\documentclass[Afour,sageh,times]{sagej}
\usepackage{moreverb,url}

\usepackage{multicol}

\usepackage{amsmath,amssymb,amsfonts}

\DeclareMathOperator*{\argmin}{arg\,min}
\DeclareMathOperator*{\argmax}{arg\,max}
\usepackage{graphicx}
\usepackage{textcomp}
\usepackage[dvipsnames,table]{xcolor}
\usepackage{algorithm}
\usepackage{algorithmic}
\usepackage[utf8]{inputenc}
\usepackage{array}
\usepackage{multirow}
\usepackage{microtype}
\usepackage{stfloats}
\usepackage{verbatim}
\usepackage{tikz}
\usepackage{svg}
\usepackage{wrapfig}
\usepackage{csquotes}
\usepackage{subcaption}

\DeclareRobustCommand{\say}[1]{\textquote{#1}}

\usepackage{booktabs}
\usepackage{amsthm}
\theoremstyle{definition}
\newtheorem{definition}{Definition}[section]

\usepackage{enumitem}
\usepackage{setspace}
\usepackage{color, colortbl}

\definecolor{Gray}{gray}{0.9}
\definecolor{LightCyan}{rgb}{0.88,1,1}
\definecolor{LightPink}{rgb}{1,0.89,0.88}
\definecolor{LightGreen}{rgb}{0.85,1,0.8}

\usepackage{xcolor}

\theoremstyle{plain}
\newtheorem{theorem}{Theorem}[section]
\newtheorem{lemma}[theorem]{Lemma}
\newtheorem{assumption}[theorem]{Assumption}

\theoremstyle{remark}

\usepackage{empheq}

\usepackage{soul}

\usepackage[dvipsnames]{xcolor} 
\usepackage[colorlinks,bookmarksopen,bookmarksnumbered,citecolor=blue,urlcolor=blue]{hyperref}

\newcommand\BibTeX{{\rmfamily B\kern-.05em \textsc{i\kern-.025em b}\kern-.08em
T\kern-.1667em\lower.7ex\hbox{E}\kern-.125emX}}

\def\volumeyear{2026}

\begin{document}


\runninghead{Xue and Figueroa}

\title{No Undesirable Equilibria, No Collisions: Combining Modulation and Control Barrier Functions for Liveness and Safety in Dynamic Collision Avoidance}

\author{Yifan Xue\affilnum{1} and Nadia Figueroa\affilnum{1}}

\affiliation{\affilnum{1} University of Pennsylvania, USA }

\corrauth{Yifan Xue, University of Pennsylvania,
the Department of Mechanical Engineering and Applied Mechanics, Philadelphia, PA 19104, USA.}

\email{yifanxue@engineering.upenn.edu}

\begin{abstract}
Control Barrier Function Quadratic Programs (CBF-QPs) provide formal safety guarantees for general control-affine systems and accommodate actuation constraints, but do not inherently guarantee \textit{liveness}, i.e., sustained progress toward and convergence to a goal. They may admit undesirable equilibria or dysfunctional circulation that impede goal convergence. In contrast, Modulation of Dynamical Systems (Mod-DS) promotes liveness through geometric reshaping of nominal vector fields, but is primarily suited to fully actuated systems and does not directly accommodate general actuation constraints. We establish theoretical connections between these paradigms, showing that normal Mod-DS is a special case of CBF-QP and that reference Mod-DS is linked to CBF-QP through a shared formulation. These connections motivate the \textbf{Modulated CBF-QP (MCBF-QP)} framework, which incorporates geometric guidance from Mod-DS into CBF-QPs to enhance liveness while retaining formal safety and control-affine generality. We develop reference MCBF-QP, which reduces undesirable equilibria and dysfunctional circulation for star-shaped obstacles, and on-manifold MCBF-QP, which eliminates undesirable equilibria and reduces dysfunctional circulation for arbitrarily shaped concave obstacles. Although no Lyapunov function is explicitly constructed, we prove that the tangent guidance constraint introduced into the CBF-QP formulation implicitly enforces the decrease condition of a hybrid Lyapunov function, yielding asymptotic goal convergence and thus liveness under the stated assumptions. Extensive simulation and hardware experiments demonstrate safe and live obstacle avoidance with the proposed MCBF-QP framework in static and dynamic environments. Project website, code, and videos are available at \url{https://yifanxueseas.github.io/mcbf_web/}.
\end{abstract}
\maketitle
\section{Introduction}
\label{Introduction}
Recent developments in autonomous systems have brought increasing research efforts into the field of robot obstacle avoidance and safe control system design. Autonomous robots colliding with environmental obstacles or their co-workers would not only reduce work efficiency but also injure users in safety-critical tasks ranging from industrial automation, household human-robot interaction to autonomous driving \citep{ROB-052}. Over the years, a broad spectrum of collision avoidance approaches has demonstrated strong empirical performance in lab settings, spanning graph-search methods, data-driven learning, sampling-based planners, closed-form reactive strategies, and control optimization \citep{lavalle1998rapidly, hart1968formal, dolgov2008practical,lutjens2019safe, zhu2022collision, hope,williams2017model, khansari2012dynamical, LukesDS, onManifoldMod, mpcdc2021safety}. Yet, it is noteworthy that strong performance in controlled test environments does not directly translate to reliable deployment in fully autonomous systems. In practice, robots inevitably encounter scenarios beyond those considered during design and evaluation.  

This motivates a \textbf{stricter notion of safety}, one where a trustworthy controller should provide formal guarantees of collision avoidance during operation or explicitly detect and report infeasibility when the formulation cannot admit a safe action. In addition, we argue that the notion of \textbf{liveness}, i.e., the property that a robotic system will eventually make progress toward and achieve its specified goals, is fundamental for reliable autonomous navigation.


In the robotics literature, collision avoidance with formal or structural safety guarantees has been pursued through several paradigms, including (i) \textbf{optimization-based} methods such as Control Barrier Functions (CBFs) and collision-constrained Model Predictive Control (MPC-DC, MPC-CBF) \citep{cbfclf,ames2019control,mpcdc2021safety}, (ii) \textbf{reachability-based} methods such as Hamilton-Jacobi (HJ) reachability \cite{hj_reachability}, and (iii) \textbf{closed-form} potential-field-inspired methods that \textbf{geometrically warp} the navigation space, such as navigation functions and Dynamical System Motion Policy Modulation (Mod-DS) \citep{apf1989real,navigationf1992exact,harmonic1997real,khansari2012dynamical,LukesDS,billard2022learning}. Within categories (i) and (ii), a broad range of control and navigation approaches have been developed to promote or establish liveness, including Control Lyapunov Functions (CLFs), reachability analysis with liveness filtering, and terminal-cost or terminal-constraint design in optimization-based control \citep{ames2019control,reach_for_liveness,MAYNE2000789}. Nevertheless, ensuring liveness for reach-avoid problems remains challenging in complex environments, particularly in the presence of concave geometries and dynamic obstacles. Existing approaches often face either computational challenges that limit their applicability to highly reactive settings or difficulties in constructing Lyapunov or terminal functions whose desirable properties are preserved across general collision-avoidance scenarios. On the other hand, geometric closed-form approaches in category (iii), such as Mod-DSs, provide a direct, real-time means of promoting goal convergence through the explicit construction of local obstacle-aware vector fields \citep{khansari2012dynamical,LukesDS,onManifoldMod}. Yet these methods are generally restricted to fully actuated systems and may induce discontinuous or rapidly varying control inputs, limiting their applicability to general control-affine systems and systems with broader actuation constraints.


Reactive collision avoidance in dynamic environments requires \textbf{both safety and liveness}, for which CBF-QP and Mod-DS offer complementary strengths.
CBF-QPs provide a generic framework for minimally modifying a nominal controller while enforcing safety through affine constraints derived from set invariance \citep{ames2019control}. They apply to general control-affine systems and accommodate actuation limits. However, CBF-QPs may admit \textbf{undesirable equilibria} even near simple convex obstacles \citep{clfcbfEquilibria}. Augmenting the QP with a soft affine CLF constraint can promote liveness and eliminate boundary equilibria \citep{ames2019control,clfcbfEquilibria}, but does not guarantee goal convergence. Near obstacle boundaries, particularly for concave obstacles, oscillatory or limit-cycle-like motion may persist despite the absence of equilibria, which we refer to as \textbf{dysfunctional circulation}. CLF design for reach-avoid problems is challenging because obstacle avoidance may require temporary motion away from the goal, conflicting with the descent condition of a goal-oriented CLF. Ensuring global convergence while accounting for such obstacle-dependent motion generally requires explicit environment information, which may be unavailable or time-varying in unknown and dynamic environments. Thus, CBF-CLF formulations provide a principled means of augmenting safety with liveness but do not fully address the broader liveness challenges induced by complex obstacle geometries.

Mod-DS approaches, inspired by Harmonic Potential Functions \citep{harmonic1997real} and Navigation Functions \citep{navigationf1992exact}, achieve obstacle avoidance by modulating a nominal autonomous dynamical system with a state-dependent modulation matrix. This matrix is constructed from obstacle distance and gradient information to reshape the nominal flow around obstacles. The original Mod-DS formulation \citep{khansari2012dynamical}, referred herein as \textbf{normal Mod-DS}, reduces undesirable equilibria relative to Artificial Potential Fields (APFs) but may still admit boundary equilibria when the nominal vector field becomes collinear with the obstacle gradient. 
\textbf{Reference Mod-DS} \citep{LukesDS} addresses this issue for convex and star-shaped obstacles by introducing a gradient offset and leveraging contraction theory. Under a nominal linear DS, it reduces the remaining equilibria to a single saddle trajectory per obstacle. \cite{onManifoldMod} further establish that this residual equilibrium is unavoidable in this setting due to the Hairy Ball Theorem \citep{hairyball_proof_1979}; and propose \textbf{on-manifold Mod-DS}, which eliminates undesirable equilibria for arbitrary obstacle geometries by treating obstacle boundaries as navigable manifolds incorporating gradient, tangent, and curvature information into the modulation. While liveness is guaranteed, Mod-DS methods rely on closed-form geometric constructions restricted to fully actuated systems. They cannot directly accommodate actuation limits or kinematic/dynamic constraints found in general control-affine systems.

\begin{figure}[!tbp]
    \centering
    \begin{subfigure}{0.47\linewidth}
        \centering
        \includegraphics[
            width=\linewidth,
            trim={23 0 23 0},
            clip=true
        ]{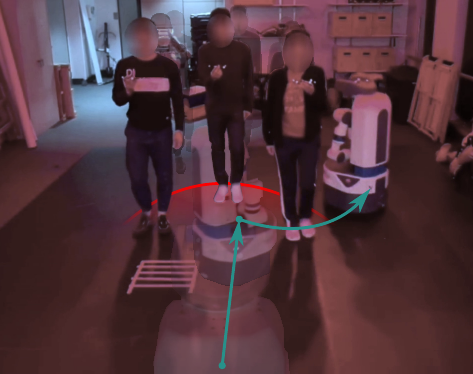}
        \caption{crowd navigation}
        \label{fig:crowd_nav}
    \end{subfigure}
    \hfill
    \begin{subfigure}{0.47\linewidth}
        \centering
        \includegraphics[
            width=\linewidth,
            trim={0 0 0cm 0.35cm},
            clip=true
        ]{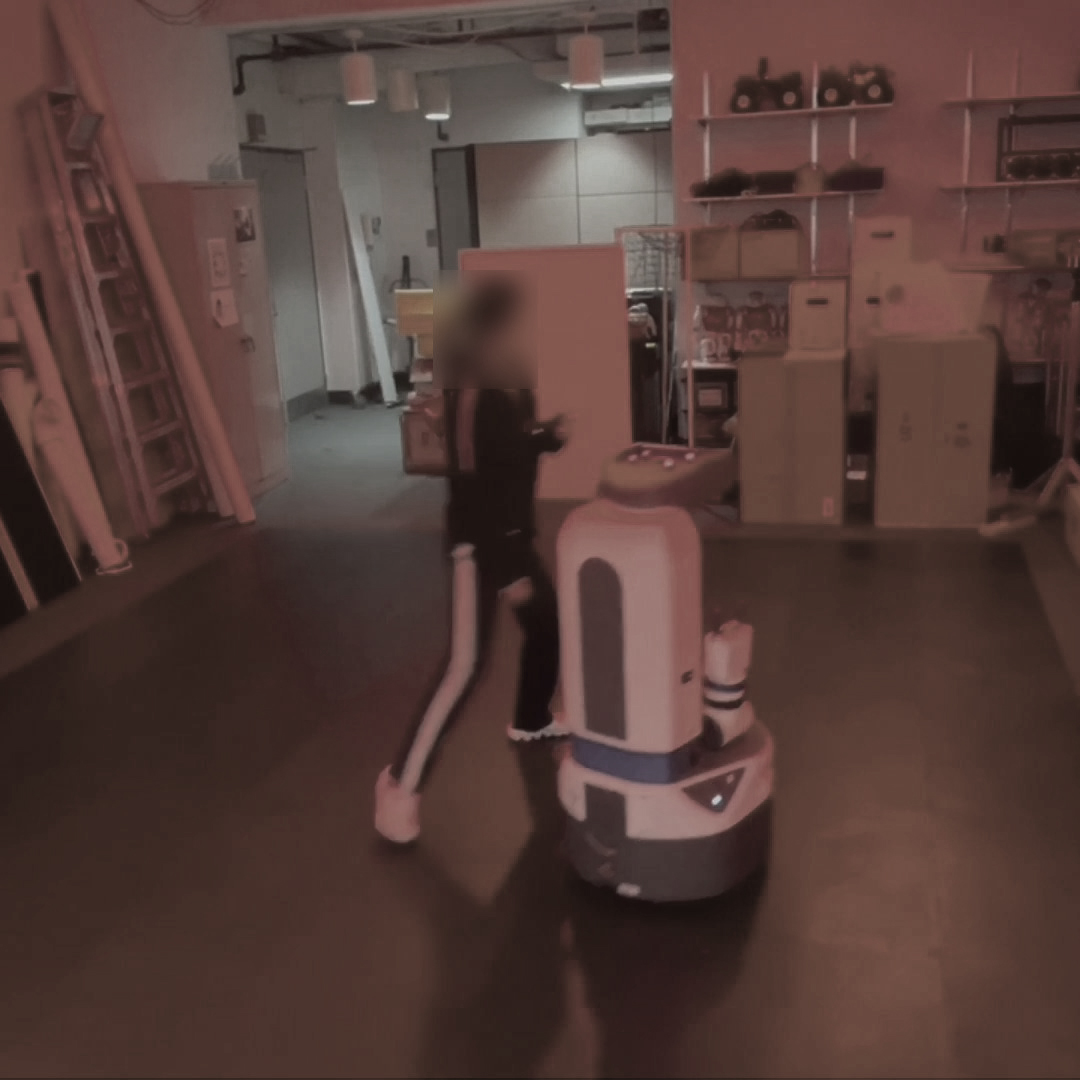}
        \caption{adversarial attack scenarios}
        \label{fig:adversarial_attack}
    \end{subfigure}

    \vspace{0.5em}

    \begin{subfigure}{0.47\linewidth}
        \centering
        \includegraphics[
            width=\linewidth,
            trim={0 0 0 0},
            clip=true
        ]{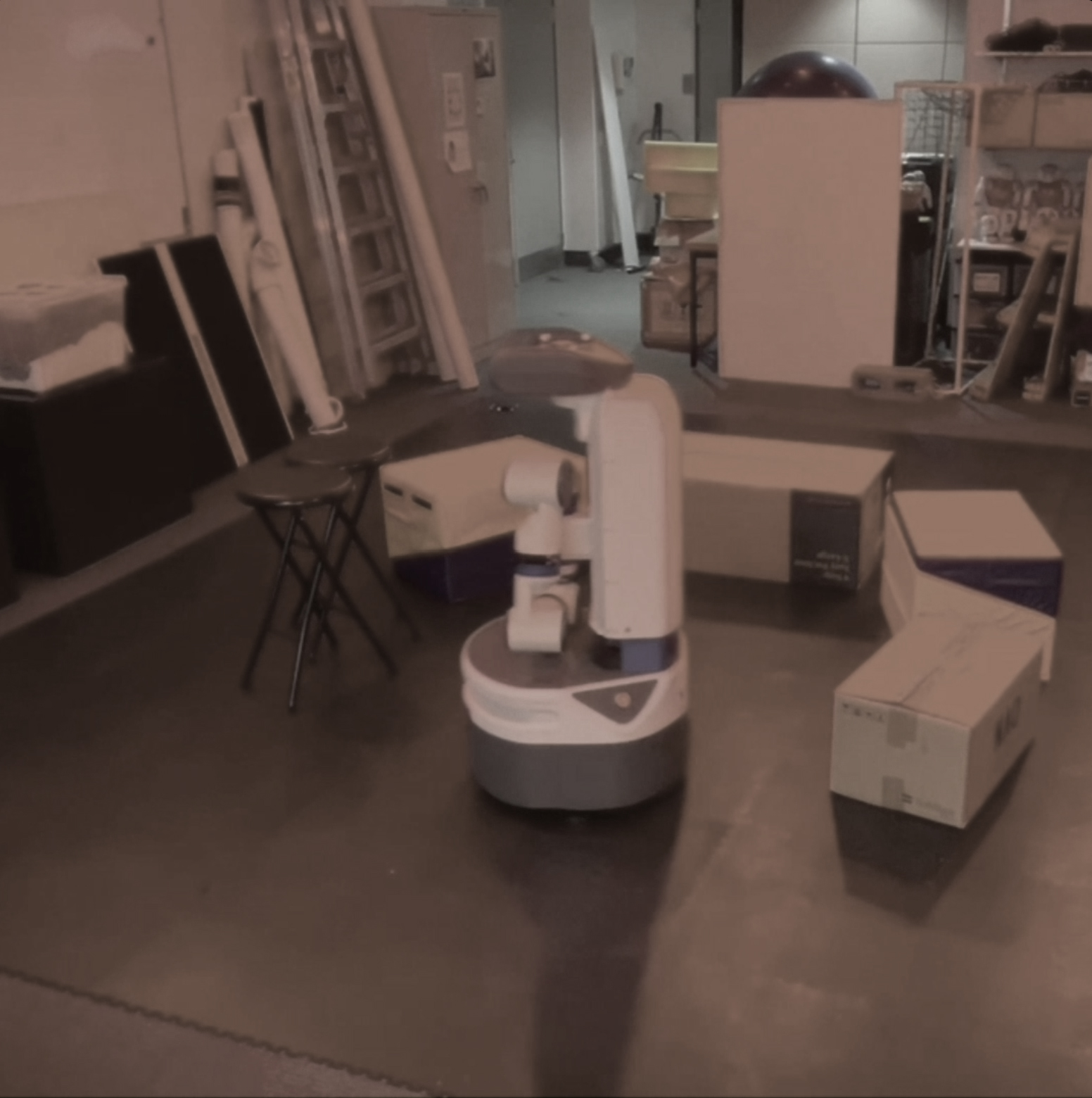}
        \caption{concave obstacle navigation}
        \label{fig:concave_nav}
    \end{subfigure}
    \hfill
    \begin{subfigure}{0.47\linewidth}
        \centering
        \includegraphics[
            width=\linewidth,
            trim={0 0 0 0},
            clip=true
        ]{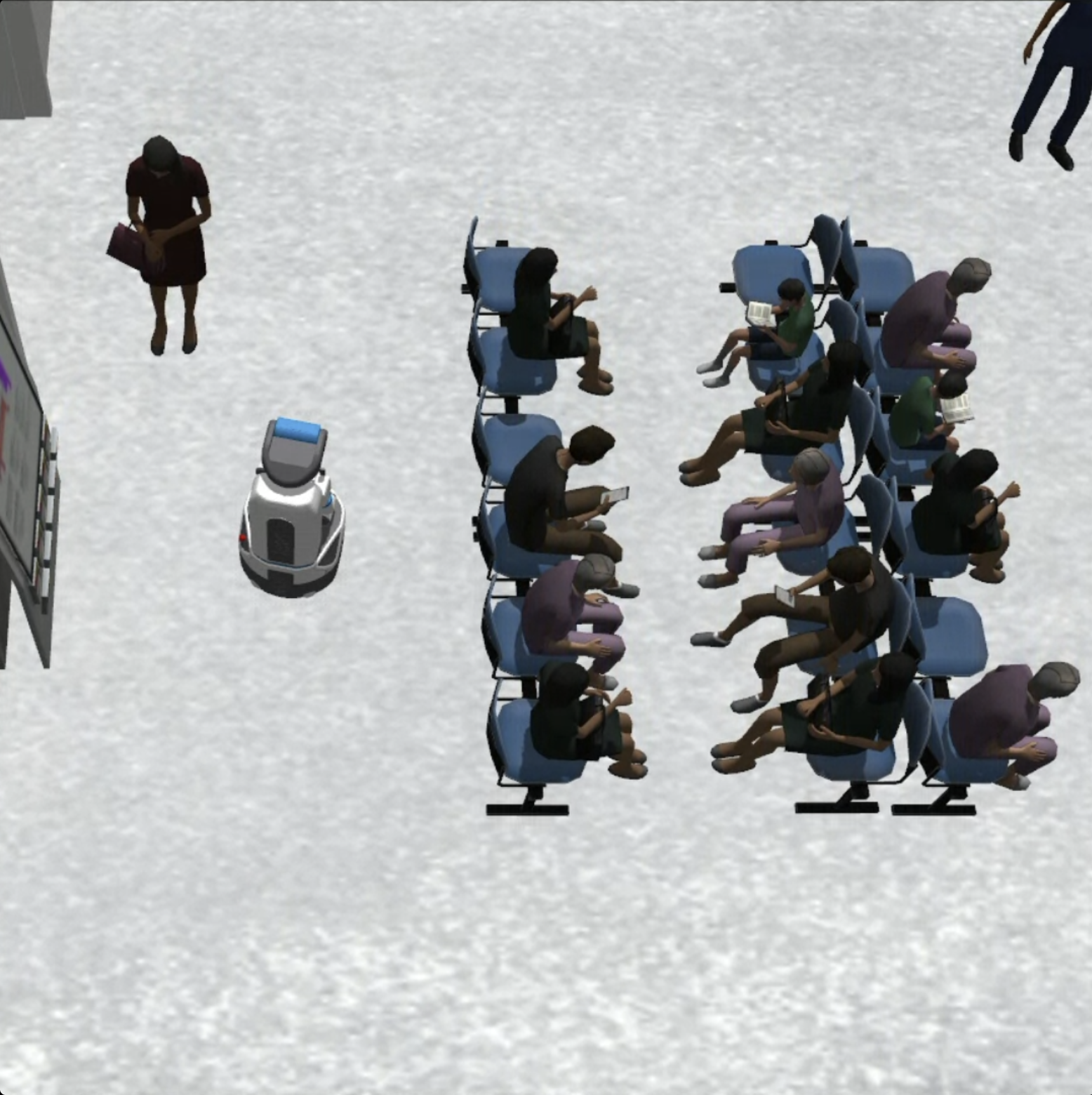}
        \caption{hospital navigation}
        \label{fig:hospital_nav}
    \end{subfigure}

    \caption{Real-world (a-c) and simulated experiments in Gazebo (d) with the proposed \textbf{MCBF-QP} strategy (Section~\ref{sec:mod-cbf}) for undesirable-equilibrium-free navigation in complex and dynamic environments. \label{fig:mainfig}}
    \vspace{-15pt}
\end{figure}

In this paper, we present a novel approach that bridges the trade-off between the safety robustness and geometric generality of CBF-QPs and the liveness properties of Mod-DS. By analyzing their behavioral similarities, differences, and theoretical connections, we introduce the \textbf{Modulated CBF-QP (MCBF-QP)} framework, which incorporates geometric principles from Mod-DS into the CBF-QP formulation. The proposed formulation can be viewed as \textbf{an extension of existing CBF-CLF formulations}. Although it does not explicitly construct a Lyapunov function, we prove that its tangent guidance constraint implicitly enforces the decrease condition of a hybrid Lyapunov function measuring progress toward the goal, thereby ensuring asymptotic goal convergence. The major contributions of this paper are:
\begin{enumerate}[noitemsep]
\item A theoretical analysis establishing the equivalence and key differences between Mod-DS and CBF-QP methods (\autoref{sec:theoretical-analysis}).

\item A family of \textbf{Modulated CBF-QP (MCBF-QP)} variants that reduce undesirable equilibria and eliminate dysfunctional circulation while preserving compatibility with control-affine systems: (i) the Reference Modulated CBF-QP (R-MCBF-QP) for star-shaped obstacles (\autoref{sec:reference mod-cbf}); and (ii) the on-manifold Modulated CBF-QP (OnM-MCBF-QP) for arbitrarily shaped concave obstacles (\autoref{sec:on-manifold cbf-qp}).


\item Theoretical guarantees of undesirable-equilibrium reduction or elimination and asymptotic goal convergence for both MCBF-QP variants (\autoref{theorem:r-mcbf-equilibrium}, \autoref{theorem:r-mcbf-convergence}, \autoref{theorem:om-mcbf-equilibrium}, \autoref{theorem:om-mcbf-convergence}).

\item Extensive validation of the MCBF-QP controllers in simulation and on real robotic hardware (\autoref{sec:experiment}).
\end{enumerate}

\noindent \textbf{Paper Organization.} 
\autoref{sec:related work} reviews real-time safe and liveness-aware navigation, highlighting limitations in addressing undesirable equilibria and dysfunctional circulation in concave environments while preserving control-affine system dynamics. \autoref{sec:problem statement} defines the robot dynamics, safety and liveness objectives, scope of this work, and assumptions considered throughout the paper. \autoref{sec:background} introduces the CBF-QP and Mod-DS frameworks. Building on these preliminaries, \autoref{sec:theoretical-analysis} establishes theoretical connections between CBF-QPs and Mod-DSs and analyzes their shared mechanisms and the key differences in Mod-DSs that improve navigation liveness. Finally, based on the theoretical connections established, \autoref{sec:mod-cbf} proposes the MCBF-QP framework.

\section{Related Work}
\label{sec:related work}
The topic of collision avoidance in robotics has been extensively studied across multiple paradigms spanning graph-search methods, data-driven learning, sampling-based planners, closed-form reactive strategies, and control optimization \citep{lavalle1998rapidly, hart1968formal, dolgov2008practical,lutjens2019safe, zhu2022collision, hope,williams2017model, khansari2012dynamical, LukesDS, onManifoldMod, mpcdc2021safety}. Despite this progress, many existing approaches lack theoretical foundations to ensure robot safety. Data-driven methods, featuring deep reinforcement learning, can capture complex behaviors but often degrade outside their training distributions and rely on auxiliary safety filters to conservatively clip unsafe actions \citep{lutjens2019safe, zhu2022collision, hope}. Sampling-based methods, such as Model Predictive Path Integral (MPPI) \citep{williams2017model}, offer real-time adaptability but may produce unsafe actions, as collision avoidance is typically encoded in the objective rather than enforced as a hard constraint.

Moreover, methods with theoretical guarantees in static environments often struggle in dynamic settings due to computational limitations. Graph-search-based planners, including Rapidly-exploring Random Trees (RRT), A*, and Hybrid A*, can generate globally feasible paths in static environments, but are too computationally expensive to update in real time when faced with dynamic obstacles that deviate from preplanned trajectories \citep{lavalle1998rapidly, hart1968formal, dolgov2008practical}. Traditional trajectory optimization approaches based on MPC, such as MPC-DC and MPC-CBF, provide reliable obstacle avoidance in relatively simple environments, but their computational cost scales poorly with the number and complexity of nonconvex constraints \citep{mpcdc2021safety, cosner2023robust}. Simplified obstacle representations, such as spherical approximations, can reduce this cost and enable near-real-time performance, but may be overly conservative and eliminate feasible solutions \citep{mpcdc2021safety}. In practice, many approaches incorporate CBF-QPs as a low-level safety filter to achieve provably safe motion planning in dynamic environments\citep{drl_cbf, mppi_cbf1, mppi_cbf2, mpc_cbf1, mpc_cbf2}. However, the performance of such filtered systems—whether the high-level controller is based on learning \citep{drl_cbf}, sampling \citep{mppi_cbf1, mppi_cbf2}, or trajectory optimization \citep{mpc_cbf1, mpc_cbf2}—are affected by undesirable equilibria and dysfunctional circulation, especially in concave environments. This motivates our design of liveness-enhanced CBF-QPs by mathematically incorporating Mod-DS’s equilibrium reduction capabilities into the QP formulation.

Several efforts have compared or combined collision avoidance controllers for their distinctive advantages. \cite{singletary2021comparative} has shown that CBF-QPs, under fully actuated assumptions, are effectively equivalent to APFs, which are known to exhibit spurious equilibria requiring exploratory strategies to escape \citep{spong}. Other studies compare CLF-CBF QPs with Hamilton–Jacobi reachability methods in static environments \citep{li2021comparison}, or contrast optimization-based safe control with deep neural approaches; for instance, \citep{9152161deep} examines differences between deep learning and MPC for adaptive cruise control. However, a comprehensive comparison of reactive obstacle avoidance techniques in real-time dynamic settings remains absent. Although Mod-DS methods have gained traction for their computational efficiency and strong guarantees on reducing topologically unnecessary equilibria \citep{LukesDS,billard2022learning,onManifoldMod}, they are rarely included in these evaluations.

Several approaches have sought to enhance the liveness of CBF-QPs. \cite{notomista2021safety} reduce undesirable equilibria by mapping the robot-obstacle space to a ball world through an online diffeomorphism; however, analytical diffeomorphisms are unavailable for general non-star-shaped obstacles, and is limited to static environments. \cite{cbf_circulation,keyumarsi2025circulation} introduce circulation constraints for non-star-shaped obstacles, but optimize over fully actuated control inputs rather than the original system inputs, sacrificing compatibility with general control-affine dynamics. \cite{clfcbfEquilibria,tracking_cbf} preserve control-affine compatibility and mitigate boundary undesirable equilibria through Lyapunov shaping and auxiliary control, but are primarily developed and evaluated for circular obstacles and remain susceptible to dysfunctional circulation in concave environments. 

\textbf{To the best of our knowledge,} no existing approach jointly (i) eliminates on-boundary equilibria induced by conflicts between nominal control and safety constraints, (ii) preserves the original control-affine system dynamics, and (iii) mitigates dysfunctional circulation for general concave obstacles. The proposed MCBF-QP addresses these limitations.

\section{Problem Statement} \label{sec:problem statement}
Given the robot dynamics defined in \autoref{sec:dynamics} and the \textbf{local} dynamic environment $\mathcal{O}$ and controller assumptions stated in \autoref{sec:assumptions}, the objective is to design a control law $u$ such that $\lim_{t \to \infty} x(t) = x^*$, while ensuring the resulting trajectories are free of collisions and undesirable equilibria as defined in \autoref{sec:safety def} and \autoref{sec:local minimum def}.
\subsection{Robot Dynamics}
\label{sec:dynamics}
We restrict the study of the proposed MCBF-QPs to general \emph{control-affine systems} defined as,
\begin{equation}
    \dot{x} = f(x) + g(x)u, 
    \label{eq:affine system}
\end{equation}
where $x \in \mathbb{R}^d$, $u \in \mathbb{R}^p$, and $f$ and $g$ are Lipschitz continuous functions. A special class of control-affine systems are fully actuated systems, whose controllable degrees of freedom (DoF) are equal to total DoF, i.e. $p=d$ and can be modeled as single integrator robots, mathematically defined as:
\begin{equation}
\label{eq:fully actuated system}
\dot{x} =  u. 
\end{equation}
In mobile robotics, although fully actuated systems such as omnidirectional drive robots can be faithfully modeled as Eq.~\eqref{eq:fully actuated system} exist, many robotic platforms are subject to nonholonomic constraints and therefore require differential or inverse kinematic formulations. Such underactuated robots include differential-drive and Ackermann-steering vehicles.
\vspace{-5pt}
\subsection{Safety in Dynamic Obstacle Avoidance}
\label{sec:safety def}
\begin{figure}[!tbp]
    \centering
    \subfloat[$h_{\text{conv}}$]{%
        \includegraphics[width=0.32\linewidth]{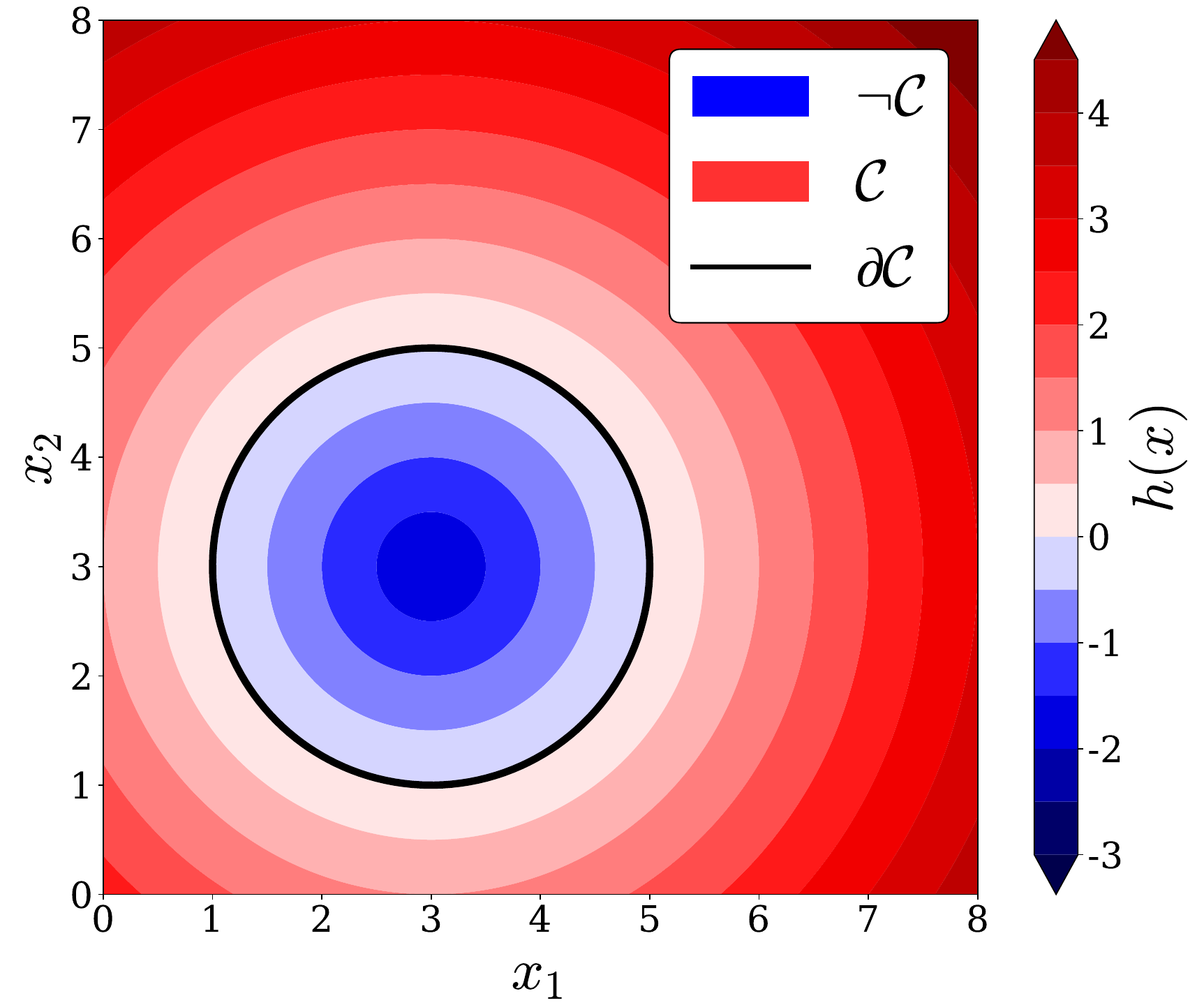}%
        \label{fig:isoline_convex}
    }
    \subfloat[$h_{\text{star}}$]{%
        \includegraphics[width=0.32\linewidth]{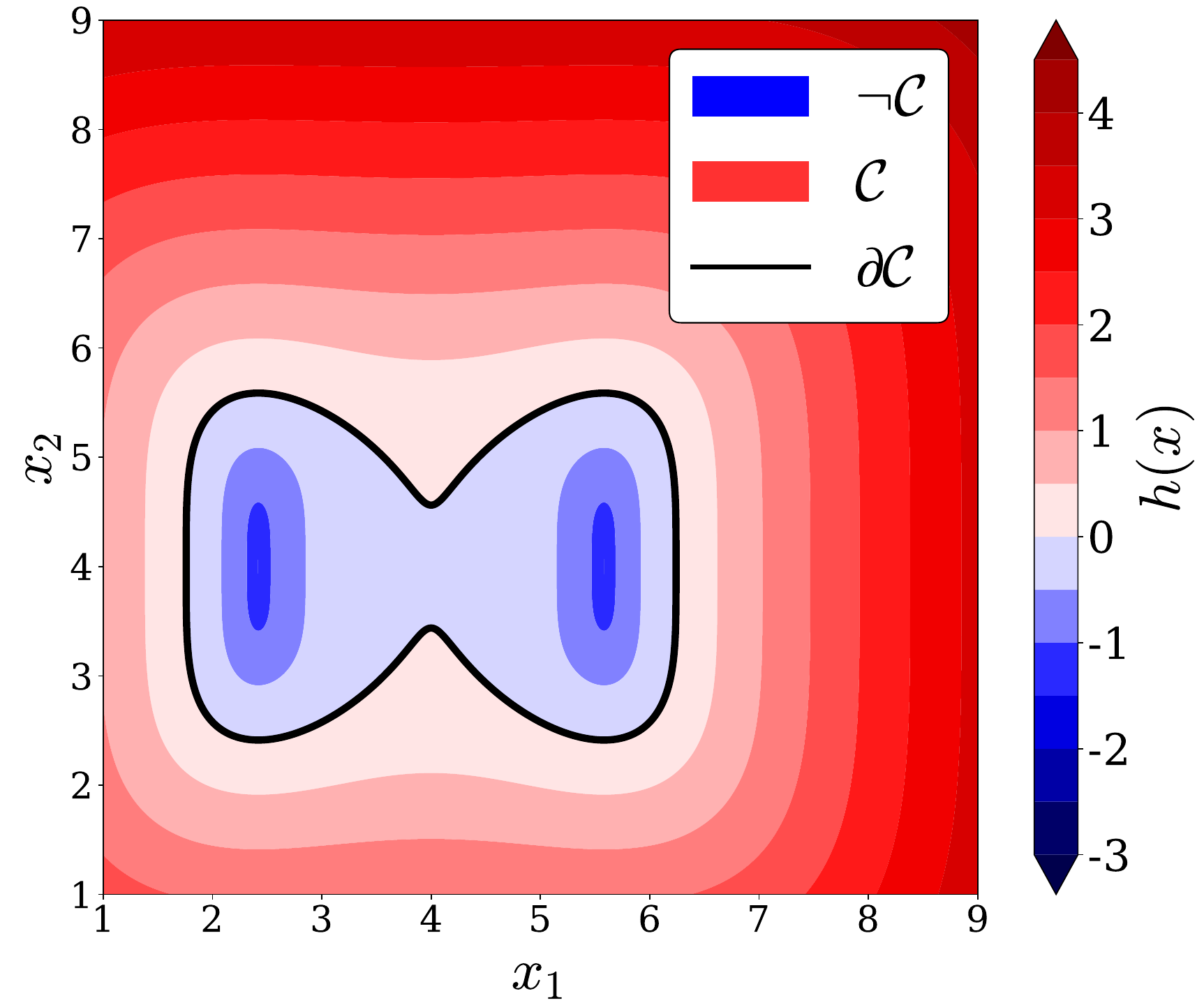}%
        \label{fig:isoline_star}
    }
    \subfloat[$h_\text{nstar}$]{%
        \includegraphics[width=0.32\linewidth]{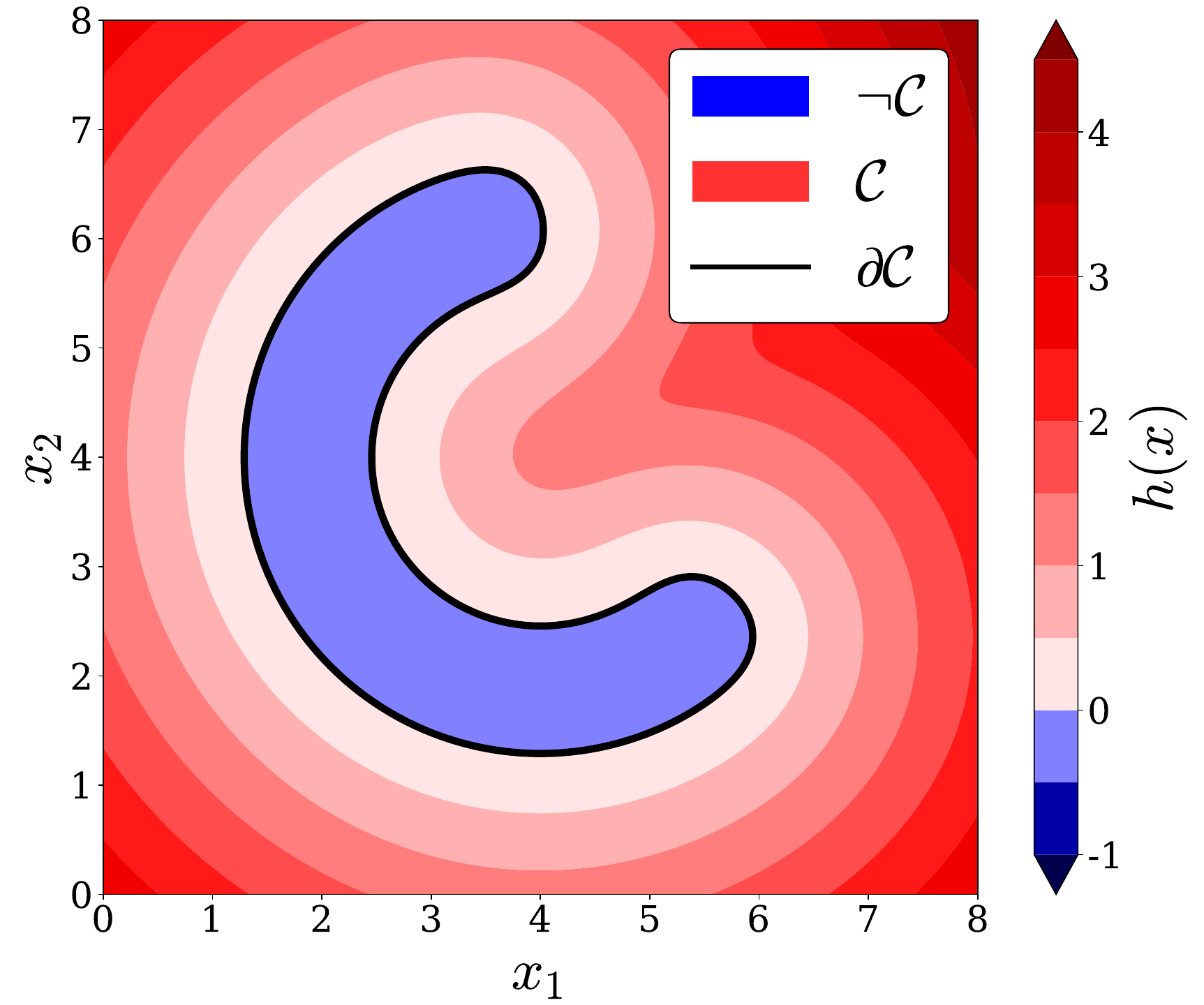}%
        \label{fig:isoline_nstar}
    }
    \caption{Isolines displaying (a) $h_{\text{conv}}$, (b) $h_{\text{star}}$, and (c) $h_\text{nstar}$ in the proximity of respectively a circle, a funnel, and a C-shaped obstacle. \label{fig:isoline} 
    }
    \vspace{-10pt}
\end{figure}

As shown in \autoref{fig:isoline}, Mod-DSs and CBF-QPs formalize robot safety and synthesize safe control algorithms using boundary functions. Given robot state $x \in \mathbb{R}^d$ and obstacle state $x_o \in \mathbb{R}^{d'}$ for an obstacle $o$ in the nearby obstacle set $\mathcal{O}$, a continuously $\mathcal{C}^1$ differentiable function $h_o:\mathbb{R}^d \times \mathbb{R}^{d'} \rightarrow \mathbb{R}$ is a boundary function iff the safe set $\mathcal{C}_o$ (outside the obstacle), the boundary set $\partial \mathcal{C}_o$ (on the boundary of the obstacle), and the unsafe set $\neg \mathcal{C}_o$ (inside the obstacles) of the system can be defined as, 
\begin{equation}
\label{eq:safe_sets}
\begin{aligned}
\mathcal{C}_o & =\{x \in \mathbb{R}^d, x_o \in \mathbb{R}^{d'}: h_o(x,x_o)>0\}\\
\partial \mathcal{C}_o & =\{x \in \mathbb{R}^d, x_o \in \mathbb{R}^{d'}: h(x,x_o)=0\}\\
\neg \mathcal{C}_o & =\{x \in \mathbb{R}^d, x_o \in \mathbb{R}^{d'}: h(x,x_o)<0\}
\end{aligned}
\end{equation}

Note that the boundary function $h_o(x, x_o)$ can be simplified to $h_o(x)$ when the obstacle $o$ is static (i.e., $x_o$ is constant), and to $h(x, x_o)$ when the environment contains only a single obstacle. In practice, $h_o(x, x_o)$ is often represented as the distance from the controlled agent to the obstacle surface, i.e., the signed distance function. Given environments defined by $h_o$ functions, the goal of a safety-critical controller is to generate an admissible input $u$ that ensures the robot state $x$ remains within the safe set $C$, defined in Eq.~\eqref{eq:safe region}, and ultimately reaches a target state $x^* \subset C$.
\begin{equation}
\mathcal{C}
=
\left\{
(x,\{x_o\}):
h_o(x,x_o)>0,\quad
\forall o\in\mathcal{O}
\right\}
\label{eq:safe region}
\end{equation}

\subsection{Undesirable Equilibrium \& Dysfunctional Circulations}
\label{sec:local minimum def}
Besides safety, a controller's navigation capability in complex environments also depends on how often it causes the robot to stop before reaching the intended target $x^*$. Such phenomenon has been referred to as \emph{undesirable equilibria} in \cite{clfcbfEquilibria, cbf_circulation, tracking_cbf}, \emph{critical points} in \cite{KODITSCHEK1990412} and \emph{spurious attractors} in \cite{billard2022learning, onManifoldMod}. For the rest of this paper, we adopt the term \emph{undesirable equilibria}, following previous CBF-QP works. Note that, in addition to conflicts between nominal control and safety constraints, robots using CBF-QP filters may also stop prematurely in dynamic-obstacle scenarios where the QP optimization becomes infeasible under actuation limits. In this work, we focus on stopping behaviors arising from geometric infeasibility (i.e., on-boundary equilibria) induced by conflicts between nominal control and safety constraints, while excluding optimization infeasibility-arising from mutual constraint conflicts-from consideration. A solution to alleviate the optimization infeasibility issue is proposed in \cite{admm_mcbf_lca} via a layered architecture with offline feedforward and online feedback loops.

\begin{definition}[\textbf{Undesirable Equilibria} in Robot Navigation]
\label{def:equilibria}
Let $\dot{x} = f(x,u)$ denote the robot dynamics, where $f:\mathbb{R}^d \times \mathbb{R}^p \to \mathbb{R}^d$. 
A point $x \in \mathbb{R}^d$ is called an \emph{undesirable equilibrium} if
\[
f(x, u) = 0 \quad \text{and} \quad x \ne x^*,
\]
i.e., the robot freezes at a point other than the target.
\end{definition}

In challenging obstacle avoidance scenarios, especially those involving concave obstacles, convergence to the target is not guaranteed despite the absence of undesirable equilibria. Such failures are caused by dysfunctional circulation, defined as oscillatory or limit-cycle-like behaviors that arise in saddle regions near obstacle boundaries, where the robot continuously circulates without being able to move sufficiently away from the obstacle.

\begin{definition}[\textbf{Dysfunctional Circulation} in Robot Navigation]
\label{def:circulation}
Let $\dot{x} = f(x,u)$ denote the robot dynamics, where $f:\mathbb{R}^d \times \mathbb{R}^p \to \mathbb{R}^d$. 
A robot is said to be trapped in \textit{dysfunctional circulation} wrt. obstacle $o$ if its state trajectory $x(t) \in \mathbb{R}^d$ satisfies the following conditions:
\[
f(x(t),u(t)) \neq 0, \quad h(x(t), x_o)\leq \epsilon_o, \quad \text{and} \quad x(t) \neq x^*,
\]
for all $t \in [0,\infty)$ where $\epsilon_o > 0$ is a small positive constant defining the maximum allowable deviation from obstacle $o$ during the circulation process.
\end{definition}

\vspace{-10pt}
\subsection{Scope \& Assumptions}
\label{sec:assumptions}

\textbf{\indent Robot: }We consider point robots, or applications in which the robot geometry can be approximated by appropriate inflation of safety margins. Scenarios requiring precise geometric fitting, such as navigating through tightly constrained passages, are excluded from consideration. 

\textbf{Controller: }The safe controller takes the current robot state $x$ and nominal control $u_{\mathrm{nom}}$ as inputs and outputs a safe control action $u$. We assume that the controller has sufficient \textbf{local} environmental information to accurately evaluate the obstacle boundary functions $h_o(x,x_o)$ for all $o\in\mathcal{O}$. The online acquisition and accumulation of environment representations through perception without prior knowledge of the environment are outside the scope of this paper and are discussed in our follow-up work \cite{li2026memoryawaremultisensorperceptionefficient}. As is standard in robotic navigation and motion planning, the nominal control is designed to drive the system toward the desired target in the absence of safety intervention, subject to bounded velocity.

\begin{assumption}[\textbf{Velocity Bounds}]
\label{assumption:velocity-bounds}
There exists a positive lower bound on the nominal velocity and a finite upper bound on the resulting closed-loop velocity. That is, there exist constants $\underline{v}_{\mathrm{nom}},\bar v>0$ such that
\begin{equation}
\label{eq:velocity_bound}
\|f(x)+g(x)u_{\mathrm{nom}}(x)\|
\geq
\underline{v}_{\mathrm{nom}},
\qquad
\|\dot{x}\|\leq\bar v.
\end{equation}
\end{assumption}

\begin{assumption}[\textbf{Nominal Controller Convergence}]
\label{assumption:nominal-convergence}
The nominal controller $u_{\mathrm{nom}}$ asymptotically stabilizes the target state $x^*$ in the absence of safety intervention. Specifically, there exists a continuously differentiable, positive definite, and proper function $V_N:\mathcal{X}\rightarrow\mathbb{R}_{\geq 0}$ satisfying
\begin{equation}
\label{eq:VN_properties}
V_N(x^*)=0,
\qquad
V_N(x)>0
\quad
\forall x\neq x^*,
\end{equation}
such that, along the nominal closed-loop dynamics
\begin{equation}
\dot{x}_{\mathrm{nom}}
=
f(x)+g(x)u_{\mathrm{nom}}(x),
\end{equation}
\begin{equation}
\label{eq:VN_nominal_descent}
\nabla V_N(x)^\top\dot{x}_{\mathrm{nom}}
\leq 0,
\end{equation}
with strict inequality for $x\neq x^*$ within the region of attraction. In addition, the nominal closed-loop velocity satisfies
\begin{equation}
\label{eq:nominal-nonvanishing-velocity}
\dot{x}_{\mathrm{nom}}(x)
=
f(x)+g(x)u_{\mathrm{nom}}(x)
\neq \mathbf{0},
\qquad
x\neq x^*.
\end{equation}
\end{assumption}

\textbf{Obstacles: }In this work, we analyze and propose safe control methods applicable to individual obstacles of arbitrary geometry, including convex, star-shaped, and non-star-shaped concave shapes, whose size and shape remain unchanged over time. Star-shaped obstacles are defined as those containing a reference point from which every ray intersects the boundary exactly once \citep{Sakaguchi1984}. All moving obstacles are assumed to undergo rigid translational and rotational motion. The proposed methods can be applied to obstacles arising in general environments, including regular floor-plan layouts and analogous outdoor environments, as demonstrated in our follow-up work \cite{li2026memoryawaremultisensorperceptionefficient}. The arbitrary-geometry claim concerns the shape of individual obstacles and does not imply that the reactive controller alone performs high-level path planning through globally constrained environments whose arrangement forms a maze-like topology. Extensions of the proposed method to deforming, intersecting, and overlapping obstacles, as well as to dynamic obstacle motion prediction, are explored in \cite{xue2026proactivelocalminimafreerobotnavigation}.

\subsection{Notations}
We use notation $\|\cdot\|_p$ to represent the p-norm of a vector and denote $\|\cdot\|_p^q = (\|\cdot\|_p)^q$, $\forall p$, $q \in \mathbb{Z}^+$. We annotate $\langle a,b\rangle $ for the dot product between vectors a and b. $\textbf{0}_{a \times b}$ is a matrix full of 0s of size $a \times b$. Given a vector $a(\cdot)$, $\hat{a}$ represents the axis whose direction is the same as $a(\cdot)$, and $a[:k]$ represents the new vector formed by the first k elements in $a$. $L_f h(\cdot)$ is the Lie derivative of the function $h(\cdot)$ along $f(\cdot)$, defined as $L_f h(x)=\nabla_x h(x)^\top f(x)$.
Frequently-used variables in this work are summarized in Appendix \ref{appendix:var}.

\section{Preliminaries} 
\label{sec:background}
In this section, for mathematical clarity, we review the CBF-QP and Mod-DS methods in single obstacle avoidance settings, denoting the boundary function $h_o(x,x_o)$ as $h(x, x_o)$.

\subsection{Control Barrier Functions}
\label{sec:prelims-cbf}
In the CBF-QP formulation, the boundary function $h(x,x_o)$ is also known as the barrier function. To guarantee the safety of the controlled agent, CBF-QP utilizes the Nagumo Set Invariance Theorem \cite{ames2019control}.
\begin{definition}[Nagumo Set Invariance]
\begin{equation}
\label{Nagumo}
C \text{ is set invariant} \iff \Dot{h}(x,x_o) \geq 0 \; \forall x \in \partial \mathcal{C}
\end{equation}
\end{definition}
Control Barrier Functions extends Nagumo Set Invariance Theorem to a control setting~\citep{ames2019control}, where the condition $\forall x \in \partial \mathcal{C}$ is rewritten mathematically using an extended $\mathcal{K}_\infty$ function $\alpha$.
\begin{gather}
 \nonumber C \text{ is set invariant} \iff \\
 \exists u \;\; \text{s.t.} \quad \Dot{h}(x,x_o, \dot{x}_o, u) \geq  -\alpha (h(x,x_o))
 \label{eq:CBF conditions}
\end{gather}

\begin{definition}[Extended $K_{\infty}$ Functions]
\label{def:K inf}
An extended $K_{\infty}$ function is a continuous, strictly increasing
function $\alpha:\mathbb{R}\rightarrow\mathbb{R}$ satisfying
$\alpha(0)=0$ and $\lim_{h\rightarrow\pm\infty}\alpha(h)=\pm\infty$.
\end{definition}

By enforcing the CBF conditions in Eq.~\eqref{eq:CBF conditions} within a quadratic program, safety can be guaranteed through the set invariance of the safety region $\mathcal{C}$ defined in Eq.~\eqref{eq:safe region}. For general control-affine systems with nominal controller $u_{\mathrm{nom}}$, the CBF-QP is formulated as follows:
\begin{gather}
u_{\text{cbf}} = \argmin_{{u} \in \mathbb{R}^p}\; (u-u_{\text{nom}})^\top(u-u_{\text{nom}}) \label{eq:cbf-qp affine} \\
\nonumber L_fh(x,x_o) + L_gh(x,x_o)u  + \nabla_{x_o} h(x,x_o)^\top\dot{x}_o \geq \\ -\alpha (h(x,x_o)) 
\end{gather}

For fully actuated systems as in Eq.~\eqref{eq:fully actuated system}, the CBF-QP becomes,
\begin{gather}
 u_{\text{cbf}} = \argmin_{u \in \mathbb{R}^d} \; (u-u_{\text{nom}})^\top(u-u_{\text{nom}}) \label{eq:cbf-qp fully actuated} \\
\nonumber \nabla_{x} h(x,x_o)^\top u + \nabla_{x_o} h(x,x_o)^\top\dot{x}_o \geq -\alpha (h(x,x_o))
\end{gather}


\textbf{CBF-CLF:} While CBFs provide safety guarantees by enforcing forward invariance of the safe set, safety alone does not guarantee progress toward a desired goal. This motivates the combination of Control Lyapunov Functions (CLFs) and CBFs in a unified quadratic program, where the CLF promotes convergence toward the goal while the CBF enforces safety. For a positive-definite function $V(x)$ with respect to the goal $x^*$, the CLF condition is typically expressed as
\begin{equation}
\label{eq:clf-condition}
\dot{V}(x,u)
=
L_fV(x)+L_gV(x)u
\leq
-\gamma(V(x)),
\end{equation}
where $\gamma$ is an extended $\mathcal{K}_{\infty}$ function. A CLF-CBF-QP can then impose Eq.~\eqref{eq:clf-condition} together with the CBF condition in Eq.~\eqref{eq:CBF conditions}, thereby simultaneously promoting goal convergence and enforcing safety. However, satisfying the CLF and CBF conditions does not by itself guarantee liveness, as the two constraints can become incompatible in the presence of obstacles, preventing the robot from maintaining progress toward the goal and potentially resulting in undesirable equilibria \citep{notomista2021safety,clfcbfEquilibria}.

\begin{table*}[!tpb]
\caption{Properties of the three main Mod-DS variants, detailing their diagonal scaling matrix $D(x, x_o)$, basis matrix $E(x, x_o)$, and the obstacle types they can traverse without generating trapping regions in the modified DS. The table also lists the number of undesirable equilibria induced per obstacle for the navigable obstacle types.}
\centering
\begin{minipage}[t]{\textwidth}
\begin{center}
\centering
\small
 {\begin{tabular}{  m{3.2cm}  m{7.3cm}  m{1.2cm} m{1.9 cm} m{1.6cm} }
    \specialrule{\heavyrulewidth}{0pt}{0pt}
    \rowcolor{gray!10} \textbf{Property} & $D(x, x_o)$ & $d(x, x_o)$ & \textbf{Obstacle Type} & \textbf{\# Equilibria}\\
    \hline
    Normal~\citep{khansari2012dynamical} & $\lambda^*(x, x_o) = 1-\frac{1}{h(x, x_o)+1}$, $\lambda_e^* (x, x_o)= 1+\frac{1}{h(x, x_o)+1}$ & $n(x, x_o)$ & convex & 1\\
    Reference~\citep{LukesDS} & $\lambda^*(x, x_o) = 1-\frac{1}{h(x, x_o)+1}$, $\lambda_e^* (x, x_o)= 1+\frac{1}{h(x, x_o)+1}$ & $r(x, x_o)$ & convex, star & 1\\
     On-Manifold~\citep{onManifoldMod} & $D_\nabla(x, x_o) + D_\phi(x, x_o)+\varphi_h(x, x_o)D_h(x, x_o)$ & $n(x, x_o)$ & all \checkmark & 0 \checkmark\\
    \specialrule{\heavyrulewidth}{0pt}{0pt}
\end{tabular}}
\label{table:modulation types}
\end{center}
\end{minipage}
\end{table*}
\subsection{Mod-DS Approach}
\label{sec:prelims-mod}
Unlike CBF-QP that applies to general control-affine systems, Mod-DS is restricted to \textbf{fully actuated systems} shown in Eq.~\eqref{eq:fully actuated system}. Thus, for fully actuated systems we assume
\begin{equation}
    \label{eq:nominal DS}
    \dot{x}_\text{nom} = u_\text{nom} \in\mathbb{R}^{d}
\end{equation}
is the nominal vector field converging to target $x^*$. Mod-DS achieves obstacle avoidance by locally reshaping this vector via a full-rank locally active matrix $M(x,x_o)\in\mathbb{R}^{d\times d}$. Specifically, it reduces the relative speed of the robot toward the obstacle while increasing or maintaining the speed along directions tangent to it \citep{LukesDS,billard2022learning}. The smooth modulated DS $\dot{x}_\text{mod}$ in the presence of an obstacle $o$, is computed as
\begin{gather}
\label{eq:ds-modulation}
u_\text{mod} = \dot{x}_\text{mod}= M(x,x_o)(\dot{x}_{\text{nom}}-\bar{\dot{x}}_{o})+\bar{\dot{x}}_{o},\\
\label{eq:dxo}
\bar{\dot{x}}_{o}=\dot{x}_o^R+\omega_{o}\times(x-x_o^R)
\end{gather}
where $x_o^R \in \mathbb{R}^d$ is the rotation center of obstacle $o$, and $\dot{x}_{o}^{R}$, $\omega_o \in \mathbb{R}^d$ are its linear and angular velocities, respectively. In a static scenario, Eq.~\eqref{eq:ds-modulation} reduces to $\dot{x}=M(x)\dot{x}_{\text{nom}}$. 

The modulation matrix $M(x,x_o)$ is constructed analogously to an eigendecomposition, with $E(x,x_o) \in \mathbb{R}^{d \times d}$ as the basis matrix and $D(x,x_o) \in \mathbb{R}^{d \times d}$ as a diagonal scaling. 
\begin{gather}
\label{eq:M matrix}
M(x,x_o)=E(x,x_o)D(x,x_o)E(x,x_o)^{-1}\\
\label{eq:E matrix}
E(x,x_o)=[d(x,x_o) \;\; H(x,x_o)]\\
\label{eq:H matrix}
H(x,x_o)= [e_1(x,x_o) \;\; e_2(x,x_o)\; ... \; e_{d-1}(x,x_o)]\\
\label{eq:D matrix}
D(x,x_o)= \text{diag}( \lambda(x,x_o), \lambda_e(x,x_o), \dots, \lambda_e(x,x_o))
\end{gather} 
The basis matrix $E(x,x_o)$, computed using Eq.~\eqref{eq:E matrix}, consists of a direction vector $d(x,x_o) \in \mathbb{R}^{d \times 1}$ pointing toward the obstacle and a hyperplane $H(x,x_o) \in \mathbb{R}^{d \times (d-1)}$ tangent to the obstacle surface, represented by an orthonormal basis $e_1(x,x_o), \dots, e_{d-1}(x,x_o) \in \mathbb{R}^{d \times 1}$ (see Eq.~\eqref{eq:H matrix}). Velocity redistribution is enforced by the full basis matrix $E(x,x_o)$ along the directions $d(x,x_o), e_1(x,x_o), \dots, e_{d-1}(x,x_o)$, while the magnitudes along these directions are scaled by $D(x,x_o)$ as in Eq.~\eqref{eq:D matrix}. There are three main variants of Mod-DS, differing in how $E(x,x_o)$ and $D(x,x_o)$ are constructed: normal, reference, and on-manifold Mod-DS \citep{khansari2012dynamical, LukesDS, onManifoldMod}, with their properties summarized in Table \ref{table:modulation types}. Since any smooth nondegenerate vector field, such as $\dot{x}_\text{mod}$, has at least as many topological undesirable equilibria as obstacles \citep{KODITSCHEK1990412}, a variant is considered capable of handling a given obstacle type if its modified DS has at most one undesirable equilibrium per such obstacle and does not induce dysfunctional circulation. All Mod-DS approaches ensure von Neumann \textit{impenetrability} at the obstacle boundary \citep{khansari2012dynamical, LukesDS,onManifoldMod}. 

For multiple obstacles, normal and reference Mod-DS compute a weighted sum of the individually stretched velocities \citep{khansari2012dynamical, constrainedModu} and on-Manifold Mod-DS constructs a single boundary function that jointly encapsulates all obstacles \citep{onManifoldMod} as shown below.
\begin{gather}
    \bar{h}(x,x_o) = -\frac{1}{\rho}\log\left(\sum_{o \in \mathcal{O}} \exp{(-\rho [h_o(x, x_o)+1])}\right)-1,
    \label{eq:uni h}
\end{gather}
where $\rho$ is a positive user-selected constant. The smaller the value of $\rho$, the smoother the edges in the distance field. 

\textbf{Normal Mod-DS~\citep{khansari2012dynamical}:} In normal Mod-DS, the vector $n(x,x_o)$, normal to the surface of obstacle $o$ and thus to the boundary function $h(x,x_o)$, is used as the direction toward the obstacle when constructing the basis matrix $E(x,x_o)$ as defined in Eq.~\eqref{eq:E matrix}:
\begin{equation}
\label{eq:standard mod d}
d_\text{n}(x,x_o) = n(x,x_o) :=  \frac{\nabla_{x} h(x,x_o)}{\|\nabla_{x} h(x,x_o)\|_2}.
\end{equation}
Popular choices for the diagonal scaling matrix entries, $\lambda^*(x,x_o)$ and $\lambda^*_e(x,x_o)$, are listed in Table \ref{table:modulation types}, but any $\lambda(x, x_o)$ and $\lambda_e(x, x_o)$ satisfying the conditions below are feasible
\begin{equation}
\label{eq:standard mod lambda}
\lim_{h(x, x_o)\rightarrow 0} \lambda(x, x_o) = 0 \;\; \text{and} \;\; \lambda_e(x, x_o) > 0, \;\; \forall x \in \mathbb{R}^d
\end{equation}
As the earliest proposed Mod-DS variant, normal Mod-DS is dysfunctional-circulation-free only in convex obstacle avoidance scenarios \citep{khansari2012dynamical}. Moreover, it still exhibits one undesirable equilibrium per obstacle for \textit{nominal linear DSs} $\dot{x}_{\rm{nom}}$.

\textbf{Reference Mod-DS~\citep{LukesDS}:}
It differs from the normal Mod-DS only in using a reference direction $r(x, x_o)$ instead of $n(x, x_o)$ to construct $E(x, x_o)$ as:
\begin{equation}
\label{eq:reference mod d}
d_\text{r}(x, x_o) = r(x, x_o) :=  \frac{x - x_r(x_o)}{\|x - x_r(x_o)\|_2}.
\end{equation}
Here, $r(x, x_o)$ points from the robot's current state $x$ to the reference point $x_r(x_o) \in \mathbb{R}^d$ of obstacle $o$, as defined in \autoref{def:reference point}. This modification enables reference Mod-DS to circumnavigate convex and star-shaped obstacles with a single saddle trajectory converging at one undesirable equilibrium per obstacle, assuming \textit{nominal linear DSs}.

\begin{definition}[\textbf{Reference Points in Star-Shaped Obstacles}]
\label{def:reference point}
A point $x_r(x_o) \in \mathbb{R}^d$ is a \emph{reference point} iff it lies strictly inside the obstacle $\neg \mathcal{C}_o$ and every ray from $x_r(x_o)$ intersects the boundary $\partial \mathcal{C}_o$ exactly once. An obstacle or unsafe set is \emph{star-shaped} if it contains at least one reference point.
\end{definition}

\textbf{On-Manifold Mod-DS~\citep{onManifoldMod}:} Uses the same basis matrix $E(x, x_o)$ as the normal variant, i.e.,
\begin{equation}
d_\text{onM}(x, x_o) = n(x, x_o) = \frac{\nabla_{x} h(x, x_o)}{\|\nabla_{x} h(x, x_o)\|_2}.
\end{equation}
However, unlike the other Mod-DS approaches that achieve safe control using a single uniform diagonal scaling policy as in Eq.~\eqref{eq:standard mod lambda}, on-manifold Mod-DS realizes \textbf{undesirable-equilibrium-free} control through the combined effect of three navigation policies replacing the diagonal scaling matrix: 
\begin{equation}
\nonumber
D(x, x_o) = D_\nabla(x, x_o) + D_\phi(x, x_o) + \varphi_h(x, x_o)D_h(x, x_o).
\end{equation}
$D_\nabla$ encodes the gradient ascent dynamics from the boundary, $D_\phi$ enables on-manifold morphing dynamics, and $D_h$ defines the regular modulation dynamics. The technical details of on-manifold Mod-DS can be found in \cite{onManifoldMod}. 

In this work, we focus on the obstacle exit strategy $\phi(x, \{x_o\})$ used in $D_\phi$ to efficiently circumnavigate both convex and concave unsafe regions. Unlike $H(x, x_o)$, which is constructed from the boundary function of a single obstacle $o$, $H(x, \{x_o\})$ in on-Manifold Mod-DS is obtained from the combined boundary function in Eq.~\eqref{eq:uni h} by using its gradient as the hyperplane normal direction. Given an initial direction $e^{(0)} \in \mathbb{R}^{d-1}$ satisfying $e^{(0)} \notin \mathcal{N}(H(x, \{x_o\}))$, geodesic approximation generates a path $[x^{(0)}, x^{(1)}, \dots, x^{(N)}]$ along the corresponding isosurface around the surrounding obstacles, where $N \in \mathbb{N}$ is the planning horizon \citep{onManifoldMod}. The tangent-space exit direction $\phi(x, \{x_o\})$ is selected from $m \ge 2^{d-1}$ candidate initial directions according to the resulting potential $P^{(N)}$:
\begin{equation}
\label{eq:geo approxi}
\begin{aligned}
x^{(i+1)} &= x^{(i)} + \beta H(x^{(i)},\{x_o\})H(x^{(i)},\{x_o\})^\top e^{(i)},\\
e^{(i+1)} &=
\frac{H(x^{(i)},\{x_o\})H(x^{(i)},\{x_o\})^\top e^{(i)}}
{\|H(x^{(i)},\{x_o\})H(x^{(i)},\{x_o\})^\top e^{(i)}\|_2},\\
P^{(i+1)} &= P^{(i)} + \beta p(x^{(i+1)}, x^*).
\end{aligned}
\end{equation}
\begin{equation}
\label{eq:phi geo}
\phi(x,\{x_o\}) =
\underset{
e^{(0)}\in\{e^{(0)}_1,\ldots,e^{(0)}_m\}
}{\arg\min}
\;
P^{(N)}\!\left(e^{(0)},x,\{x_o\}\right).
\end{equation}
Here, $\beta$ is the step size, and $p(x,x^*)$ is a user-defined penalty function, often chosen as the distance from $x$ to the target $x^*$.

\begin{figure*}[tbp]
    \centering
    \includegraphics[width=\textwidth]{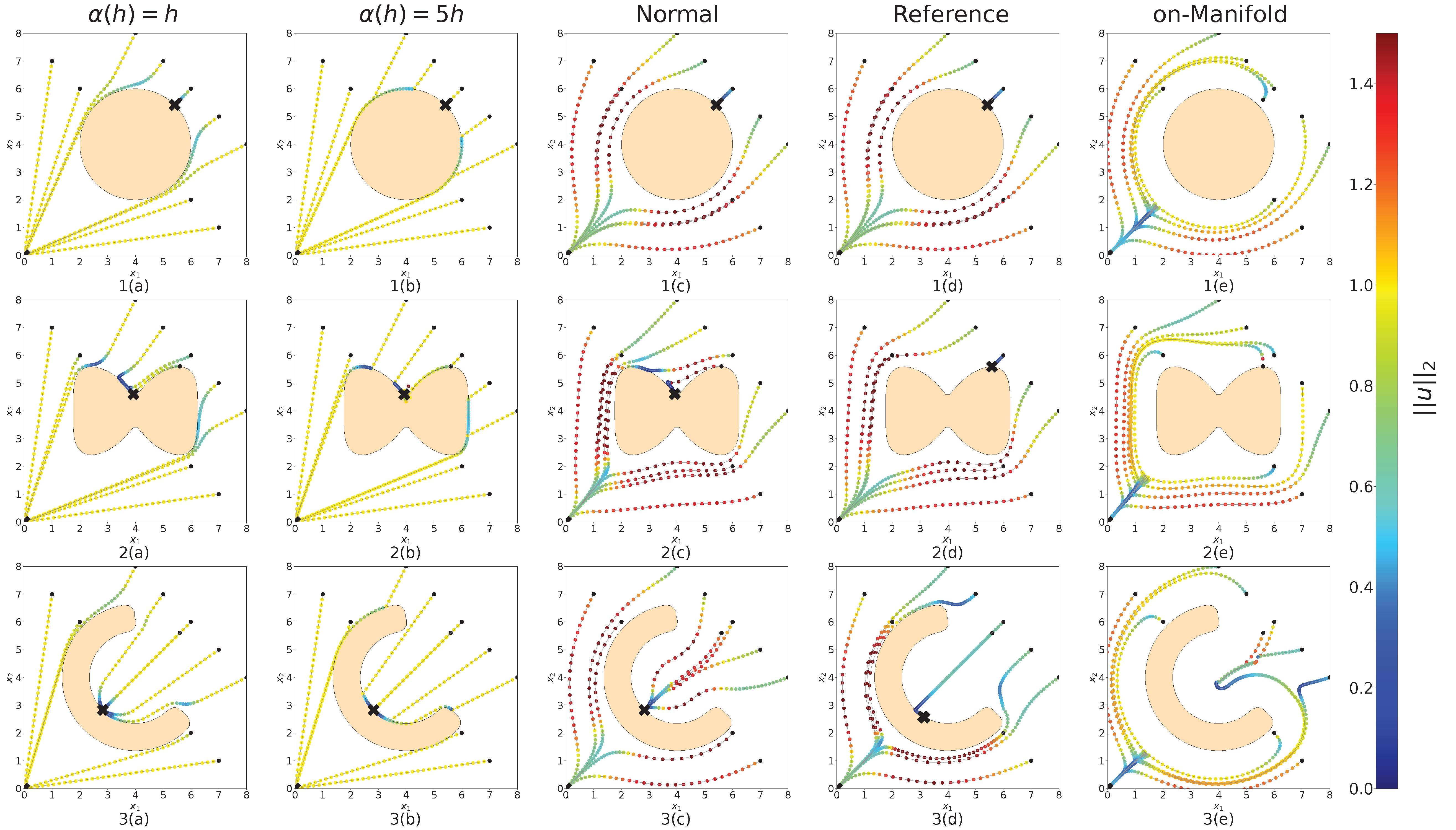}
    \vspace{-10pt}
     \caption{Trajectories from 10 initial locations toward a target at the origin generated by CBF-QP (columns a-b) and Mod-DS (column c-e) obstacle avoidance methods around 1) a convex, 2) a star-shaped, and 3) a non-star-shaped obstacle. The nominal controller is $\dot{x}_{\text{nom}}  =  u_{\text{nom}} = -\frac{1}{\|x\|_2}x $. Colors along the trajectories indicate the robot speed magnitude at each state \(x\).\label{fig:comparison data}}
\end{figure*}

\section{Theoretical Connections between Mod-DS and CBF-QP in Fully Actuated Systems}
\label{sec:theoretical-analysis}
To fairly compare Mod-DS and CBF-QP, we restrict the comparison to fully actuated systems as defined in Eqs.~\eqref{eq:fully actuated system} and~\eqref{eq:nominal DS}, since the Mod-DS baselines are developed for single-integrator dynamics and do not admit a control-affine formulation. \autoref{fig:comparison data} characterizes the behaviors of the Mod-DS variants and CBF-QP approaches for fully actuated systems in a single obstacle environment. In addition to demonstrating the ability of the reference and on-manifold Mod-DS variants to mitigate undesirable equilibria in the presence of concave obstacles, the comparison reveals shared characteristics between the standard Mod-DS and CBF-QP approaches. 
Specifically, given the same input state $x$ and nominal controller $u_\text{nom}$, both methods exhibit undesirable equilibria caused by the collinearity between $u_\text{nom}$ and $d(x, x_o)$. Building on this empirical observation, in \autoref{sec:cbf mod equivalance} we theoretically prove the existence of such undesired on-boundary equilibrium for CBF-QP, stemming from the same pathological case as in Mod-DS; i.e., collinearity between $u_{\text{nom}}$ and $\nabla_x h(x)$. We further demonstrate that normal Mod-DS is mathematically equivalent to CBF-QP under specific choices of $\lambda$ and $\lambda_e$ satisfying Eq.~\eqref{eq:standard mod lambda}. Leveraging this equivalence, in \autoref{sec: reference mod-ds vs cbf} we prove that reference Mod-DS satisfies the CBF-QP conditions in Eq.~\eqref{eq:CBF conditions} and derive a conversion relation with CBF-QP, revealing the closed-form structure that enables robust navigation in star-shaped obstacle environments with only one single undesirable equilibrium.

All proofs assume static obstacles for mathematical interpretability and simplicity, though results are extendable to dynamic obstacle scenarios. For clarity, we simplify notations and let the boundary function $h(x, x_o)$, direction $d(x, x_o)$, normal vector $n(x, x_o)$, and reference $r(x, x_o)$ vectors to $h(x)$, $d(x)$, $n(x)$, and $r(x)$, respectively.

\subsection{Normal Mod-DS and CBF-QP Equivalence}
\label{sec:cbf mod equivalance}

Both normal and reference Mod-DS can exhibit undesirable equilibria when the nominal controller is inversely collinear with their respective directional vectors $d(x)$, defined as the obstacle normal $n(x)$ for normal Mod-DS (Eq.~\eqref{eq:standard mod d}) and the reference vector $r(x)$ for reference Mod-DS (Eq.~\eqref{eq:reference mod d}). Specifically, inverse collinearity,
\begin{equation}
\left\langle d(x),\frac{u_{\mathrm{nom}}}{\|u_{\mathrm{nom}}\|}\right\rangle=-1,
\end{equation}
causes the modulated velocity to vanish as the agent approaches the obstacle boundary, resulting in an undesirable equilibrium \citep{khansari2012dynamical}. While undesirable equilibria have been studied in CLF-CBF-QPs, where conflicts between CLF and CBF constraints can induce boundary equilibria even for convex obstacles \citep{clfcbfEquilibria,tracking_cbf}, the corresponding mechanism in standard CBF-QPs has received less attention. For fully actuated systems, a standard CBF-QP can exhibit the same inverse-collinearity mechanism as normal Mod-DS.

\begin{theorem}[\textbf{Equilibria in CBF-QP}]
\label{theorem:saddle-cbfqp}
Consider a fully actuated CBF-QP in a \emph{static single-obstacle environment}. If, starting from any $x_{\mathrm{ini}}\in\mathcal{C}$, the controller continuously encounters inverse collinearity between $\nabla_x h(x)$ and $u_{\mathrm{nom}}$, i.e.,
\begin{equation}
\left\langle n(x),\frac{u_{\mathrm{nom}}}{\|u_{\mathrm{nom}}\|}\right\rangle=-1,
\end{equation}
then the solution of Eq.~\eqref{eq:cbf-qp fully actuated} converges to an undesirable equilibrium at the obstacle boundary:
\begin{equation}
\lim_{h(x)\rightarrow 0}\dot{x}
=
\lim_{h(x)\rightarrow 0}u_{\mathrm{cbf}}
=
\mathbf{0}.
\end{equation}
\textbf{Proof:} See Appendix~\ref{app:proof-equilibria-cbfqp} \hfill $\blacksquare$
\end{theorem}

The proof shows that inverse collinearity causes the CBF-QP solution to remain directed toward the obstacle while its magnitude vanishes as $h(x)\rightarrow0$, trapping the agent at the boundary.
The similarity between normal Mod-DS and fully actuated CBF-QP in terms
of undesirable equilibria originates from their equivalence in static
environments.

\begin{theorem}[\textbf{Normal Mod-DS and CBF-QP Equivalence}]
\label{theorem: equivalance}
In static environments, given a fully actuated CBF-QP controller with
any extended $K_{\infty}$ function $\alpha$, there exists a pair of
$\lambda$ and $\lambda_e$ defined as, 
\begin{equation}
\lambda=
\begin{cases}
1, & \nabla_xh^\top u_\text{nom}\geq-\alpha,\\[4pt]
-\dfrac{\alpha}{\nabla_xh^\top u_\text{nom}}, & \nabla_xh^\top u_\text{nom}<-\alpha,
\end{cases}
\qquad
\lambda_e=1.
\end{equation}
such that the normal Mod-DS defined in
Eqs.~\eqref{eq:ds-modulation}--\eqref{eq:standard mod d} is equivalent to the CBF-QP; i.e,. $u_\text{mod}^N-u_\text{cbf} = 0$.\\
\textbf{Proof:} See Appendix~\ref{app:proof-mod-ds-cbf-equivalence} \hfill $\blacksquare$
\end{theorem}

\subsection{Reference Mod-DS and CBF-QP Similarities}
\label{sec: reference mod-ds vs cbf}

In \autoref{fig:comparison data}, the reference Mod-DS was shown to outperform the CBF-QP in star-shaped obstacle avoidance using fully actuated systems. Nevertheless, the fact that a linear DS modified by reference Mod-DS exhibits the same undesirable equilibrium as the CBF-QP in circular obstacle avoidance suggests an underlying connection between these seemingly distinct approaches. Characterizing their similarities and differences is important for understanding how their respective safety and motion-shaping mechanisms affect navigation liveness and goal convergence. We therefore first show that reference Mod-DS satisfies the CBF condition in the clean mathematical setting of static single-obstacle environments and then quantify its difference from the CBF-QP. These results provide a theoretical basis for combining their complementary properties in the safe and liveness-oriented control approach developed in \autoref{sec:mod-cbf}. 


\begin{theorem}[\textbf{Safe Set Invariance in Reference Mod-DS}]
\label{theorem:set invariance in reference}
Given fully actuated nominal controller $u_\text{nom}$ in Eq.~\eqref{eq:fully actuated system} and any $\lambda$ and $\lambda_e$ satisfying Eq.~\eqref{eq:standard mod lambda}, there exists an extended class $K_{\infty}$ function such that the Mod-DS safe controller $u_\text{mod}$ defined in Eqs.~\eqref{eq:ds-modulation}, \eqref{eq:M matrix}, and \eqref{eq:reference mod d} satisfies the CBF condition in Eq.~\eqref{eq:cbf-qp fully actuated} when $\dot{x}_o=\mathbf{0}$.\\
\textbf{Proof:} See Appendix~\ref{app:proof-set-invariance-reference} \hfill $\blacksquare$
\end{theorem}

\begin{theorem}[\textbf{Quantitative Difference between Reference Mod-DS and CBF-QP}]
\label{theorem:quantitative diff reference}
Consider a fully actuated nominal linear DS $u_\text{nom}=k(x^*-x)$ with $k>0$, as required by the reference Mod-DS formulation. When the CBF constraint is active, the difference between the outputs of reference Mod-DS and CBF-QP using that same nominal controller is
\begin{equation}
u_\text{mod}^R-u_\text{cbf}=-\lambda^+HH^\top\left(I-\dfrac{rn^\top}{n^\top r}\right)(x^*-x),
\end{equation}
where $\lambda^+>0$ under the standard reference Mod-DS parameterization with $\lambda_e>\lambda$, $r=r(x,x_o)$, $H=H(x,x_o)$, and $n=n(x,x_o)$ are defined in Eqs.~\eqref{eq:reference mod d}, \eqref{eq:H matrix}, and \eqref{eq:standard mod d}, respectively.\\
\textbf{Proof:} See Appendix~\ref{app:proof-quantitative-difference} \hfill $\blacksquare$
\end{theorem}

\section{Modulated Control Barrier Functions (MCBF)}
\label{sec:mod-cbf}
Mod-DS methods provide effective obstacle avoidance for fully actuated systems, with fewer undesirable equilibria and dysfunctional circulation than conventional CBF-QPs in non-convex environments. CBF-QPs, however, offer broader applicability to control-affine systems and explicit enforcement of actuation constraints. To bridge this gap, we propose the \textbf{Modulated CBF-QP (MCBF-QP)} safety filter with two variants: (i) a reference-modulated formulation, described in Section~\ref{sec:reference mod-cbf}, for efficient obstacle avoidance in star-shaped environments; and (ii) an on-manifold formulation, described in Section~\ref{sec:on-manifold cbf-qp}, for obstacle avoidance in arbitrarily shaped concave environments. 

For both variants, we establish theoretical guarantees of undesirable-equilibrium reduction or elimination and asymptotic goal convergence without explicitly designing or imposing a global Control Lyapunov Function (CLF). Instead, \textit{liveness} is enhanced by incorporating the Mod-DS geometric principles of reference-direction and tangent-hyperplane modulation into the CBF-QP formulation. Specifically, Theorems~\ref{theorem:r-mcbf-convergence} and~\ref{theorem:om-mcbf-convergence} show that the resulting closed-loop systems admit a hybrid Lyapunov function $V_H$ that strictly decreases away from the goal over the relevant safe configuration space.
To extend these geometric principles to general control-affine systems, we formulate the geometric modulation concepts of Mod-DS directly in the system's admissible control space. This enables application to underactuated and nonholonomic systems, e.g., a unicycle with non-Euclidean state $x=[p_x,p_y,\theta]^\top\in\mathbb{R}^2\times\mathbb{S}^1$.

\subsection{Reference MCBF for Star-shaped Obstacle Avoidance}\label{sec:reference mod-cbf}
In \autoref{theorem:set invariance in reference}, reference Mod-DS and CBF-QP are shown to exhibit equivalent behavior along the $n(x,x_o)$ direction. \autoref{theorem:quantitative diff reference} further shows that, for fully actuated systems with a linear DS as the nominal input, the principal difference between the two methods lies in the off-diagonal components within the tangent hyperplane. These components are key to reducing undesirable equilibria for star-shaped obstacle avoidance. This motivates augmenting the CBF-QP with constraints on $u_{\mathrm{cbf}}$ \textcolor{magenta}{along the tangent hyperplane}. Among possible formulations, we introduce the following linear-constraint formulation, termed reference MCBF-QP (\textbf{R-MCBF-QP}):
\begin{gather}
\nonumber u_{\text{mcbf}} = \argmin_{u \in \mathbb{R}^p}(u-u_\text{nom})^\top(u-u_\text{nom})\\ 
\nonumber L_fh(x,x_o) + L_gh(x,x_o)u + \nabla_{x_o} h(x,x_o)^\top\dot{x}_o\geq \\
\qquad \qquad \qquad \qquad -\alpha (h(x,x_o)) \quad \forall o \in \mathcal{O},\label{eq:mod-r-cbf-affine}\\
\nonumber \textcolor{magenta}{H_o(x,x_o)^\top\left(I - \frac{r_o(x,x_o)\nabla_xh_o(x,x_o)^\top}{\nabla_xh_o(x,x_o)^\top r_o(x,x_o)}\right)} \\
\nonumber \textcolor{magenta}{\qquad  \cdot [f(x)+g(x)u -d_g(x,x^*, u_\text{nom})]= \textbf{0} \quad \forall o \in \mathcal{O}_r,}
\end{gather}
where
\begin{align}
\label{eq:dg}
d_g(x,x^*, u_\text{nom}) = \frac{\|f(x)+g(x)u_\text{nom}\|}{\|x^*-x\|}(x^*-x),
\end{align}
and
\begin{gather}
\nonumber
\mathcal{O}_r
=
\Big\{
o\in\mathcal O
\;\Big|\;
L_f h_o(x,\{x_o\})
+
L_g h_o(x,\{x_o\})u_\text{nom}\\
+
\nabla_{x_o}h_o(x,\{x_o\})^\top\dot{x}_o
<
-\alpha(h_o(x,\{x_o\}))
\Big\}.
\label{eq:o_r}
\end{gather}
Here, $g(x)$ is defined in Eq.~\eqref{eq:affine system}, and $H_o(x,x_o)$ and $r_o(x,x_o)$ denote the tangent hyperplane and reference direction of obstacle $o$, respectively, extended to the general control-affine robot state space. Note that the term
\begin{equation*}
\resizebox{.98\hsize}{!}{$E_r(x,x_o)^{-1}[2:d,:]\dot{x} = H(x,x_o)^\top \Big(I - \frac{r(x, x_o)\nabla_x h(x,x_o)^\top}{\nabla_x h(x,x_o)^\top r(x, x_o)} \Big)\dot{x}$}
\end{equation*}
represents the projection of $\dot{x} = f(x) + g(x)u$ onto the hyperplane $H(x,x_o)$ in the generalized non-orthogonal basis $E_r(x,x_o)$. Multiplying by $H(x,x_o)$ recovers the operator $HH^\top(I-rn^\top/(n^\top r))$ in Eq.~\eqref{eq:quantitative diff}, establishing that the \textcolor{magenta}{magenta-highlighted} affine constraints in Eq.~\eqref{eq:mod-r-cbf-affine} enforce the same hyperplane-projection patterns as the reference Mod-DS. As shown later, this eliminates undesirable equilibria except when $d_g(x,x^*,u_\text{nom})$ is inversely collinear with $r_o(x,x_o)$. Since these constraints are linear equalities in $u$, they preserve QP convexity with only modest computational overhead.

\begin{figure}[!t]
    \centering
     \includegraphics[width=\linewidth]{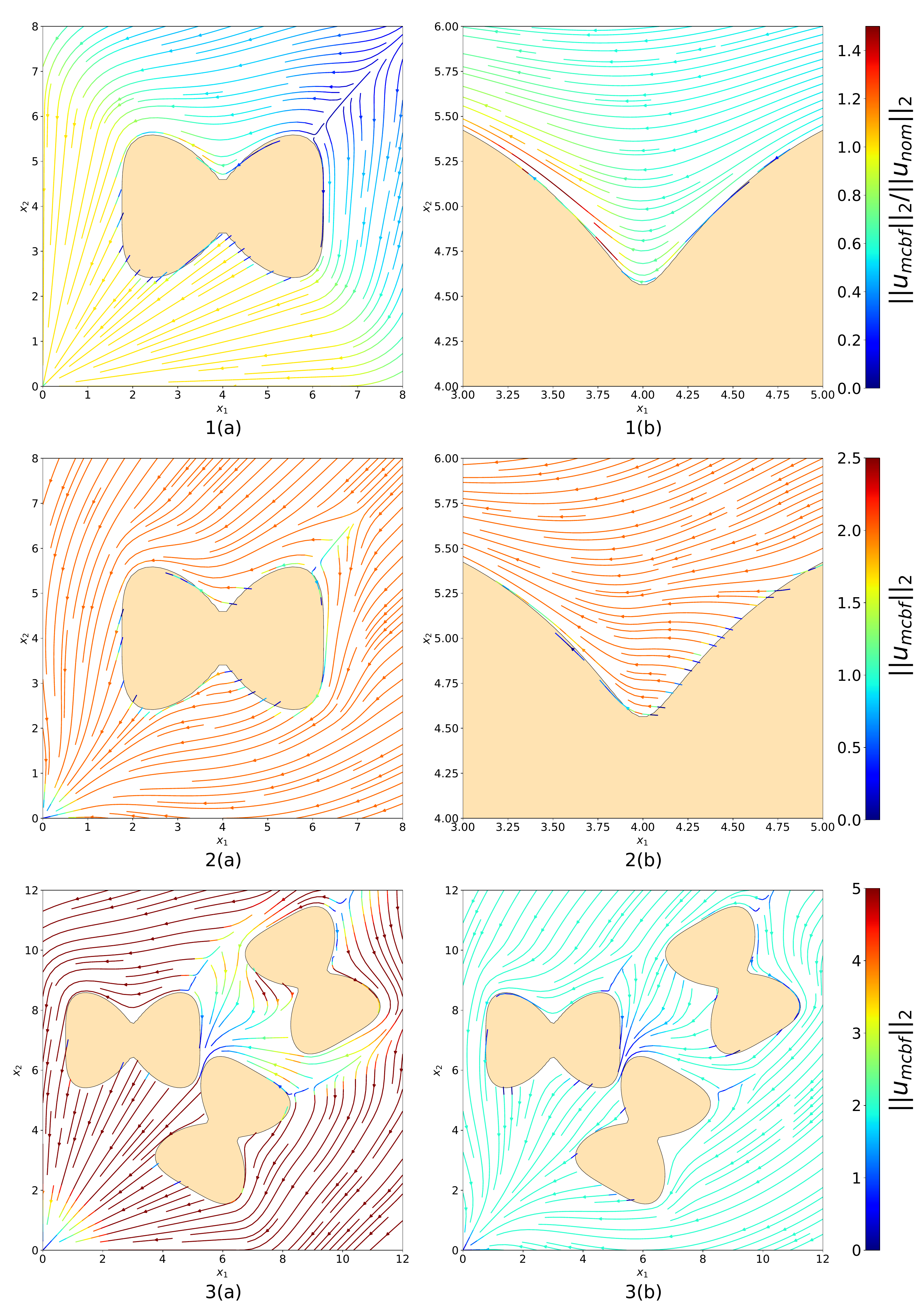}
    \vspace{-20pt}
     \caption{Performance of reference MCBF-QP in single obstacle avoidance with no robot input constraints (first row, (a): full view, (b): close view), in single obstacle avoidance with robot input constraints of $\|u_\text{mcbf}\|_2\leq2$ (second row, (a): full view, (b): close view), and in multi-obstacle avoidance respectively with and without robot input constraints of $\|u_\text{mcbf}\|_2\leq2$ (third row (a),(b)).\label{fig:mod r cbf}}
     \vspace{-10pt}
\end{figure}

The introduction of additional equality constraints, however, can significantly shrink the feasible set of $u$ and may render the optimization problem infeasible under tight actuation limits or when the robot is squeezed between multiple obstacles. Therefore, in practice, we instead employ the following relaxed formulation, which preserves the feasibility properties and computational efficiency of the original CBF-QP:
\begin{gather}\label{eq:mod-r-cbf-affine relaxed} 
\nonumber u_{\text{mcbf}} = \argmin_{u \in \mathbb{R}^p,\{\rho_o \in \mathbb{R}^{d-1}\}}(u-u_\text{nom})^\top(u-u_\text{nom})+\sum_{o\in \mathcal{O}}\rho_o^\top\rho_o\\ 
\nonumber L_fh(x,x_o) + L_gh(x,x_o)u + \nabla_{x_o} h(x,x_o)^\top\dot{x}_o\geq \\
\qquad \qquad \qquad \qquad -\alpha (h(x,x_o)) \quad \forall o \in \mathcal{O},\\
\nonumber \textcolor{magenta}{H_o(x,x_o)^\top\left(I - \frac{r_o(x,x_o)\nabla_xh_o(x,x_o)^\top}{\nabla_xh_o(x,x_o)^\top r_o(x,x_o)}\right)} \\
\nonumber \textcolor{magenta}{\qquad \qquad \qquad \cdot  [g(x)u -d_g(x,x^*, u_\text{nom})]= \rho_o \quad \forall o \in \mathcal{O}_r,}
\end{gather}
where $\rho_o$ is a relaxation parameter, $c_p \in \mathbb{R}^+$ is a user-defined penalty coefficient. Since $\rho_o$ is unconstrained, the relaxed projection constraints can always be satisfied for any admissible control input $u$. Therefore, \textbf{the relaxed R-MCBF-QP is feasible if and only if the underlying CBF-QP without the \textcolor{magenta}{additional projection constraints} is feasible}. Furthermore, as $c_p$ increases, violations of the projection constraints become increasingly penalized. Consequently, whenever the hard-constrained R-MCBF-QP in Eq.~\eqref{eq:mod-r-cbf-affine} is feasible, the solution of the relaxed problem in Eq.~\eqref{eq:mod-r-cbf-affine relaxed} converges to that of the hard-constrained formulation as $c_p\rightarrow\infty$. The performance of relaxed R-MCBF-QP in star-shaped obstacle environments, using the robot system defined in Eq.~\eqref{eq:fully actuated system} and the linear nominal controller $u_{\text{nom}} = -2x$, is validated in \autoref{fig:mod r cbf}.

\subsubsection{Theoretical Guarantees}
In this section, we establish theoretical guarantees for the reference MCBF-QP in a single-obstacle environment. Specifically, Theorem~\ref{theorem:r-mcbf-equilibrium} establishes that the reference MCBF-QP eliminates undesirable equilibria except for inverse-collinear configurations, while Theorem~\ref{theorem:r-mcbf-convergence} establishes asymptotic convergence to the goal for star-shaped obstacles under the stated assumptions. Since undesirable equilibria and dysfunctional circulation concern the robot's motion relative to the obstacle, we conduct the analysis in the obstacle-fixed frame, in which the obstacle is stationary and $\dot{x}_o=\mathbf{0}$, as is common practice in prior works. Detailed proofs and supporting lemmas are provided in Appendix~\ref{app:proofs rmcbf}.

\begin{theorem}[\textbf{Equilibrium-Reduction Guarantees of R-MCBF-QP}]
\label{theorem:r-mcbf-equilibrium}
Consider the R-MCBF-QP defined in Eq.~\eqref{eq:mod-r-cbf-affine} for a general control-affine system operating in an environment containing a star-shaped obstacle. Suppose Assumptions~\ref{assumption:velocity-bounds}, \ref{assumption:nominal-convergence}, and \ref{assumption:nominal-cone-non-increase}, and the feasibility and control-authority conditions of the R-MCBF-QP are satisfied. Then any undesirable equilibrium of the resulting closed-loop system can occur only at a state $x$ for which the goal direction $x^*-x$ is inversely collinear with the reference direction $r(x,x_o)$, i.e.,
\begin{equation}
x^*-x=c^{-} r(x,x_o)
\end{equation}
for any $c^{-}<0$.\\
\textbf{Proof:} Full derivation in Appendix~\ref{app:proof-r-mcbf-equilibrium} \hfill $\blacksquare$
\end{theorem}

To establish convergence to the goal, we impose the following condition on the nominal controller.

\begin{assumption}[\textbf{Nominal Cone Non-Reversal}]
\label{assumption:nominal-cone-non-increase}
The nominal controller does not reverse the robot's angular progress toward the goal around the obstacle.
Specifically, the magnitude of the reference angle $\theta$ is nonincreasing under the nominal dynamics:
\begin{equation}
\label{eq:nominal-cone-non-increase}
\frac{d}{dt}|\theta(x)|
\bigg|_{\dot{x}=\dot{x}_{\mathrm{nom}}}
\leq 0,
\qquad
\theta(x)\neq0,
\end{equation}
where $\theta(x)$ is the signed reference angle between the reference direction $r(x,x_o)$ and the goal direction $x^*-x$. At the collinear configuration $\theta(x)=0$, the nominal angular rate satisfies
\begin{equation}
\label{eq:nominal-cone-zero}
\dot{\theta}_{\mathrm{nom}}=0.
\end{equation}
\end{assumption}



\begin{theorem}[\textbf{Asymptotic Convergence of R-MCBF-QP}]
\label{theorem:r-mcbf-convergence}
Consider the R-MCBF-QP in a star-shaped-obstacle environment. Under Assumptions~\ref{assumption:velocity-bounds}, \ref{assumption:nominal-convergence}, and \ref{assumption:nominal-cone-non-increase}, and the feasibility and control-authority conditions of the R-MCBF-QP, for any initial condition that is not inverse-collinear, i.e.,
\begin{equation}
|\theta(x(0))|
\leq
\pi-\epsilon_\pi,
\qquad
\epsilon_\pi>0,
\end{equation}
the closed-loop system asymptotically converges to the goal $x^*$; i.e., $\lim_{t\rightarrow\infty}||x-x^*||=0$.\\
\textbf{Proof:} Full derivation in Appendix~\ref{app:proof-r-mcbf-convergence} \hfill $\blacksquare$
\end{theorem}

\subsection{On-Manifold MCBF for Non-star-shaped Obstacle Avoidance}
\label{sec:on-manifold cbf-qp}
By retaining the standard CBF-QP constraints to capture Mod-DS behavior along the normal directions $n_o(x,x_o)$ and introducing the new constraints in Eq.~\eqref{eq:mod-r-cbf-affine} to reproduce the reference Mod-DS actions within the tangent hyperplanes $H_o(x,x_o)$, the proposed R-MCBF-QP combines the constraint-handling capability of CBF-QPs with the liveness properties of reference Mod-DS for star-shaped obstacles. The R-MCBF-QP formulation, however, inherits several limitations of reference Mod-DS. In particular, the convergence guarantee of reference Mod-DS is established for linear nominal DS only \citep{LukesDS}. Although the R-MCBF-QP formulation extends the approach to more general nominal controllers, the corresponding \textit{liveness} guarantee still requires the nominal controller to satisfy Assumptions~\ref{assumption:nominal-cone-non-increase}. Moreover, the reference direction $r(x,x_o)$ becomes increasingly difficult to construct in higher-D state spaces, limiting the scalability of R-MCBF-QP.

To develop a more general liveness-enhanced CBF-QP, we instead draw on on-manifold Mod-DS, which supports arbitrary obstacle geometries and general goal-converging nominal controllers. Although on-manifold Mod-DS' combined geometric construction is too intricate to admit a direct theoretical correspondence with CBF-QP, its key mechanism remains compatible with the proposed framework. As it regulates $\dot{x}$ along the normal direction $n(x,x_o)$ while guiding tangent-space motion using $\phi(x,x_o)$ to avoid undesirable equilibria and dysfunctional circulation. 

Here, $\phi(x,x_o)$ denotes a tangent-space direction that guides the robot out of concave or near-boundary regions and is obtained by evaluating the geometric approximation in Eq.~\eqref{eq:geo approxi} and selecting the minimum-potential direction according to Eq.~\eqref{eq:phi geo}.


As discussed next, the underlying geodesic approximation mechanism can be generalized beyond Euclidean position spaces to provide guidance in general robot state spaces $x$, such as the configuration space of a unicycle model $x \in \mathbb{R}^2\times\mathbb{S}_1$. Motivated by these properties, we incorporate the $\phi(x,\{x_o\})$-induced tangent-space constraint into the CBF-QP formulation, yielding the on-manifold MCBF-QP (OnM-MCBF-QP) presented in \autoref{sec: onm formulation}.

\subsubsection{Geodesic Approximation in Non-Holonomic Space}
\label{sec:geodesic}

\begin{figure}[!tbp]
    \centering
    \subfloat[]{%
        \includegraphics[width=0.49\linewidth]{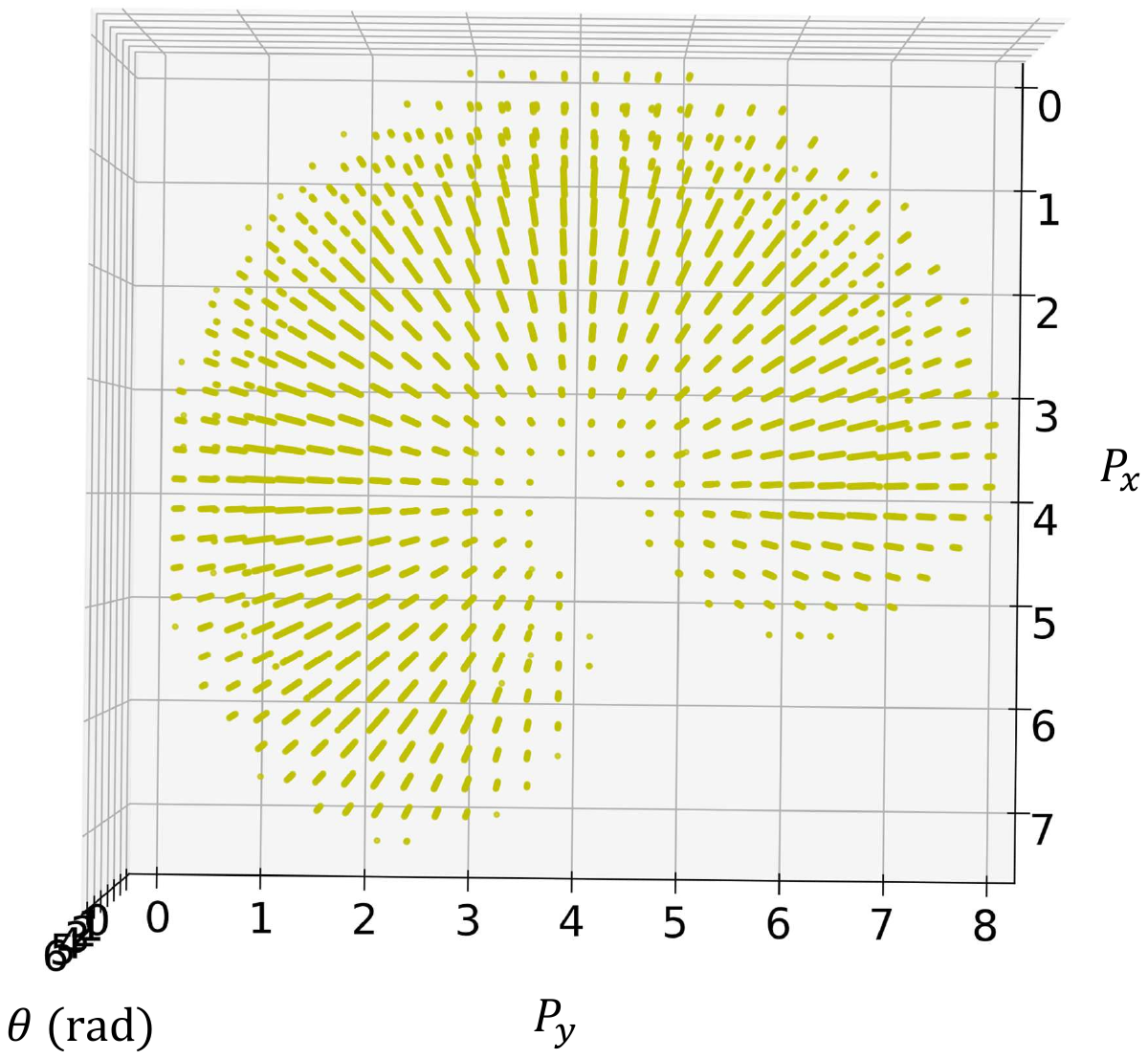}%
        \label{fig: p top}%
    }
    \subfloat[]{%
        \includegraphics[width=0.49\linewidth]{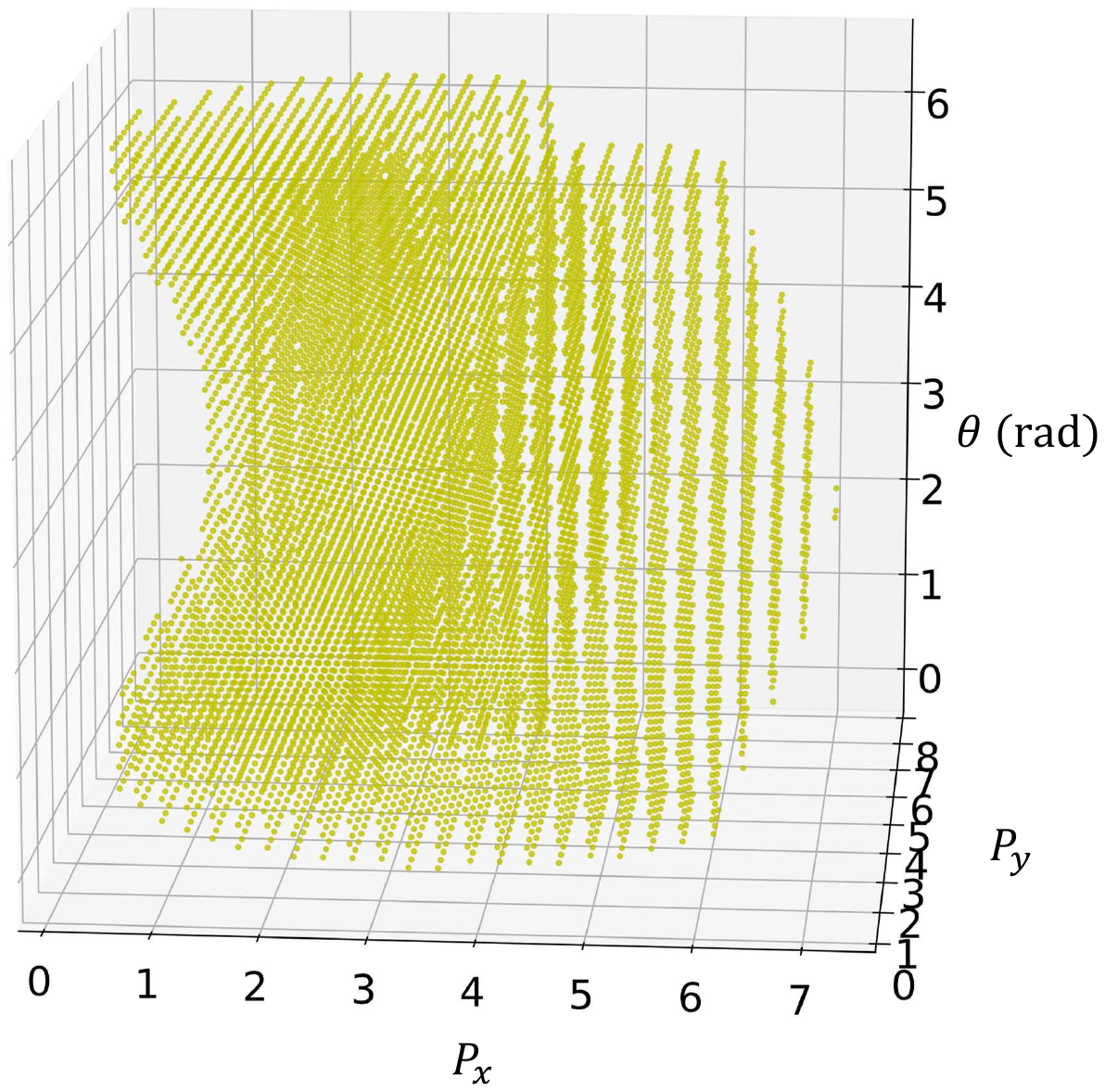}%
        \label{fig: p side}%
    }
    
    \subfloat[]{%
        \includegraphics[width=0.49\linewidth]{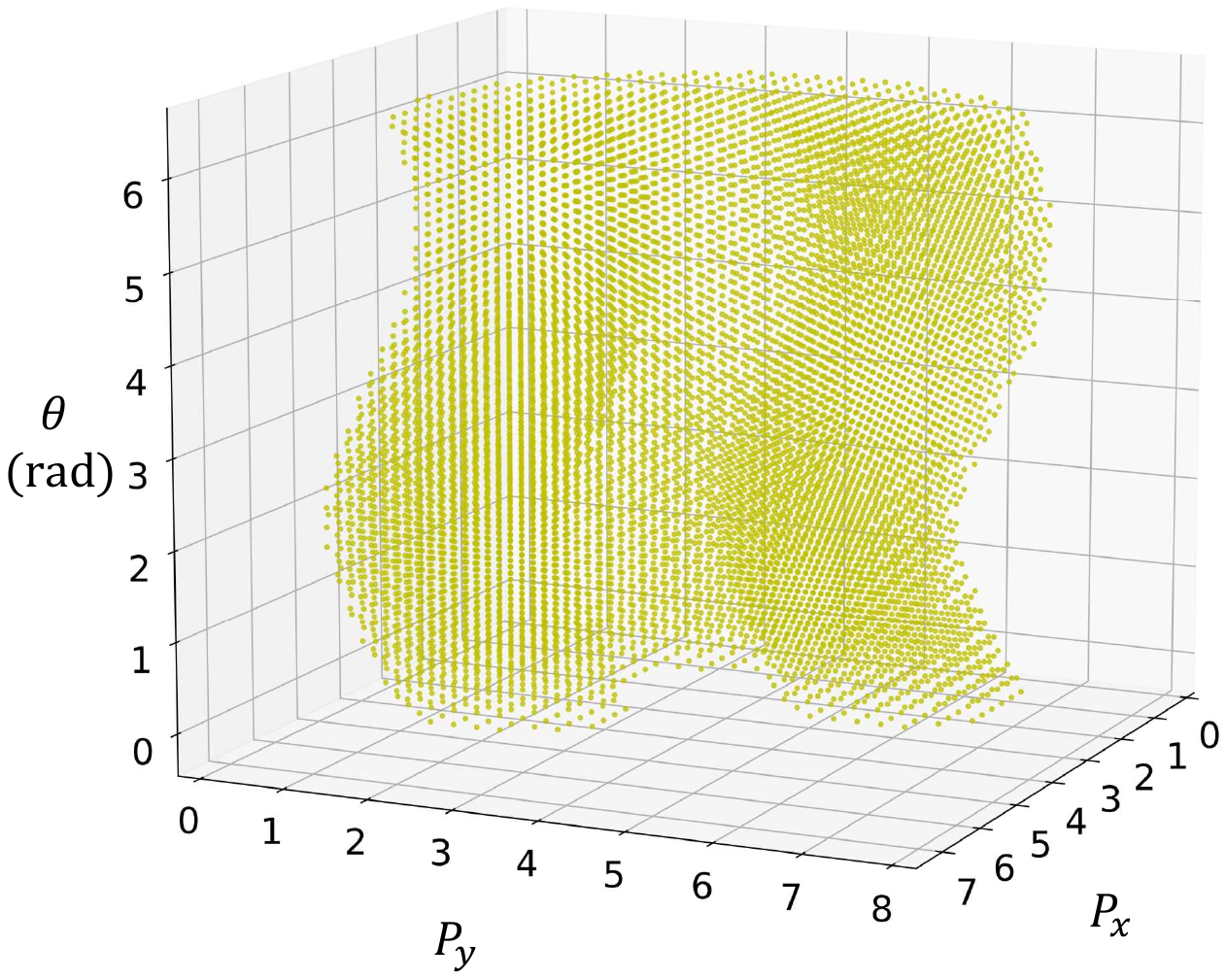}%
        \label{fig: p tri}%
    }
    \subfloat[]{%
        \includegraphics[width=0.49\linewidth]{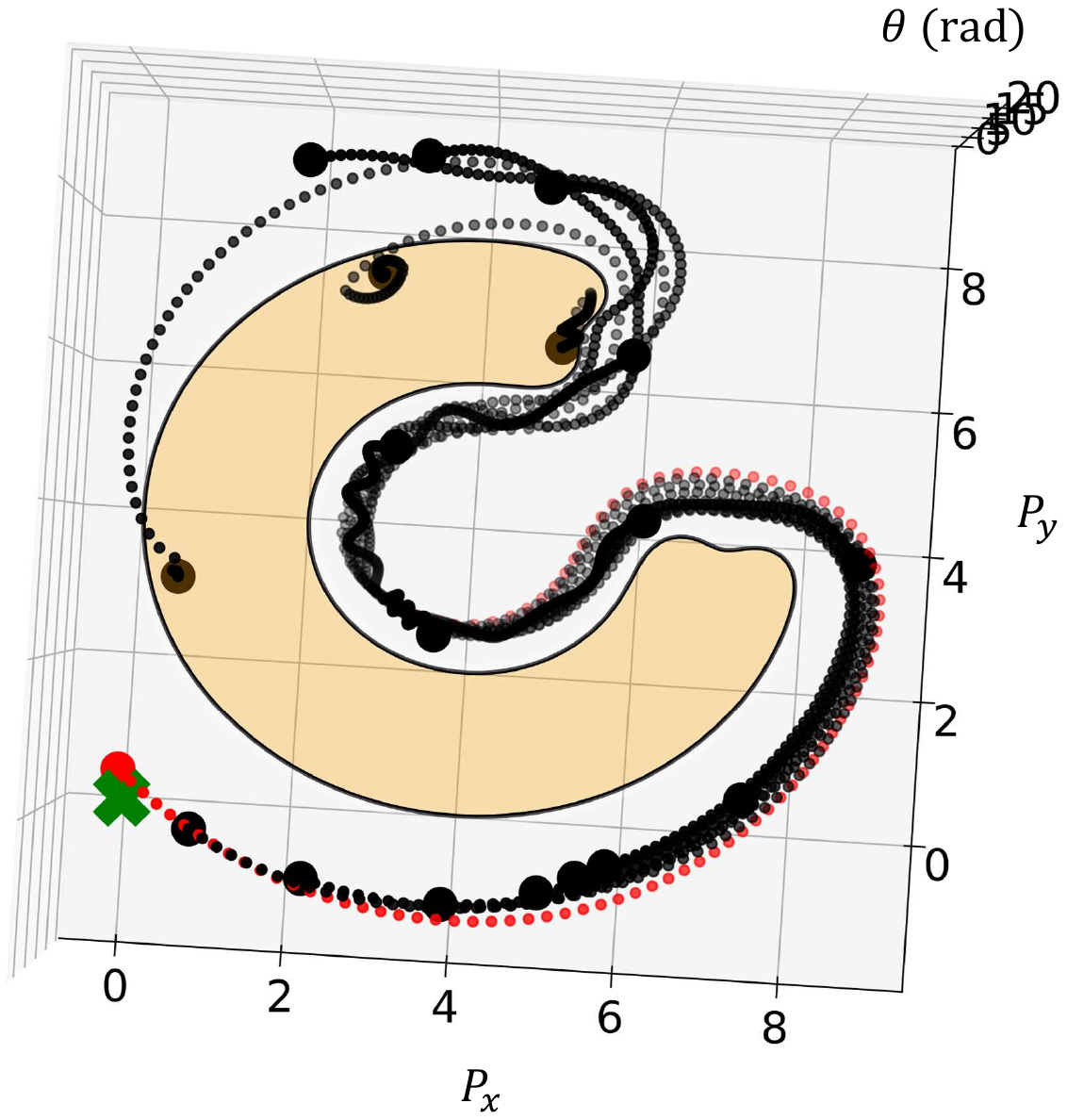}%
        \label{fig: p 20 top}%
    }
    
    \subfloat[]{%
        \includegraphics[width=0.49\linewidth]{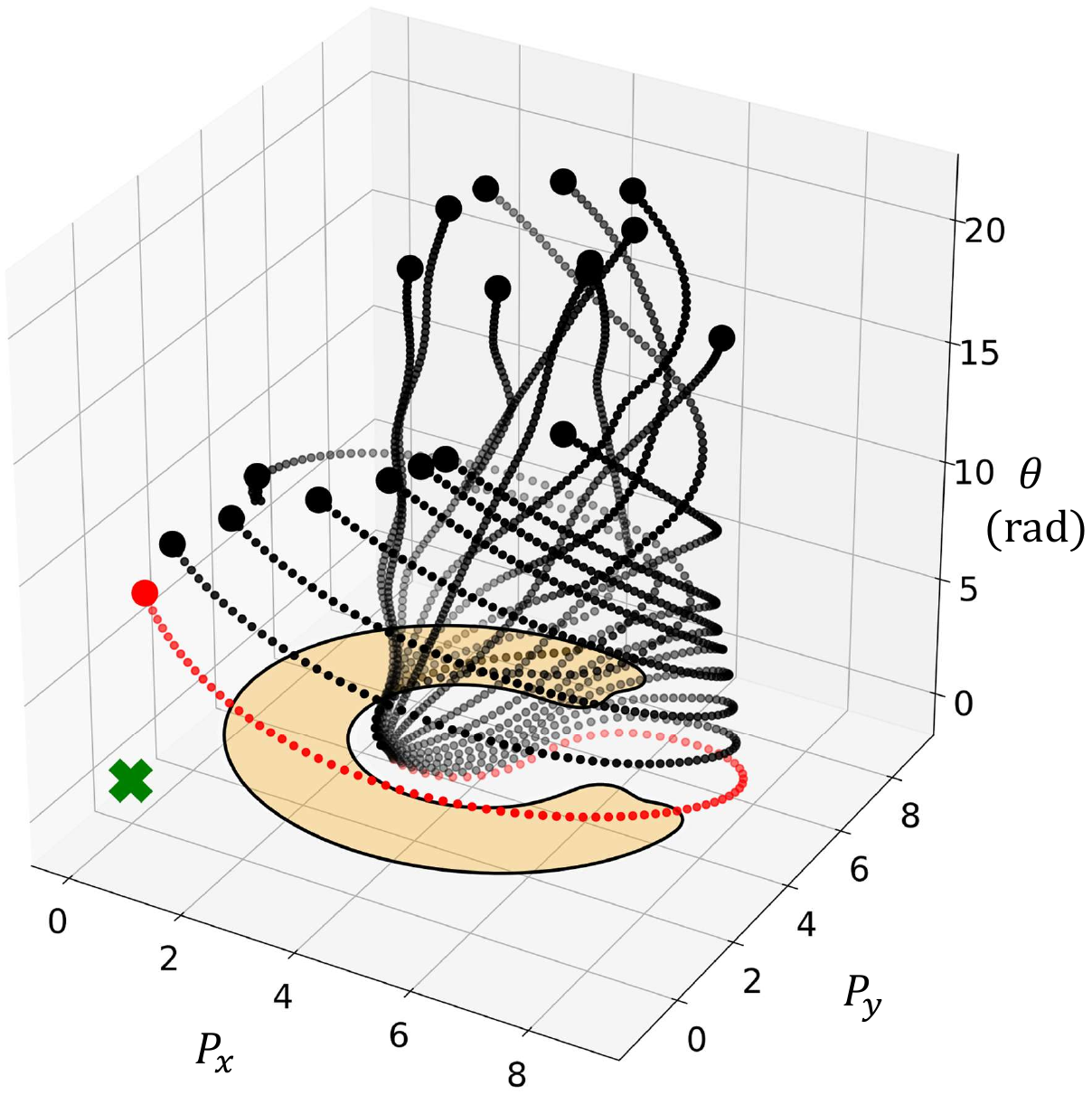}%
        \label{fig: p 20 tri}%
    }
    \subfloat[]{%
        \includegraphics[width=0.49\linewidth]{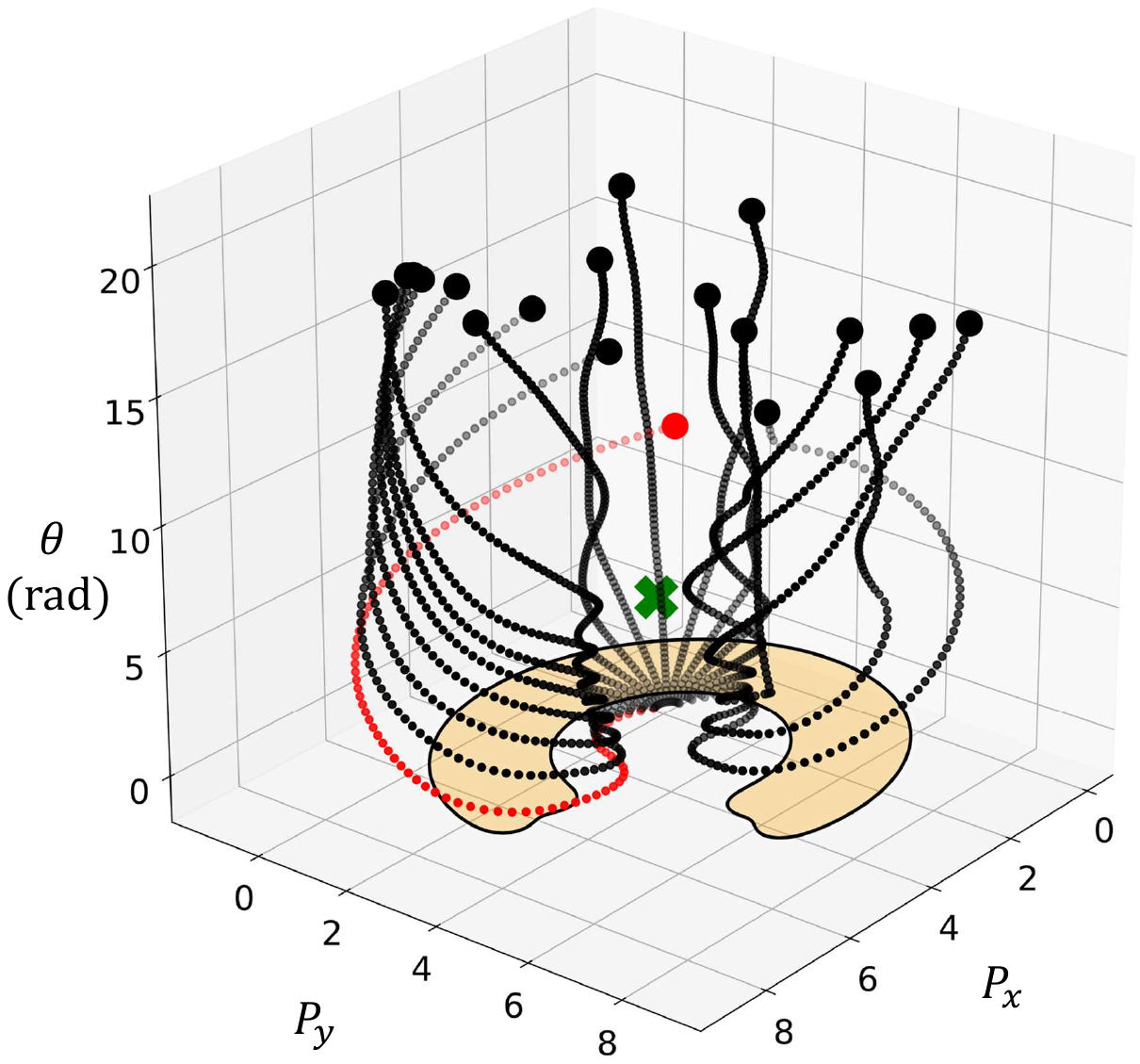}%
        \label{fig: p 20 tri2}%
    }
    \caption{Figures (a) - (c) show the top, side, and diagonal views of the unsafe regions defined by the augmented barrier function $h_\text{aug}$ in Eq.~\eqref{eq:augmented barrier function}. Figures (d) - (f) illustrate the same views with the geodesic approximation method applied: the selected circumventing trajectory is shown in red, leading efficiently toward the target location (green 'X') despite the non-Euclidean nature of the state space. }
    \label{fig:geo non-eucl}
\end{figure}

Empirically, we extend the geodesic approximation method to non-holonomic state space and validate it using the unicycle model $x = [p_x, p_y, \theta]^\top$. \autoref{fig:geo non-eucl}~(a)--(c) illustrate the unsafe region defined by the augmented CBF in Eq.~\eqref{eq:augmented barrier function} within underactuated robot state space. For clarity and interpretability, only the section with $\theta \in [0, 2\pi)$ is shown. Starting from the robot state $x^{(0)}=[3,3,0]$, the figure illustrates 20 vectors $e^{(0)}$ uniformly sampled in the plane tangent to the gradient of $h_\text{aug}(x^{(0)}, x_o)$.

Using the geodesic approximation strategy in Eq.~\eqref{eq:geo approxi}, the tangent directions are evaluated to identify the circumventing trajectory that advances toward the target most efficiently, as quantified by the penalty function $p(x^{(i)}, x^*) = \|x^{(i)}[:2]-x^*\|_2$. In \autoref{fig:geo non-eucl}~(d)--(f), the resulting $\phi(x,\{x_o\})$ from Eq.~\eqref{eq:phi geo} demonstrates that the generalized formulation provides guidance not only for omnidirectional translational motion but also for the robot orientation $\theta$.


Notably, for this extension to remain meaningful, the augmented barrier function must be explicitly defined over the full robot state space rather than purely as a signed distance function. See \autoref{sec: augmented CBF} for further discussion on augmented CBF design.

\subsubsection{On-Manifold MCBF-QP Formulations}
\label{sec: onm formulation}
For general control-affine systems, safe and efficient motion filtering for arbitrary obstacle shapes—free of on-boundary undesirable equilibria and mitigating dysfunctional circulation—can be achieved via the following optimization problem:
\begin{gather}\label{eq:mod-phi-cbf-affine} 
 u_{\text{mcbf}} = \argmin_{u \in \mathbb{R}^p}\;(u-u_\text{nom})^\top(u-u_\text{nom})\\
 L_f\bar{h}(x,\{x_o\}) + L_g\bar{h}(x,\{x_o\})u +
\dot{\bar{h}}_o\geq \label{eq:mod-phi cbf constraint}\\
\nonumber\qquad \qquad \qquad \qquad -\alpha (\bar{h}(x,\{x_o\})),\\
\label{eq:mod-phi phi constraint} \textcolor{magenta}{\phi(x, \{x_o\})^\top\!\left(f(x)+g(x)u\right)
\ge
\nu_{\phi},
\qquad
x\in\bar{\Omega}_{\mathrm{obs}}},
\end{gather}
where
\begin{equation}
\label{eq:bar_h_obs_motion}
\dot{\bar{h}}_o
=
\sum_{o\in\mathcal{O}}
\frac{
\exp\!\left(-\rho[h_o(x,x_o)+1]\right)
}{
\sum_{j\in\mathcal{O}}
\exp\!\left(-\rho[h_j(x,x_j)+1]\right)
}
\nabla_{x_o}h_o(x,x_o)^\top\dot{x}_o .
\end{equation}
denotes the obstacle-motion contribution to the time derivative of
$\bar{h}$ obtained by differentiating Eq.~\eqref{eq:uni h} with respect
to the obstacle states, and
\begin{gather}
\nonumber
\bar{\Omega}_{\mathrm{obs}}
=
\left\{
x\in\mathcal X
\;\middle|\;
L_f\bar{h}(x,\{x_o\})
+
L_g\bar{h}(x,\{x_o\})u_\text{nom}
+
\dot{\bar{h}}_o
\right.
\\
\left.
\qquad\qquad\qquad\qquad
<
-\alpha(\bar{h}(x,\{x_o\}))
\right\}.
\label{eq:o_obs}
\end{gather}
\textcolor{magenta}{$\phi(x,\{x_o\})$} denotes the tangent guiding vector computed via the geodesic approximation in Eq.~\eqref{eq:geo approxi} using the combined augmented barrier function $\bar{h}(x,\{x_o\})$, which is constructed from the individual augmented barrier functions according to Eq.~\eqref{eq:uni h}. The parameter $\nu_{\phi}\in\mathbb{R}^+$ is user-defined. In Eq.~\eqref{eq:mod-phi-cbf-affine}, the \textcolor{magenta}{additional tangent-guidance constraint} is activated only when the direct line segment from the robot to the goal intersects the obstacle set (\(x\in\bar{\Omega}_{\mathrm{obs}}\)). Analogous to the on-manifold Mod-DS approach~\citep{onManifoldMod} described in Section~\ref{sec:prelims-mod}, it enforces a non-zero system velocity along the tangent guiding direction $\phi(x,\{x_o\})$, steering the robot toward an exit from dysfunctional circulation while avoiding convergence to undesirable boundary equilibria. The parameter $\nu_{\phi}$ specifies the minimum allowable velocity component along the tangent guiding direction.

The practical implementation of $\phi(x,\{x_o\})$ follows the geodesic approximation procedure proposed in \cite{onManifoldMod}. For simple concave obstacle geometries, a sufficiently large propagation horizon $N$ and sufficiently small step size $\beta$ are selected, while the propagation is terminated once the propagated state $x^{(i)}$ satisfies
\[
\frac{\nabla_{x^{(i)}}h(x^{(i)},\{x_o\})^\top(x^*-x^{(i)})}{\|\nabla_{x^{(i)}}h(x^{(i)},\{x_o\})\|\,\|x^*-x^{(i)}\|}
\ge \eta,
\]
where $\eta\in(0,1)$ is a user-defined threshold. This criterion identifies a suitable exit point on the obstacle boundary, beyond which the robot can leave the obstacle and continue moving toward the target. For highly concave obstacle geometries or dynamically deforming obstacle clusters, fixed approximation parameters may no longer provide consistent geodesic guidance. A more general geodesic approximation algorithm that automatically updates the approximation parameters during propagation is proposed in our follow-up work~\cite{xue2026proactivelocalminimafreerobotnavigation}.

\begin{figure}[!t]
    \centering
     \includegraphics[width=\linewidth]{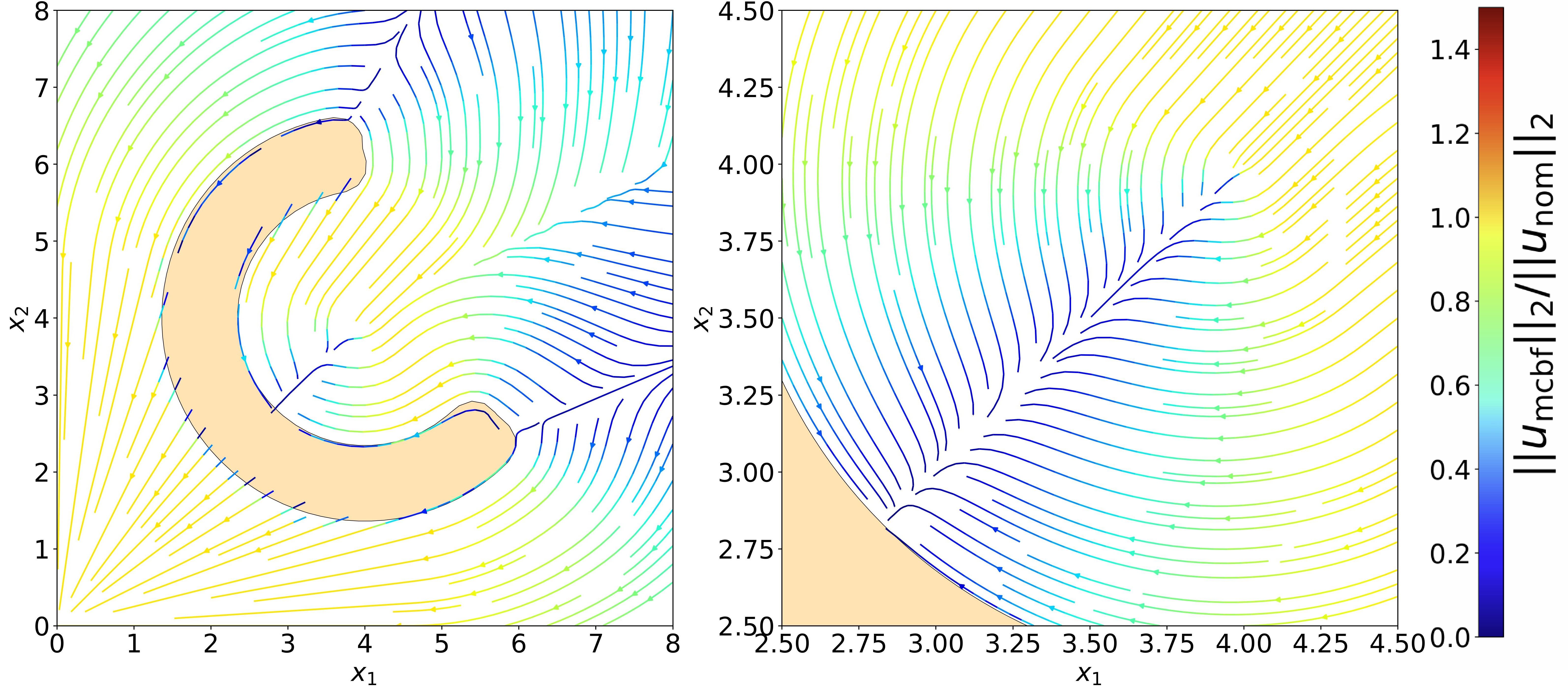}
    \vspace{-20pt}
     \caption{Standard CBF-QP induces undesirable equilibrium behavior in single non-star-shaped obstacle avoidance (left: full view; right: close-up view).\label{fig: cbf nstar}} 
     \vspace{-10pt}
\end{figure}

\begin{figure}[!t]
    \centering
     \includegraphics[width=\linewidth]{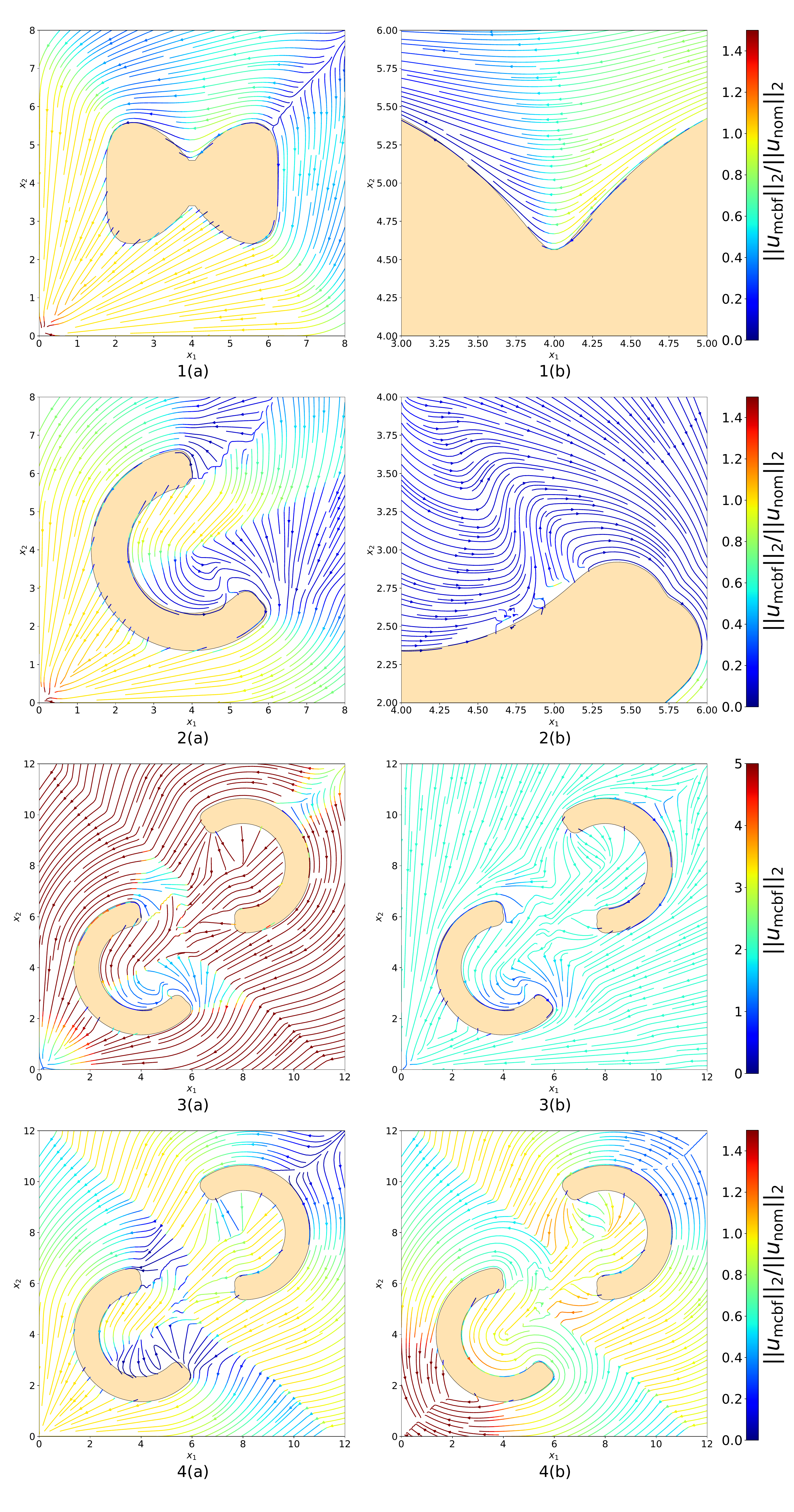}
    \vspace{-20pt}
     \caption{Performance of on-manifold MCBF-QP in single star-shaped (first row) and non-star-shaped (second row, (a): full view, (b): close view) obstacle avoidance with no robot input constraints, in multi-concave obstacle avoidance(third row, (a): without input constraints, (b): with robot input constraints of $\|u_\text{mcbf}\|_2\leq2$), given $\nu_{\phi}=1$. Lastly, pictures on the fourth row show the effects of $\nu_{\phi}$ sizes on the resulted safe trajectories ((a): $\nu_{\phi}=0.1$, (b): $\nu_{\phi}=10$). \label{fig:mod onM cbf} } 
    \vspace{-10pt}
\end{figure}

Unlike reference MCBF-QP, whose feasibility is preserved whenever the original CBF-QP is feasible (without the additional $H_o(x,x_o)$ constraints), the feasible set of on-manifold MCBF-QP is generally more restrictive due to the introduced hard tangent-hyperplane constraint. Additionally, the larger $\nu_{\phi}$ is, the smaller the feasible domain of on-manifold MCBF-QP will be. To mitigate this loss of feasibility, the tangent guidance is derived from the uniform representation $\bar{h}(x,\{x_o\})$ and enforced through a single linear inequality constraint rather than one constraint for each detected obstacle. The performance of the on-manifold MCBF-QP in concave obstacle environments is validated in \autoref{fig:mod onM cbf} using the fully actuated system defined in Eq.~\eqref{eq:fully actuated system} and the linear nominal controller $u_{\text{nom}} = -2x$, and compared against the standard CBF-QP results shown in \autoref{fig: cbf nstar}.

In the remainder of this section, we first discuss the modifications to the geodesic approximation required for the resulting guiding direction $\phi(x,\{x_o\})$ to function properly within the CBF-QP framework for dynamic obstacle avoidance (\autoref{sec:adapt geo}). Given these modifications and the on-manifold MCBF-QP formulation in Eq.~\eqref{eq:mod-phi-cbf-affine}, we establish conditions under which the proposed formulation eliminates undesirable equilibria and dysfunctional circulation, thereby guaranteeing convergence to the target for general control-affine systems (\autoref{sec:mcbf proof}).


\subsubsection{Adapting Geodesic Approximation for CBFs}
\label{sec:adapt geo}
The geodesic approximation mechanism adopted by on-manifold Mod-DS performs well for fully-actuated systems navigating around static obstacles. However, its direct application to underactuated systems and navigation among dynamic obstacle clusters is limited. To ensure that the computed tangent guidance vector $\phi(x,\{x_o\})$ remains consistent with the underlying geodesic direction while preserving the properties of a CBF-based safety filter, we propose the following modifications.

\textbf{Fixed-Manifold Geodesic Approximation:}
In the on-manifold Mod-DS framework, \cite{onManifoldMod} perform geodesic
approximation on a user-selected manifold to regulate the velocity field throughout the workspace. When the robot state does not lie on the selected manifold, additional gradient-descent dynamics are introduced to drive the
system toward the manifold before following the geodesic flow.
Such a gradient-descent mechanism, however, is undesirable in a QP-based safety filter. First, it can introduce significant deviations from the
nominal velocity direction, which may lead to undesirable changes in robot
motion and become incompatible with underactuated systems. Second, enforcing
manifold convergence through additional dynamics can dominate the QP
objective, reducing its ability to minimally modify the nominal controller,
which is the primary motivation for using a QP-based safety filter.

Without a manifold-convergence mechanism, the selection of the fixed
manifold becomes critical. Let the selected manifold be defined as an
offset level set of the combined barrier function,
\begin{equation}
\mathcal{M}_{\epsilon}
=
\left\{
x\in\mathbb{R}^{d}
\mid
\bar{h}(x,\{x_o\})=\epsilon
\right\},
\label{eq:M_e}
\end{equation}
where $\epsilon\geq0$ denotes the distance from the obstacle boundary. An
improper choice of $\epsilon$ may cause the geodesic approximation to ignore
feasible passages whose widths are smaller than approximately $2\epsilon$.
This can result in inefficient paths or, in extreme cases, prevent target
reaching when the only feasible passage is excluded by the selected
manifold.

To maximize the generality of the proposed MCBF-QP framework, we select a
small manifold offset $\epsilon\approx0$, such that the selected manifold
closely approximates the obstacle boundary while avoiding the need for
additional manifold-convergence dynamics. Given a robot state $x$, the
non-position states $x^\text{other}$ are fixed while the position states
$x^\text{pos}=[p_x,p_y,p_z]$ are propagated along the direction $x^*-x$ for a
predefined distance. If the propagated trajectory intersects the selected
manifold at
\[
x_{\mathrm{int}}
=
[x^\text{pos}_{\mathrm{int}},x^\text{other}],
\]
the tangent guidance vector
\[
\phi(x_{\mathrm{int}}(x,x^*),\{x_o\})
\]
is computed using the geodesic approximation method in
Eq.~\eqref{eq:geo approxi}. Otherwise, the robot state is outside the
obstacle-induced region, i.e., $x\notin\bar{\Omega}_{\mathrm{obs}}$, and the
standard CBF constraint in Eq.~\eqref{eq:mod-phi-cbf-affine} is applied.

\textbf{Geodesic Approximation for Crowd Navigation:}
The geodesic approximation adopted by on-Manifold Mod-DS evaluates the tangent guiding vector under the assumption of static obstacles.
While this assumption is generally sufficient for small- to medium-sized moving obstacles, it becomes less effective in crowd navigation, where humans often form large, dynamic clusters whose geometry continuously deforms and translates over time.
Consequently, computing the tangent guiding vector $\phi(x,\{x_o\})$ solely from the current obstacle configuration neglects the future evolution of the cluster, potentially causing the robot to select a longer or less efficient circumvention path.
In extreme cases, the selected tangent direction may coincide with the direction of the cluster's motion, causing the robot to continue following the moving cluster instead of progressing past it.
To extend the proposed framework to dynamic human environments and further reduce undesirable equilibria in crowd navigation, we incorporate predicted obstacle motion into the geodesic approximation by replacing each obstacle position $x_o$ with its one-step prediction $x_o+\beta\dot{x}_o$:
\begin{align}
\nonumber x^{(i+1)}&= x^{(i)} +\beta H\!\left(x^{(i)},\{x_o^{(i)}\}\right)
\nonumber H^\top\!\left(x^{(i)},\{x_o^{(i)}\}\right)e^{(i)},\\
\nonumber e^{(i+1)}&=\frac{H\!\left(x^{(i)},\{x_o^{(i)}\}\right)
H^\top\!\left(x^{(i)},\{x_o^{(i)}\}\right)e^{(i)}}{\left\|
H\!\left(x^{(i)},\{x_o^{(i)}\}\right)H^\top\!\left(x^{(i)},\{x_o^{(i)}\}\right)e^{(i)}
\right\|_2 
},\\
 P^{(i+1)}&=P^{(i)}+\beta p(x^{(i+1)},x^*),\label{eq:geo approx xo}\\
\nonumber x_o^{(i+1)}&=x_o^{(i)}+\beta\dot{x}_o^{(i)},
\quad
\forall x_o^{(i)}\in\{x_o\}^{(i)}.
\end{align}
\begin{equation}
\label{eq:phi geo xo}
\phi =
\underset{
e^{(0)}\in\{e^{(0)}_1,\ldots,e^{(0)}_m\}
}{\arg\min}
\;
P^{(N)}\!\left(e^{(0)},x,\{x_o\},\{\dot{x}_o\}\right).
\end{equation}

Here, the obstacle velocity $\dot{x}_o$ may be obtained from a trajectory prediction model or approximated using the obstacle's current measured velocity.
By accounting for predicted obstacle motion during the geodesic approximation, the computed tangent guiding vector better reflects the anticipated evolution of dynamic obstacle clusters, enabling the robot to select more efficient boundary-following paths and reducing the likelihood of prolonged interactions with moving obstacle clusters.

\subsubsection{Theoretical Guarantees}
\label{sec:mcbf proof}
This section establishes \textit{liveness} guarantees for the proposed OnM-MCBF-QP in multi-obstacle environments with arbitrarily shaped obstacles. Specifically, we show that the closed-loop system admits no undesirable equilibrium (Theorem~\ref{theorem:om-mcbf-equilibrium}) and asymptotically converges to the target under the stated conditions (Theorem~\ref{theorem:om-mcbf-convergence}). The analysis considers the general control-affine system in Eq.~\eqref{eq:affine system}.

\begin{theorem}[\textbf{Equilibrium-Free Guarantee of OnM-MCBF-QP}]
\label{theorem:om-mcbf-equilibrium}
Consider the OnM-MCBF-QP defined in Eq.~\eqref{eq:mod-phi-cbf-affine} for the general control-affine system. Under Assumption~\ref{assumption:nominal-convergence}, together with the feasibility and control-authority conditions of the OnM-MCBF-QP, the resulting closed-loop system admits no undesirable equilibrium.\\
\textbf{Proof:} The proof is provided in Appendix~\ref{app:onm-mcbf-proofs}. \hfill $\blacksquare$
\end{theorem}

\begin{definition}[\textbf{Geodesic Cost-to-Go}]
\label{def:geodesic-cost}
The geodesic cost-to-go $V_G(x,\{x_o\})$ denotes the remaining path length from the current state to the target along the geodesic guidance path induced by the environment. It satisfies
\begin{equation}
V_G(x,\{x_o\})\geq0,\qquad V_G(x^*,\{x_o\})=0.
\end{equation}
The construction of $V_G$ from the nominal trajectory and obstacle geometry is provided in Appendix~\ref{app:onm-mcbf-proofs}.
\end{definition}

To establish convergence to the goal, we impose the following assumptions on the geodesic representation and closed-loop system.

\begin{assumption}[\textbf{Geodesic Cost-to-Go Regularity}]
\label{assumption:VG-regularity}
The geodesic cost-to-go $V_G(x,\{x_o\})$ in Definition~\ref{def:geodesic-cost} is locally Lipschitz with respect to the robot state $x$ and obstacle configurations $x_o$ on the relevant safe configuration space. 
\end{assumption}

\begin{assumption}[\textbf{Geodesic Guidance Consistency}]
\label{assumption:geodesic-guidance-consistency}
The guidance direction $\phi(x,\{x_o\})$ computed by the proposed geodesic approximation in Eqs.~\eqref{eq:geo approxi} and \eqref{eq:geo approx xo}, together with the minimum tangential velocity $\nu_{\phi}>0$ specified in Eq.~\eqref{eq:mod-phi phi constraint}, induces sufficient descent of the geodesic cost-to-go. Specifically, for every point at which geodesic guidance is required, there exists a function $c_G(\nu_{\phi})>0$ such that, at every differentiable point of $V_G$,
\begin{align}
\label{eq:VG_phi_descent_assumption}
\dot{V}_G
&=
\nu_{\phi}\nabla_x V_G(x,\{x_o\})^\top\phi(x,\{x_o\})
\nonumber\\
&\quad+
\sum_{o\in\mathcal{O}}\nabla_{x_o} V_G(x,\{x_o\})^\top\dot{x}_o
\leq -c_G(\nu_{\phi}).
\end{align}
The first term captures the change in geodesic cost due to the predicted robot motion along $\phi$, while the second term captures the change due to the evolution of the obstacle configuration. More generally, let $\partial_C V_G(x,\{x_o\})$ denote the Clarke generalized gradient \citep{clarke1975generalized} of $V_G$ with respect to the joint variable
$(x,\{x_o\})$. For any
$\zeta_G\in\partial_C V_G(x,\{x_o\})$, write
\begin{equation}
\label{eq:zetaG_joint_decomposition}
\zeta_G
=
\left(
\zeta_{G,x},
\{\zeta_{G,o}\}_{o\in\mathcal{O}}
\right),
\end{equation}
where $\zeta_{G,x}$ and $\zeta_{G,o}$ denote the generalized gradient
components with respect to $x$ and $x_o$, respectively. At points
where $V_G$ is not differentiable, the generalized descent condition
is
\begin{equation}
\label{eq:VG_phi_descent_clarke}
\max_{\zeta_G\in\partial_C V_G}
\left[
\nu_{\phi}\zeta_{G,x}^\top\phi(x,\{x_o\})
+
\sum_{o\in\mathcal{O}}
\zeta_{G,o}^\top\dot{x}_o
\right]
\leq
-c_G(\nu_{\phi}).
\end{equation}
Furthermore, the geodesic descent margin is assumed to be
nondecreasing with $\nu_{\phi}$, i.e.,
\begin{equation}
\label{eq:cG_monotonicity}
\nu_{\phi 2}\geq\nu_{\phi 1}>0
\quad\Longrightarrow\quad
c_G(\nu_{\phi 2})\geq c_G(\nu_{\phi 1}).
\end{equation}
\end{assumption}

\begin{assumption}[\textbf{Bounded Residual Components}]
\label{assumption:residual-bounds}
The components of the nominal and geodesic gradients orthogonal to the subspace spanned by the geodesic guidance direction $\phi$ and the normalized aggregate barrier gradient $n_h$ are bounded. Specifically, with
\begin{equation}
n_h=\frac{\nabla\bar h}{\|\nabla\bar h\|},\qquad
P_T=I-\phi\phi^\top-n_hn_h^\top,
\label{eq:nh_definition}
\end{equation}
there exist $\epsilon_N,\epsilon_G\geq0$ such that
\begin{equation}
\|P_T\nabla V_N\|\leq\epsilon_N,\qquad \|P_T\zeta_{G,x}\|\leq\epsilon_G\approx0,
\end{equation}
for $\zeta_G\in\partial_CV_G$. 
The accuracy of $\phi$ as an approximation of the geodesic descent direction implies that the residual geodesic component $\epsilon_G$ is sufficiently small.
\end{assumption}

\begin{assumption}[\textbf{Active-Set Control Authority}]
\label{assumption:active-control-authority}
Consider the control-affine system in Eq.~\eqref{eq:affine system} with the input matrix $g$. When the $\phi$ and CBF constraints are simultaneously active, the control input can independently regulate motion along $\phi$ and $n_h$. Specifically,
\begin{equation}
\label{eq:normal_control_authority}
g^\top n_h\neq0,
\end{equation}
and
\begin{equation}
\label{eq:phi_control_independence}
g^\top\phi
\notin
\operatorname{span}\{g^\top n_h\}.
\end{equation}
Equivalently,
\begin{equation}
\label{eq:Gram_determinant}
\Delta_g
=
\|g^\top\phi\|^2
\|g^\top n_h\|^2
-
\left(
(g^\top\phi)^\top(g^\top n_h)
\right)^2
>0.
\end{equation}
\end{assumption}

\noindent\textbf{Interpretation of Assumptions:}
Assumption~\ref{assumption:VG-regularity} is reasonable because $V_G$ is constructed in Definition~\ref{def:geodesic-cost} from finite-length geodesic segments and is continuous across geodesic entry and exit transitions. Local Lipschitz continuity then permits nondifferentiable points caused by changes in the active detour to be handled through the Clarke generalized gradient.

Assumption~\ref{assumption:geodesic-guidance-consistency} follows from the proposed geodesic approximation (\autoref{sec:adapt geo}).
$\phi$ is selected as the circumvention direction that brings the robot closest to the target among tangent directions with the same propagated distance. Increasing the sampling resolution $m$ and horizon $N$ while reducing $\beta$ improves this approximation, with obstacle motion incorporated through Eq.~\eqref{eq:geo approx xo}.

Assumption~\ref{assumption:residual-bounds} is consistent with this construction because $r_G$ lies in the same tangent space searched by the geodesic approximation. Hence, with sufficiently dense sampling, $\epsilon_G$ can ideally approach zero, while finite sampling, numerical approximation, sensing noise, and obstacle-prediction error may produce a nonzero residual.

Assumption~\ref{assumption:active-control-authority} imposes a configuration-dependent actuation requirement rather than a restriction to fully actuated systems. Some underactuated robots can retain independent authority along $\phi$ and $n_h$ throughout their operating configurations. When this authority is lost, $\Delta_g>0$ excludes those configurations from the guaranteed regime, which may instead be recovered through external perturbations or additional control actions that restore the required independent motion.

\begin{theorem}[\textbf{Asymptotic Convergence of OnM-MCBF-QP}]
\label{theorem:om-mcbf-convergence}
Consider the control-affine system in Eq.~\eqref{eq:affine system} under the proposed OnM-MCBF-QP in Eq.~\eqref{eq:mod-phi-cbf-affine}. Under Assumptions~\ref{assumption:velocity-bounds}, \ref{assumption:nominal-convergence}, \ref{assumption:VG-regularity}, \ref{assumption:geodesic-guidance-consistency}, \ref{assumption:active-control-authority}, and \ref{assumption:residual-bounds}, there exist admissible choices of $\nu_{\phi}>0$ and class-$\mathcal{K}$ function $\alpha(\cdot)$ such that the closed-loop system asymptotically converges to the target $x^*$; i.e., $\lim_{t\rightarrow\infty}||x-x^*||=0$.\\
\textbf{Proof:} The proof is provided in Appendix~\ref{app:onm-mcbf-proofs}-\ref{app:onm-mcbf-discussion} \hfill $\blacksquare$
\end{theorem}

The choices of $\alpha(\cdot)$ and $\nu_{\phi}$ in Eq.~\eqref{eq:mod-phi cbf constraint} and \eqref{eq:mod-phi phi constraint}
for target convergence depend on the geodesic approximation accuracy $\epsilon_G$, control authority, and obstacle motion. The existence of admissible choices is established in Appendix~\ref{app:onm-mcbf-proofs}, while their dependence on these factors is further analyzed in Appendix~\ref{app:onm-mcbf-discussion}.

\section{Robot Experiment}
\label{sec:experiment}

\textbf{Experiment Set \#1 [Linear DS as $u_{\rm{nom}}$]} The obstacle avoidance performance of the proposed MCBF-QP controllers (\autoref{sec:mod-cbf}) is first validated in Gazebo simulations using linear dynamical systems as nominal controllers (\autoref{sec:hospital gazebo}). The Gazebo trials consider both fully-actuated and general control-affine systems with nonholonomic constraints, using the omnidirectional-drive Ridgeback and differential-drive Fetch robots, respectively. The proposed MCBF-QP variants are compared with Mod-DS and CBF-QP baselines. Since the original Mod-DS approaches in \cite{khansari2012dynamical, LukesDS, onManifoldMod} do not account for robot actuation limits, we instead use our proposed speed-constrained Mod-DS variant documented in Appendix \ref{appendix:speed-modulation}, which optimally enforces speed constraints while maintaining theoretical safety guarantees. Note that all Mod-DS strategies evaluate their methods with a linear DS as the nominal controller. Thus, it is an important benchmark to follow.


\textbf{Experiment Set \#2 [MPC/MPPI as $u_{\rm{nom}}$]} To demonstrate the compatibility of the proposed safety filters with diverse motion planning algorithms, \autoref{sec:hospital python} integrates on-manifold MCBF-QP with state-of-the-art planners, including penalty-based MPC and MPPI, in Python-simulated hospital and crowd navigation environments. The resulting systems are evaluated against the corresponding unfiltered planners, standard CBF-QP and circulation-embedded CBF-QP (CE-CBF-QP) safety filters, as well as discretized receding-horizon methods such as MPC-DCBF and MPC-DC.

\textbf{[Hardware Validation]} Finally, to assess the effectiveness of the proposed OnM-MCBF-QP in real-world underactuated robot navigation, the algorithm was deployed on a hardware Fetch robot and compared with that of the standard CBF-QP. 

\subsection{Robot Dynamics}
For fully actuated systems, the trajectories generated by the proposed reference and MCBF-QP methods (\textbf{R-MCBF-QP} and \textbf{OnM-MCBF-QP}) are compared with those from the CBF-QP and constrained Mod-DS baselines through physics-based Gazebo experiments using an omnidirectional Ridgeback robot. The Ridgeback is modeled as a holonomic platform with first-order kinematics:
\begin{equation}
\label{eq:ridgeback sys}
\begin{bmatrix}
u_x \\[3pt]
u_y
\end{bmatrix} = \begin{bmatrix}
\dot{p}_x \\[3pt]
\dot{p}_y
\end{bmatrix},
\end{equation}
where \(p_x\) and \(p_y\) denote the robot’s position along the \(x\)- and \(y\)-axes, respectively.

For underactuated systems, controller performance is evaluated using a
differential-drive Fetch robot in Gazebo. Each controller computes
control inputs based on the standard unicycle kinematic model:
\begin{equation}
\label{eq:standard model}
\dot{x} =
\begin{bmatrix}
\dot{p}_x\\[2pt]\dot{p}_y\\[2pt]\dot{\theta}
\end{bmatrix}
=
\begin{bmatrix}
\cos\theta & 0\\[2pt]
\sin\theta & 0\\[2pt]
0 & 1
\end{bmatrix}
\begin{bmatrix}
v\\[2pt]\omega
\end{bmatrix},
\end{equation}
where \(v\) and \(\omega\) denote the linear and angular velocities,
respectively.

Because the barrier functions \(h(x,x_o)\) defined in
Section~\ref{sec:assumptions} depend only on position, their partial
derivatives with respect to the orientation are zero, i.e.,
\(\partial h / \partial \theta = 0\). As a result, the CBF constraint in
Eq.~\eqref{eq:cbf-qp affine} simplifies to
\begin{align*}
\frac{\partial h}{\partial p_x}\dot p_x +
\frac{\partial h}{\partial p_y}\dot p_y +
\nabla_{x_o}h\,\dot x_o
\ge -\alpha(h).
\end{align*}
Substituting \(\dot{p}_x\) and \(\dot{p}_y\) from
Eq.~\eqref{eq:standard model} into the above expression shows that the
resulting CBF-QP depends only on \(v\), while \(\omega\) does not appear.
Consequently, the standard CBF-QP cannot directly regulate the robot’s
heading, which often limits its ability to guarantee obstacle avoidance.

\subsubsection{Shifted-Model Approach} To address this limitation, we shift
the robot’s reference point by a distance \(a>0\) along its heading,
yielding the modified kinematic model
\begin{equation}
\label{eq:shifted model}
\dot{x} =
\begin{bmatrix}
\dot{p}_x\\[2pt]\dot{p}_y\\[2pt]\dot{\theta}
\end{bmatrix}
=
\begin{bmatrix}
\cos\theta & -a\sin\theta\\[2pt]
\sin\theta & a\cos\theta\\[2pt]
0 & 1
\end{bmatrix}
\begin{bmatrix}
v\\[2pt]\omega
\end{bmatrix}.
\end{equation}
This shift couples the translational and rotational dynamics, allowing
both \(v\) and \(\omega\) to directly affect the barrier function. Hence, the MCBF-QP and CBF-QP controllers can jointly regulate translation and heading for safe navigation. 

A practical advantage of this formulation is that on-manifold MCBF-QP can compute the guiding vector \(\phi(x,x_o)\) used in Eq.~\eqref{eq:mod-phi-cbf-affine} through simple geodesic approximation in the two-dimensional Euclidean position space, similar to the fully actuated case. However, this trick is robot-specific and not readily generalizable to other underactuated robots. We therefore develop a more general on-manifold MCBF-QP implementation that performs geodesic approximation in the non-Euclidean configuration space \([p_x,p_y,\theta]^\top\). For clarity, controllers using the shifted model are denoted S-OnM-MCBF-QP and S-CBF-QP. 

\begin{figure}[!h] 
\centering 
\includegraphics[width=0.5\linewidth]{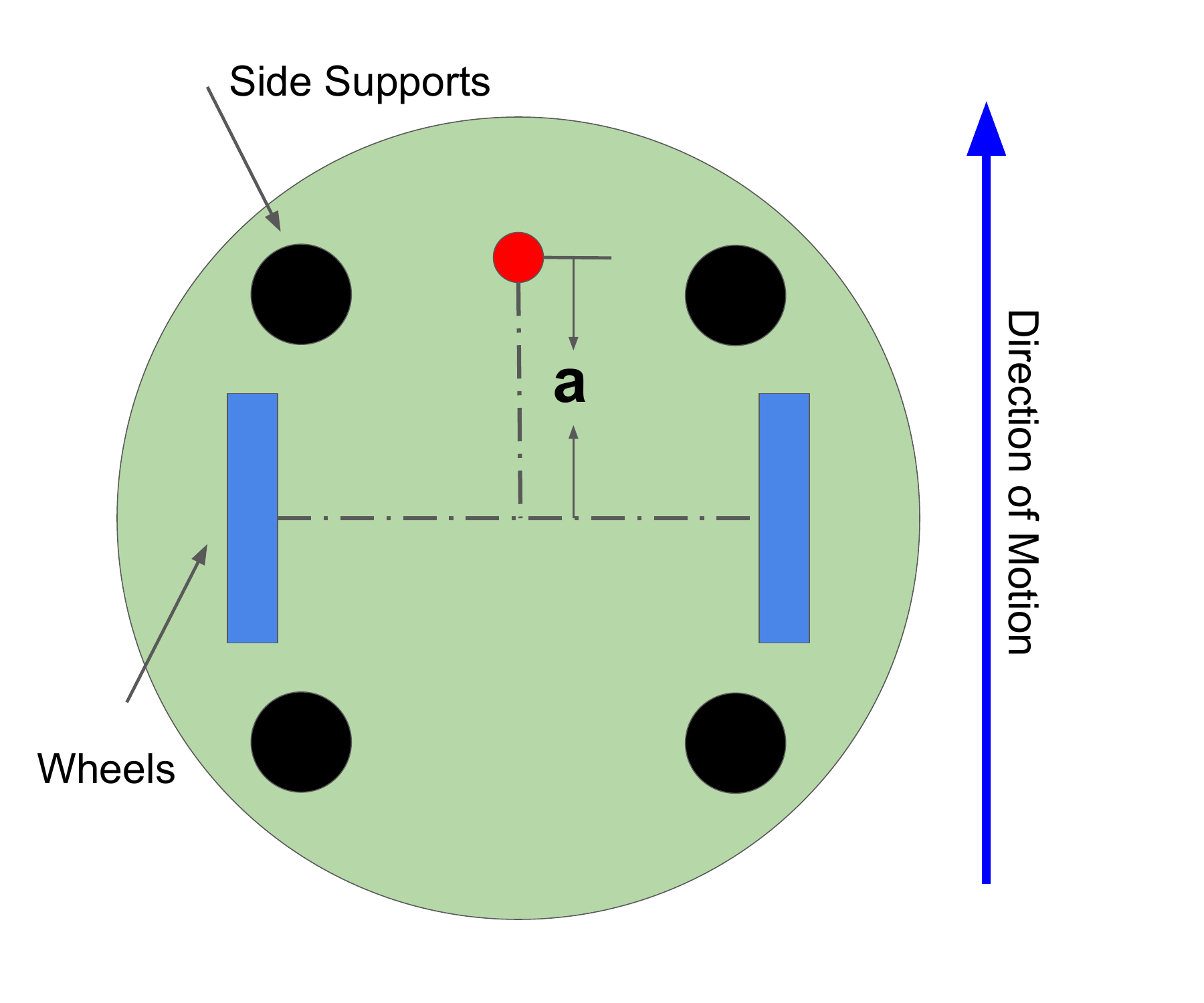} 
\vspace{-10pt}
\caption{Reference point shift along the robot’s heading in the shifted unicycle model.} 
\label{fig:shift point} 
\end{figure}

\subsubsection{Augmented-CBF Approach} 
\label{sec: augmented CBF}
To handle general underactuated systems, we design an augmented control barrier function that incorporates the
robot’s orientation into the barrier definition:
\begin{equation}
\label{eq:augmented barrier function}
h_\text{aug}(x,x_o) =
h(p_x,p_y,x_o)
+ w\frac{\partial h}{\partial p_x}\cos\theta
+ w\frac{\partial h}{\partial p_y}\sin\theta,
\end{equation}
where \(w \in \mathbb{R}^+\) is a design parameter. The added terms capture how the robot’s heading aligns with the gradient of the barrier
function, introducing explicit dependence on \(\theta\). This orientation embedding allows the controller to account for rotational effects directly through the barrier function, without altering the kinematic model. Substituting \(h_\text{aug}\) for \(h\) in the classical CBF constraint yields the augmented CBF constraints used in the A-CBF-QP and A-OnM-MCBF-QP formulations.

\subsection{Evaluation Metrics}
Performance in \autoref{sec:hospital gazebo}, \autoref{sec:hospital python}, and \autoref{sec:real world} is evaluated using four metrics: (i) average travel duration in seconds from start to target (Duration (s)), (ii) percentage of collision-free trajectories (Safe~\%), (iii) success rate of reaching the target (Reached~\%), and (iv) average number of infeasible QP instances per trajectory (Infeasible \#). In Gazebo simulations, each navigation task is repeated ten times from a fixed initial pose to the target, and the collected results are averaged across trials. In Python simulations, each task is evaluated from the same initial position with ten different initial orientations. In the tables, \say{NA} denotes "not applicable," indicating that the travel duration cannot be recorded because the robot fails to reach the target.

\subsection{MCBF-QPs vs. CBF-QPs vs. Mod-DSs}
\label{sec:hospital gazebo} 
\begin{figure}[!tbp]
    \centering
    \subfloat[]{%
        \includegraphics[width=0.45\linewidth]{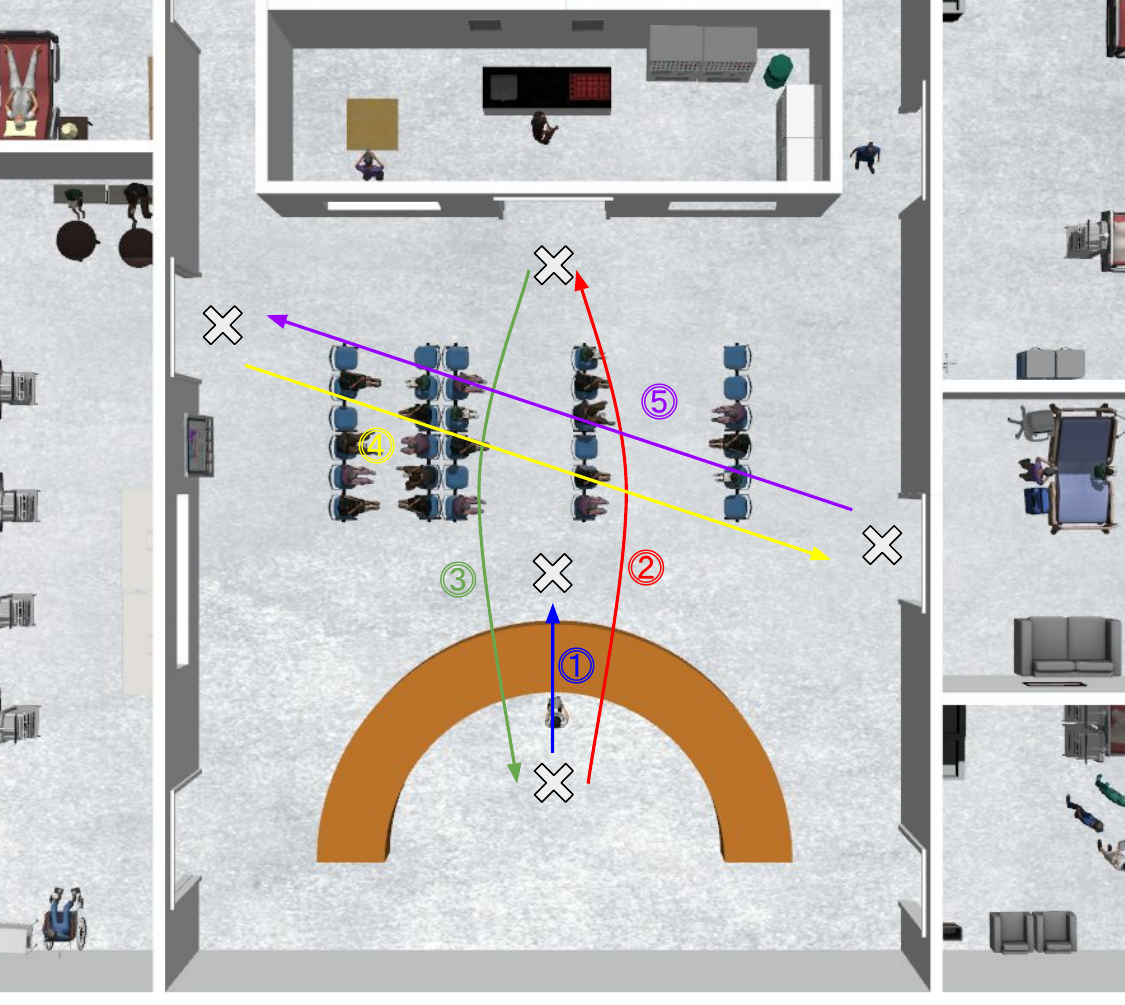}%
        \label{fig:hop sim}%
    }
    \subfloat[]{%
        \includegraphics[width=0.52\linewidth, trim=0 100 0 0pt, clip]{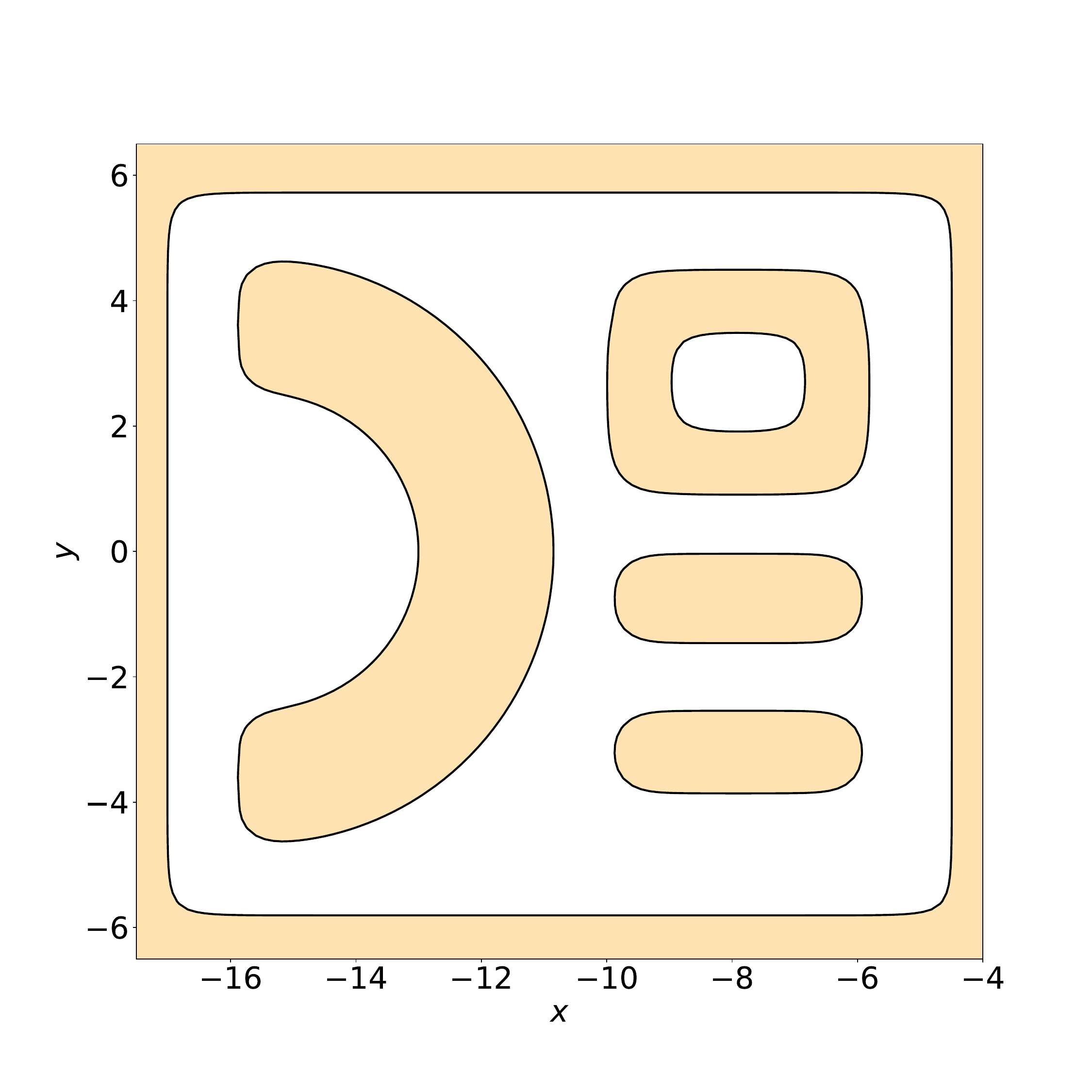}%
        \label{fig:hop rep}%
    }
    \caption{Simulation environment and controller setup for robot comparison tests. Figure (a) presents the five test scenarios, while (b) depicts the robot's perceived view of the hospital environment.}
    \label{fig:hos sim rep}
\end{figure}

\textbf{Environment Settings:} This section evaluates the proposed MCBF-QP controllers in a structured indoor environment resembling a hospital. The environment, adapted from Arena Rosnav \citep{ArenaRosnav2025}, includes both convex and concave static furniture as well as dynamic obstacles such as nurses and patients. Five representative scenarios with different initial and target locations are designed, as shown in \autoref{fig:hop sim}. These scenarios capture typical challenges encountered by delivery robots operating in crowded hospital settings. Scenario~1 tests navigation around a concave obstacle (front desk); Scenarios~2–3 (``mix'' and ``mix-reverse'') combine concave and convex obstacle-avoidance challenges; and Scenarios~4–5 involve only convex obstacles (benches). 

\textbf{Nominal Controller:} For fully actuated systems, the nominal
controller \(u_\text{nom}^l = -\epsilon x\) with
\(\epsilon = 1/\|x\|_2\) is used on the Ridgeback platform. For
underactuated systems, the nominal control inputs are approximated from
the linear dynamical system as
\begin{equation}
\begin{aligned}
v_\text{nom} = \|u_\text{nom}^l\|_2, \quad 
\omega_\text{nom} = \frac{\psi}{\Delta t},
\end{aligned}
\label{eq:dubin nominal system}
\end{equation}
where \(\psi\) is the angular difference between the current orientation
\(\theta\) and the desired orientation inferred from
\(u_\text{nom}^l\), and \(\Delta t\) is the control update period. This
approximation is inexact due to the nonholonomic constraints of
differential-drive robots.

\begin{figure*}[!h]
    \centering
    \subfloat[Fully Actuated Systems in Static Environment]{%
        \includegraphics[width=\linewidth]{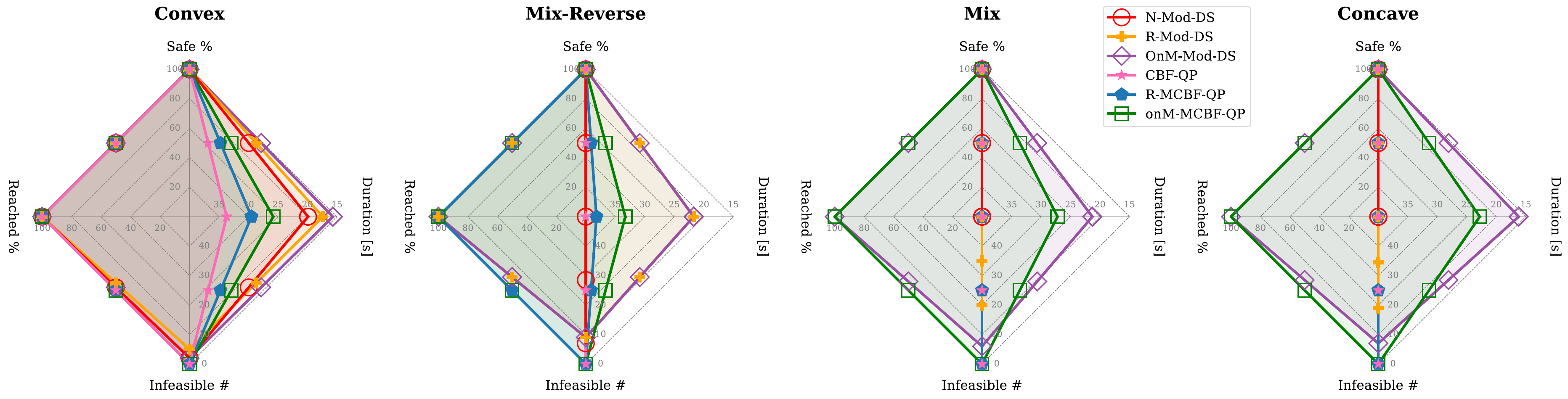}%
        \label{fig:fully-static}%
    }

    \subfloat[Fully Actuated Systems in Dynamic Environment]{%
        \includegraphics[width=\linewidth]{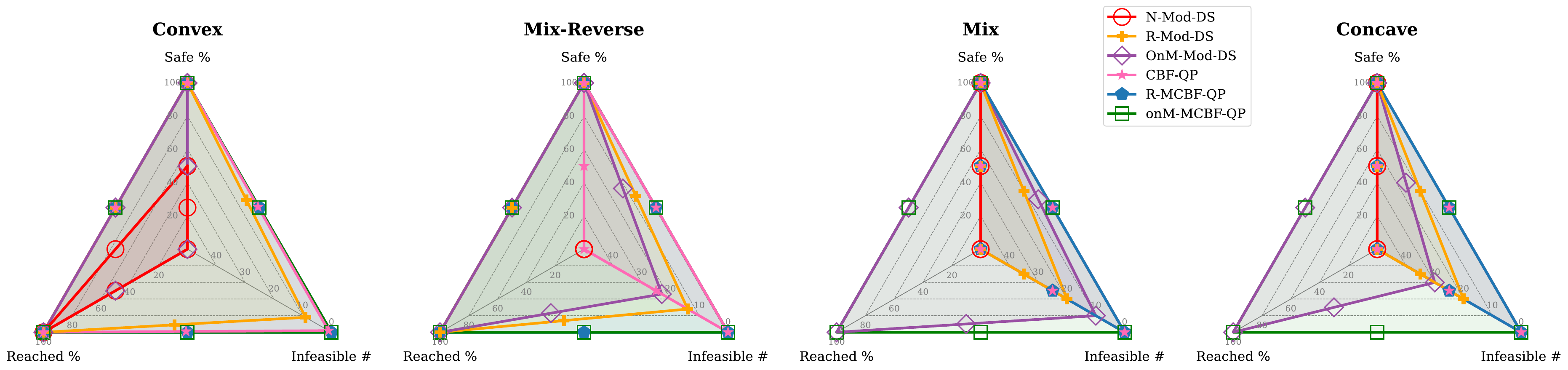}%
        \label{fig:fully-dynamic}%
    }

    \subfloat[Under-Actuated Systems in Static Environment]{%
        \includegraphics[width=\linewidth]{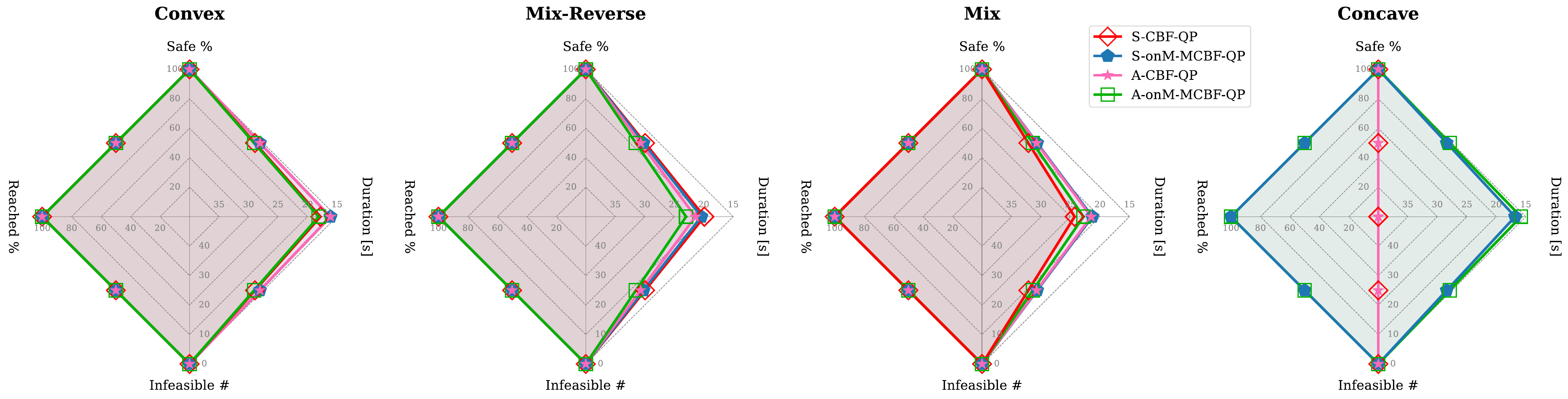}%
        \label{fig:under-static}%
    }

    \subfloat[Under-Actuated Systems in Dynamic Environment]{%
        \includegraphics[width=\linewidth]{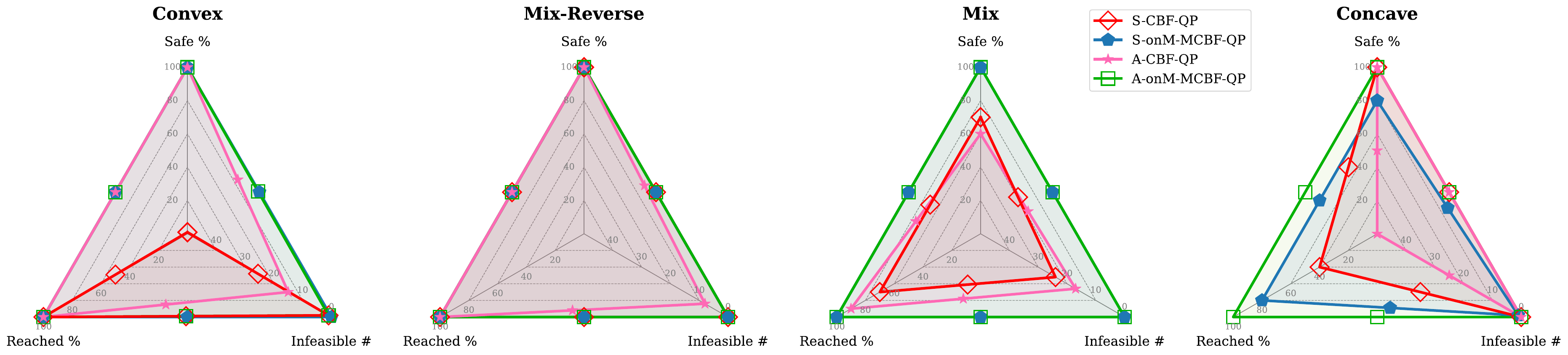}%
        \label{fig:under-dynamic}%
    }

    \caption{Spider plot summarizing controller performance in hospital navigation at 20~Hz. Values are normalized between the center and outermost layer, where a larger radial extent denotes better performance.\label{fig:hospital sim results}}
    \label{fig:hospital_sim_results}
\end{figure*}

The quantitative performance of the proposed Modulated CBF-QP controllers and existing reactive safety controllers is evaluated across hospital scenarios 1--5. For fully actuated systems using the omnidirectional Ridgeback robot, R-MCBF-QP and OnM-MCBF-QP are compared with normal Mod-DS (N-Mod-DS), reference Mod-DS (R-Mod-DS), on-manifold Mod-DS (OnM-Mod-DS), and CBF-QP. For underactuated differential-drive systems using the Fetch robot, S-OnM-MCBF-QP is compared with S-CBF-QP, while A-OnM-MCBF-QP is evaluated against A-CBF-QP. Since Mod-DS approaches are not applicable to underactuated systems, they are excluded from these comparisons. All results are summarized in \autoref{fig:hospital sim results}, where the metrics are normalized between the center and outermost layers of each spider plot, with tick marks indicating the corresponding value ranges. A larger radial extent indicates better performance for the corresponding metric.

From the collected results, several conclusions emerge. In fully actuated systems (\autoref{fig:hospital_sim_results}~(a)--(b)), Mod-DS methods (N-Mod-DS, R-Mod-DS, OnM-Mod-DS) generally achieve shorter travel durations than CBF-QP-based controllers (CBF-QP, R-MCBF-QP, OnM-MCBF-QP). Both approaches ensure safety by decelerating along obstacle normals, but Mod-DS further accelerates tangentially, improving efficiency while increasing infeasibility risks in dense, multi-obstacle settings. When obstacles are closely spaced, Mod-DS exhibits reduced safety performance, as indicated by Safe~\% values of 50\% and 0\% for N-Mod-DS in the dynamic convex and mix-reverse environments, respectively. In contrast, CBF-QP, proven theoretically equivalent to N-Mod-DS in single-obstacle avoidance (\autoref{theorem: equivalance}), remains robust under these conditions. The proposed R-MCBF-QP and OnM-MCBF-QP methods overcome these limitations by maintaining the 100~Safe~\%, equivalent to that of CBF-QP, while substantially reducing infeasibility compared to Mod-DS approaches. Furthermore, R-MCBF-QP achieves a 100~\% reaching rate in the mix-reverse scenario, and OnM-MCBF-QP attains 100~\% reachability across all tested environments. In contrast, CBF-QP fails to complete any tasks under mix, mix-reverse, or concave scenarios. 

In underactuated differential-drive systems (\autoref{fig:hospital_sim_results}~(c)-(d)), both conventional CBF-QP methods (S-CBF-QP and A-CBF-QP) and the proposed modulated variants (S-OnM-MCBF-QP and A-OnM-MCBF-QP) provide strong safety guarantees, with collisions occurring only in rare cases of QP infeasibility in dynamic environments. However, the proposed MCBF-QP controllers significantly improve goal-reaching performance in concave environments. Furthermore, in most scenarios, they achieve shorter travel durations, as the tangential velocity component inherited from OnM-Mod-DS guides the robot along more efficient paths.

Overall, experiments in simulated hospital environments demonstrate that the proposed Modulated CBF-QP methods outperform both Mod-DS and standard CBF-QP baselines across fully actuated and underactuated control-affine systems.

\subsection{More Nominal Controllers (MPC \& MPPI) and Safety Filter Baselines}
To demonstrate that the proposed MCBF-QP filters are not limited to simple nominal controllers based on linear DS, but can instead serve as general safety filters for diverse high-level path planners, we consider sampling-based Model Predictive Path Integral (MPPI) and penalty-based Model Predictive Control (MPC) controllers in addition to the linear DS used in \autoref{sec:hospital gazebo}. We further consider crowd navigation with moving obstacles to demonstrate that these liveness issues are not specific to static environments and can also occur during dynamic obstacle avoidance. For each nominal controller without inherent safety guarantees, we evaluate its performance both without safety filtering and with the proposed on-manifold MCBF-QP, and compare the results with standard CBF-QP \citep{ames2019control}, circulation-embedded CBF-QP (CE-CBF) \citep{cbf_circulation}, MPC-DC, and MPC-DCBF \citep{mpcdc2021safety}. The latter two incorporate obstacle-based safety constraints directly into their MPC formulations rather than using a separate safety-filtering layer.

\begin{figure*}
    \includegraphics[width=\linewidth]{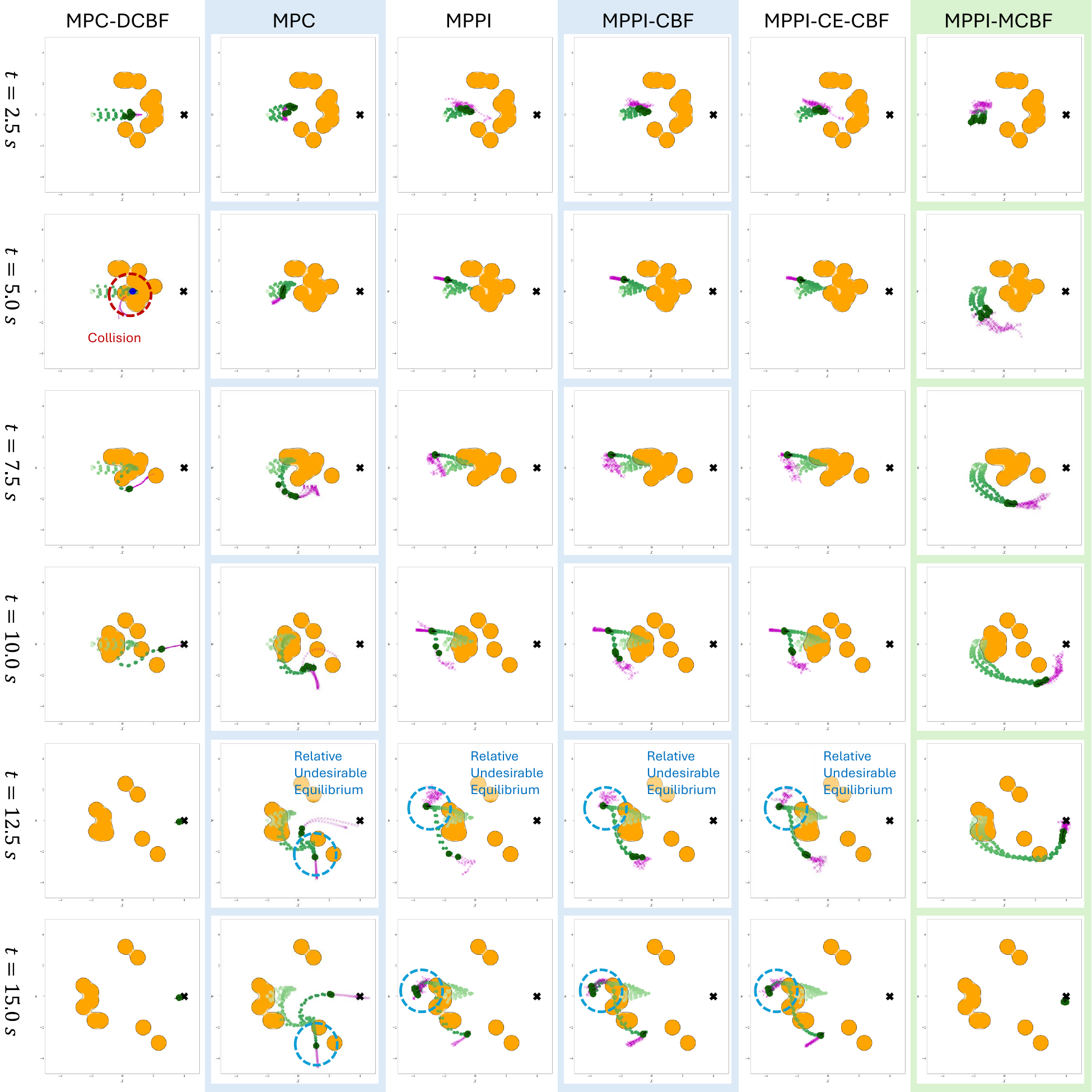}
    \caption{Comparison of the proposed MCBF-QP safety filters with CBF-QP safety filters and MPC-CBF solvers for crowd navigation scenarios using sampling-based MPPI planners as nominal controllers.\label{fig:crowd_nav_sim}}
\end{figure*}

As shown in \autoref{table:nominal experiment} and \autoref{fig:crowd_nav_sim}, both penalty-based and sampling-based MPC controllers achieve low safety rates without safety filtering, despite reporting no solver infeasibility. This demonstrates that solver feasibility alone does not ensure safe execution, as these controllers neither guarantee collision avoidance nor explicitly detect when no safe action is available. Standard CBF-QP improves safety but can leave the robot trapped near undesirable equilibria. In dynamic environments, such stagnation can allow moving obstacles to approach or surround the robot, resulting in higher infeasibility and collision rates than the proposed MCBF-QP, as shown in \autoref{table:nominal experiment}. Although CE-CBF alleviates local minima around concave obstacles, its reliance on an approximated fully actuated model can weaken its safety guarantees for underactuated robots, resulting in the lowest safety rate among the evaluated baselines. Moreover, its circulation mechanism can induce dysfunctional circulation when the robot is simultaneously constrained by multiple nearby obstacles. Consequently, CE-CBF improves the goal-reaching rate over standard CBF-QP but still fails to complete all evaluated scenarios and generally requires substantially longer traversal times than MCBF-QP. MPC-DC and MPC-DCBF incorporate safety constraints over a prediction horizon and therefore can account for future obstacle interactions, but their nonconvex optimization problems do not guarantee globally optimal solutions and can still become infeasible or converge to undesirable equilibria. This results in a 0\% goal-reaching rate in geometrically complex environments. Among all evaluated safety-filtering approaches, the proposed on-manifold MCBF-QP is the only method that consistently achieves both 100\% safety and 100\% goal-reaching rates across all considered nominal controller types.

\label{sec:hospital python}
\begin{table}[!t]
\caption{Performance comparison of the proposed modulated, circulation-embedded, and conventional CBF-QP formulations under different nominal controllers.}
\label{table:nominal experiment}
\centering
\resizebox{0.95\linewidth}{!}{\begin{tabular}{ | m{0.3cm} |m{1.8cm} |m{2.1cm} | m{1.2cm} | m{1.7 cm} |m{1.7cm}|m{1.7cm}|}
  \hline
  \rowcolor{Gray}  & Nominal Controller & Safety Filter & Safe \%  & Reached \% & Infeasible \# &Duration (s)\\ 
  \hline
  \multirow{13}{*}{\rotatebox{90}{Hospital}} & MPC-DC & None & 100 & 0 & 55.4 & NA\\
  \cline{2-7}
  & MPC-DCBF & None & 90 & 0 & 7.3 & NA\\
  \cline{2-7}
   & \multirow{3}{*}{Linear DS} & CBF-QP  & 75 & 75 & 16.2 & \cellcolor{green!10} \textbf{27.3}\\
   & & CE-CBF & 85 & 90 & 21.9 & 37.0\\
  & & \cellcolor{green!10} \textbf{MCBF (ours)} & \cellcolor{green!10} \textbf{100} & \cellcolor{green!10} \textbf{100} & \cellcolor{green!10} \textbf{0.0} & \cellcolor{blue!10} 28.3\\
  \cline{2-7}
  & \multirow{4}{*}{\parbox{1.7cm}{Penalty-based MPC}} & None & 100 & 5 & 0.0 & \cellcolor{green!10} \textbf{23.6}\\
  & & CBF & 90 & 15 & 8.4 & 54.6\\
  & & CE-CBF & 90 & 90 & 10.9 & 47.1\\
  & & \cellcolor{green!10} \textbf{MCBF (ours)} & \cellcolor{green!10} \textbf{100} & \cellcolor{green!10} \textbf{100} & \textbf{0.0} & \cellcolor{blue!10} 32.4\\
  \cline{2-7}
  & \multirow{4}{*}{MPPI} & None & 90 & 5 & 0.0 & 40.8\\
  & & CBF & 100 & 20 & 0.0 & 46.7\\
  & & CE-CBF & 55 & 55& 19.8 & 38.3\\
  & & \cellcolor{green!10} \textbf{MCBF (ours)} & \cellcolor{green!10} \textbf{100} & \cellcolor{green!10} \textbf{100} & \cellcolor{green!10} \textbf{0.0} & \cellcolor{green!10} \textbf{35.0}\\
  \hline
  \multirow{13}{*}{\rotatebox{90}{Crowd Navigation}} & MPC-DC & None & 0 & 100 & 101.1 & 11.3\\
  \cline{2-7}
  & MPC-DCBF & None & 0 & 100 & 69.9 & 11.2\\
  \cline{2-7}
   & \multirow{3}{*}{Linear DS} & CBF-QP  & 100 & 40& 58.6 & 14.4\\
   & & CE-CBF & 0 & 40 & 68.0 & 17.7\\
  & & \cellcolor{green!10} \textbf{MCBF (ours)} & \cellcolor{green!10} \textbf{100} & \cellcolor{green!10} \textbf{100} & \cellcolor{green!10} \textbf{0.0}  & \cellcolor{green!10} \textbf{9.7}\\
  \cline{2-7}
  & \multirow{4}{*}{\parbox{1.7cm}{Penalty-based MPC}} & None & 100 & 20 & 0.0 & \cellcolor{green!10}\textbf{17.5}\\
  & & CBF & 100 & 20& 0.0 & 17.6\\
  & & CE-CBF & 100 & 90 & 0.0 & 17.9\\
  & & \cellcolor{green!10} \textbf{MCBF (ours)} & \cellcolor{green!10} \textbf{100} & \cellcolor{green!10} \textbf{100} & \cellcolor{green!10}\textbf{0.0}  & \cellcolor{blue!10} 20.8\\
  \cline{2-7}
  & \multirow{4}{*}{MPPI} & None & 100 & 30 & 0.0 & 22.7\\
  & & CBF & 100 & 50 & 0.0 & 22.9\\
  & & CE-CBF & 100 & 20 & 0.0 & 22.8\\
  & & \cellcolor{green!10} \textbf{MCBF (ours)} & \cellcolor{green!10}\textbf{100} & \cellcolor{green!10}\textbf{100} & \cellcolor{green!10} \textbf{0.0} & \cellcolor{green!10}\textbf{13.97}\\
  \hline
\end{tabular}}
\end{table}

\begin{table}[!t]
\caption{Performances of proposed On-Manifold Modulated CBF-QP methods in comparison to CBF-QP formulations in real-life experiments using Fetch at 20Hz.}
\label{table:experiment}
\centering
\resizebox{0.9\linewidth}{!}{\begin{tabular}{ | m{1.3cm} | m{2.4cm} | m{1.3cm} | m{1.7 cm} |m{1.5cm}|}
  \hline
  \rowcolor{Gray} Scenario & Methods & Safe \%  & Reached \% & Duration (s)\\ 
  \hline
   \multirow{4}{*}{\parbox{1.3cm}{convex dynamic}} & S-CBF  & 100 & 0 & NA\\
   & \cellcolor{green!10}\textbf{S-OnM-MCBF} & \cellcolor{green!10}\textbf{100} & \cellcolor{green!10}\textbf{100 }& \cellcolor{green!10}\textbf{12.4}\\
  & A-CBF & 100 & 0 & NA\\
  &\cellcolor{green!10}\textbf{A-OnM-MCBF} & \cellcolor{green!10}\textbf{100} & \cellcolor{green!10}\textbf{100 }& \cellcolor{green!10}\textbf{14.3}\\
  \hline
  \multirow{4}{*}{\parbox{1.3cm}{static concave}} &  S-CBF  & 100 & 0 & NA\\
  & \textbf{S-OnM-MCBF} & \textbf{100} & \textbf{100} & \textbf{14.9}\\
  & A-CBF & 100 & 0 & NA\\
  & \cellcolor{green!10}\textbf{A-OnM-MCBF} & \textbf{}\textbf{100} &\cellcolor{green!10}\textbf{80} & \cellcolor{green!10}\textbf{30.7}\\
  \hline
\end{tabular}}
\end{table}

\subsection{Real World Experiments}
\label{sec:real world}
In addition to simulations, the proposed S-OnM-MCBF-QP and A-OnM-MCBF-QP methods are compared to S-CBF-QP and A-CBF-QP in real-life experiments using Fetch given nominal controller in Eq. \eqref{eq:dubin nominal system}. In the first experiment, Fetch is challenged to reach the goal position on the other side of the room despite active attempts from a walking human to block its path. In the second experiment, the robot is tasked to navigate out of a C-shaped cluster region towards the target. These two experiments are repeated five times with different robot initial poses in static scenarios and different human walking trajectories in dynamic scenarios. Lastly, we validated our proposed method in a real-life crowd navigation experiment using the robot Fetch to avoid a 3-person cluster while reaching the target, as presented in \autoref{fig:crowd_nav}. In all hardware experiments, a Vicon motion tracking system was used to capture real-time positions and velocities of both humans and the robot. Videos of these hardware experiments are provided in our project website \url{https://yifanxueseas.github.io/mcbf_web/} and the supplementary multimedia attachment.

The performance of the OnM-MCBF-QPs and CBF-QPs in hardware experiments are summarized in \autoref{table:experiment}. The results demonstrate that CBF-induced undesirable equilibria will deteriorate robot task completion performance, while the introduction of OnM-Mod-based constraints into the same QP problem can sufficiently eliminate those undesirable equilibria. In 2D navigation tasks using Fetch, though both A-OnM-MCBF-QP and S-OnM-MCBF-QP drastically improve robot task completion performance, S-OnM-MCBF-QP using the shifted unicycle model outperforms A-OnM-MCBF-QP using the standard unicycle model in terms of path efficiency and task completion rate. The reason for that is the unsafe region defined by A-CBF $h_\text{aug}(x, x_o)$ \eqref{eq:augmented barrier function}, as shown in \autoref{fig: p tri}, is not the same as the actual obstacle region, causing the geodesic approximation strategy to output sometimes non-optimal choices of tangent vectors $\phi(x,x_o)$ for the robot to track. Additionally, finding an appropriate step size $\beta$ and horizon $N$ to use during geodesic approximation is more challenging for A-MCBF-QP that searches for $\phi(x,x_o)$ in 3D non-Euclidean state spaces, compared with S-MCBF-QP that looks for $\phi(p_x,p_y,x_o)$ instead in 2D Euclidean spaces. Nevertheless, A-OnM-MCBF-QP offers the merit of generalizability, as it can, in principle, be applied to any control-affine robot model.

\section{Discussion and Future Directions}
\label{future}
This paper establishes the theoretical foundations of MCBF-QPs for rigid-body obstacle avoidance. Quantitative, qualitative, and analytical comparisons with optimization-based CBF-QP and closed-form Mod-DS methods in static and dynamic environments show that MCBF-QP combines the undesirable-equilibrium elimination of Mod-DS with the safety robustness and generality of CBF-QP, outperforming both approaches across diverse obstacle-avoidance scenarios.

The additional modulation constraints in on-manifold MCBF-QPs inevitably reduce the instantaneous feasible set. Empirically, when combined with conventional nominal controllers such as MPC or linear dynamical systems, the safety filter can dominate motion generation, leading to abrupt motions and less dynamically feasible trajectories. To mitigate this issue, our follow-up work \citep{admm_mcbf_lca} introduces an ADMM-based higher-level path planner that adapts online using feedback from the MCBF-QP. Future work includes extending Modulated CBF-QPs beyond locomotion to manipulation tasks, where concave joint-space geometries may also induce undesirable equilibria.

\section*{Statements and Declarations}

\subsection*{Ethical considerations}
Not applicable.

\subsection*{Consent to participate}
Not applicable.

\subsection*{Consent for publication}
Not applicable.

\subsection*{Declaration of conflicting interest}
The author(s) declared no potential conflicts of interest with respect to the research, authorship, and/or publication of this article.

\subsection*{Funding statement}
This work was supported by the National Science Foundation (NSF) Foundational Research in Robotics (FRR) program under NSF CAREER Award Grant No. FRR-2443721.

\subsection*{Data availability}
Not applicable.

{
\bibliographystyle{SageH}
\bibliography{refs.bib}
}

\appendix
\newpage
\section{Variable Dictionary}\label{appendix:var}
\vspace{10pt}
\begin{center}
\begin{minipage}{\columnwidth}
\centering
\captionof{table}{Key Variables in This Paper.}
\begin{tabular}{lp{0.65\linewidth}}
\toprule
\textbf{Variable} & \textbf{Description} \\
\midrule
$x$ & Robot state vector \\
$x^*$ & Target position for trajectory planning \\
$p_x$ & Robot position along the $x$-axis \\
$p_y$ & Robot position along the $y$-axis \\
$\theta$ & Robot heading/orientation \\
$x_o$ & Obstacle state vector \\
$u$ & Robot control input vector \\
$v$ & Unicycle linear velocity \\
$\omega$ & Unicycle angular velocity \\
$u_\text{nom}$ & Nominal control input \\
$\alpha(\cdot)$ & Class-$\mathcal{K}$ function \\
$h(\cdot)$ & Control barrier function (CBF)\\
$h_\text{aug}(\cdot)$ & Augmented Control Barrier Function (A-CBF) \\
$H(\cdot)$ & Hyperplane tangent to obstacle surface \\
$E(\cdot)$ & Orthonormal obstacle-based frame \\
$E_r(\cdot)$ & Non-orthonormal reference Mod-DS frame \\
$n(\cdot)$ & Unit vector normal to the obstacle boundary \\
$r(\cdot)$ & Reference vector defined in Eq.~\eqref{eq:reference mod d} \\
$e_i(\cdot)$ & Orthonormal basis vectors spanning the tangent hyperplane $H(\cdot)$ \\
$\lambda$ & Normal direction scaling coefficient in Mod-DS\\
$\lambda_e$ & Tangent direction scaling coefficient in Mod-DS\\
$\rho$ & Slack variables in Eq.~\eqref{eq:mod-r-cbf-affine} \\
$\beta$ & Step size parameter in geodesic approximation \\
$N$ & Horizon size in geodesic approximation \\
$p(x, x^*)$ & User-defined penalty function, typically chosen as the distance from $x$ to the target $x^*$ \\
$\phi(\cdot)$ & Tangent-hyperplane guidance vector estimated using geodesic approximation\\
$\nu_{\phi}$ & User-defined positive real number in Eq.~\eqref{eq:mod-phi-cbf-affine}\\
\bottomrule
\end{tabular}
\end{minipage}
\end{center}

\section{Supplementary Proofs of the CBF-QP and Mod-DS Relationship}
\label{app:proofs}

\subsection{Closed-Form CBF-QP Solutions}
\label{app:closed-form-cbf}

For a single obstacle (or single boundary function) and unconstrained
inputs, the CBF-QP admits a closed-form solution obtained from the
Karush--Kuhn--Tucker (KKT) conditions. When the CBF constraint is
inactive, $u_\text{cbf}=u_\text{nom}$. When the constraint is active,
the solutions to Eqs.~\eqref{eq:cbf-qp affine} and
\eqref{eq:cbf-qp fully actuated} are, respectively,
\begin{align}
u_\text{cbf}
&=
u_\text{nom}
-
\frac{L_fh+L_ghu_\text{nom}+\alpha}
{L_ghL_gh^\top}L_gh^\top,
\label{eq:explicit cbf affine}
\\
u_\text{cbf}
&=
u_\text{nom}
-
\frac{\nabla_xh^\top u_\text{nom}+\alpha}
{\|\nabla_xh\|_2^2}\nabla_xh.
\label{eq:explicit cbf fully actuated}
\end{align}
For notational simplicity, we let $\alpha = \alpha (h(x,x_o))+\nabla_{x_o} h(x,x_o)^\top\dot{x}_o$, $L_fh=L_fh(x,x_o)$, $L_gh=L_gh(x,x_o)$ and $\nabla_xh = \nabla_xh(x,x_o)$.

\subsection{Proof of Theorem~\ref{theorem:saddle-cbfqp}}
\label{app:proof-equilibria-cbfqp}

\textbf{Proof:}
Under inverse collinearity, there exists a scalar $c>0$ such that
\begin{equation}
\nonumber u_{\mathrm{nom}}=-c\nabla_x h(x).
\end{equation}
Substituting this relation into Eq.~\eqref{eq:explicit cbf fully actuated} gives
\begin{equation}
\nonumber
u_{\mathrm{cbf}} =
\begin{cases}
-c\nabla_x h(x),
&
\|\nabla_x h(x)\|_2^2
\leq -\dfrac{\alpha(h(x))}{c},
\\[8pt]
-\dfrac{\alpha(h(x))}
{\|\nabla_x h(x)\|_2^2}\nabla_x h(x),
&
\|\nabla_x h(x)\|_2^2
> -\dfrac{\alpha(h(x))}{c}.
\end{cases}
\label{eq:cbf-equilibrium-cases}
\end{equation}

In the first case,
\begin{equation}
\nonumber
\dot{h}(x)
=
\nabla_x h(x)^\top u_{\mathrm{cbf}}
=
-c\|\nabla_x h(x)\|_2^2
<0,
\end{equation}
so the agent moves toward the obstacle boundary until the second case becomes active.

In the second case, the control remains directed along $-\nabla_x h(x)$, with magnitude
\begin{equation}
\nonumber
\|u_{\mathrm{cbf}}\|
=
\frac{\alpha(h(x))}
{\|\nabla_x h(x)\|_2}.
\end{equation}
Since $\alpha(h(x))\rightarrow0$ as $h(x)\rightarrow0$, and assuming
$\|\nabla_x h(x)\|_2\neq0$ on $\partial\mathcal{C}$,
\begin{equation}
\nonumber
\lim_{h(x)\rightarrow0}
\|u_{\mathrm{cbf}}\|
=0.
\end{equation}
Therefore,
\begin{equation}
\nonumber
\lim_{h(x)\rightarrow0}\dot{x}
=
\lim_{h(x)\rightarrow0}u_{\mathrm{cbf}}
=
\mathbf{0},
\end{equation}
and the agent converges to an undesirable equilibrium on $\partial\mathcal{C}$.
\hfill$\blacksquare$
\subsection{Proof of Theorem~\ref{theorem: equivalance}}
\label{app:proof-mod-ds-cbf-equivalence}

\textbf{Proof:} For simplicity, let $\alpha=\alpha(h(x))$, $\nabla_xh=\nabla_xh(x)$, $H=H(x)$, and $E=E(x)$. The normal Mod-DS basis matrix is
\begin{equation}
E=\begin{bmatrix}\dfrac{\nabla_xh}{\|\nabla_xh\|_2} & H\end{bmatrix}.
\end{equation}
Since $H^\top\nabla_xh=\mathbf{0}$ and the columns of $H$ form an orthonormal basis, $E$ is orthonormal and hence $E^{-1}=E^\top$. Using the closed-form CBF-QP solution in Eq.~\eqref{eq:explicit cbf fully actuated}, its projection onto the obstacle-based frame is
\begin{equation}
E^{-1}u_\text{cbf}=\begin{bmatrix}\dfrac{\nabla_xh^\top u_\text{cbf}}{\|\nabla_xh\|_2}\\ H^\top u_\text{cbf}\end{bmatrix}.
\end{equation}
Thus,
\begin{equation}
E^{-1}u_\text{cbf}=
\begin{cases}
\begin{bmatrix}\dfrac{\nabla_xh^\top u_\text{nom}}{\|\nabla_xh\|_2}\\ H^\top u_\text{nom}\end{bmatrix}, & \nabla_xh^\top u_\text{nom}\geq-\alpha,\\[10pt]
\begin{bmatrix}-\dfrac{\alpha}{\|\nabla_xh\|_2}\\ H^\top u_\text{nom}\end{bmatrix}, & \nabla_xh^\top u_\text{nom}<-\alpha.
\end{cases}
\label{eq:cbf projected obstacle frame}
\end{equation}
For normal Mod-DS, Eqs.~\eqref{eq:M matrix} and \eqref{eq:D matrix} give
\begin{equation}
E^{-1}u_\text{mod}=\begin{bmatrix}\lambda\dfrac{\nabla_xh^\top u_\text{nom}}{\|\nabla_xh\|_2}\\ \lambda_e H^\top u_\text{nom}\end{bmatrix}.
\label{eq:mod proj}
\end{equation}
Choose
\begin{equation}
\lambda=
\begin{cases}
1, & \nabla_xh^\top u_\text{nom}\geq-\alpha,\\[4pt]
-\dfrac{\alpha}{\nabla_xh^\top u_\text{nom}}, & \nabla_xh^\top u_\text{nom}<-\alpha,
\end{cases}
\qquad
\lambda_e=1.
\label{eq:lambda value for equivalence}
\end{equation}
Because $\alpha(h)\rightarrow0$ as $h\rightarrow0$ for an extended $K_{\infty}$ function, these choices satisfy the requirements of Eq.~\eqref{eq:standard mod lambda}. Substituting Eq.~\eqref{eq:lambda value for equivalence} into Eq.~\eqref{eq:mod proj} gives $E^{-1}u_\text{mod}=E^{-1}u_\text{cbf}$. Since $E$ is invertible, $u_\text{mod}=u_\text{cbf}$. Thus, normal Mod-DS and the fully actuated CBF-QP are equivalent under the specified choice of $\lambda$ and $\lambda_e$. \hfill$\blacksquare$

\begin{figure*}[!tbp]
    \centering
    \subfloat[CBF-QP with $\alpha(h)=h$]{%
        \includegraphics[width=0.47\linewidth]{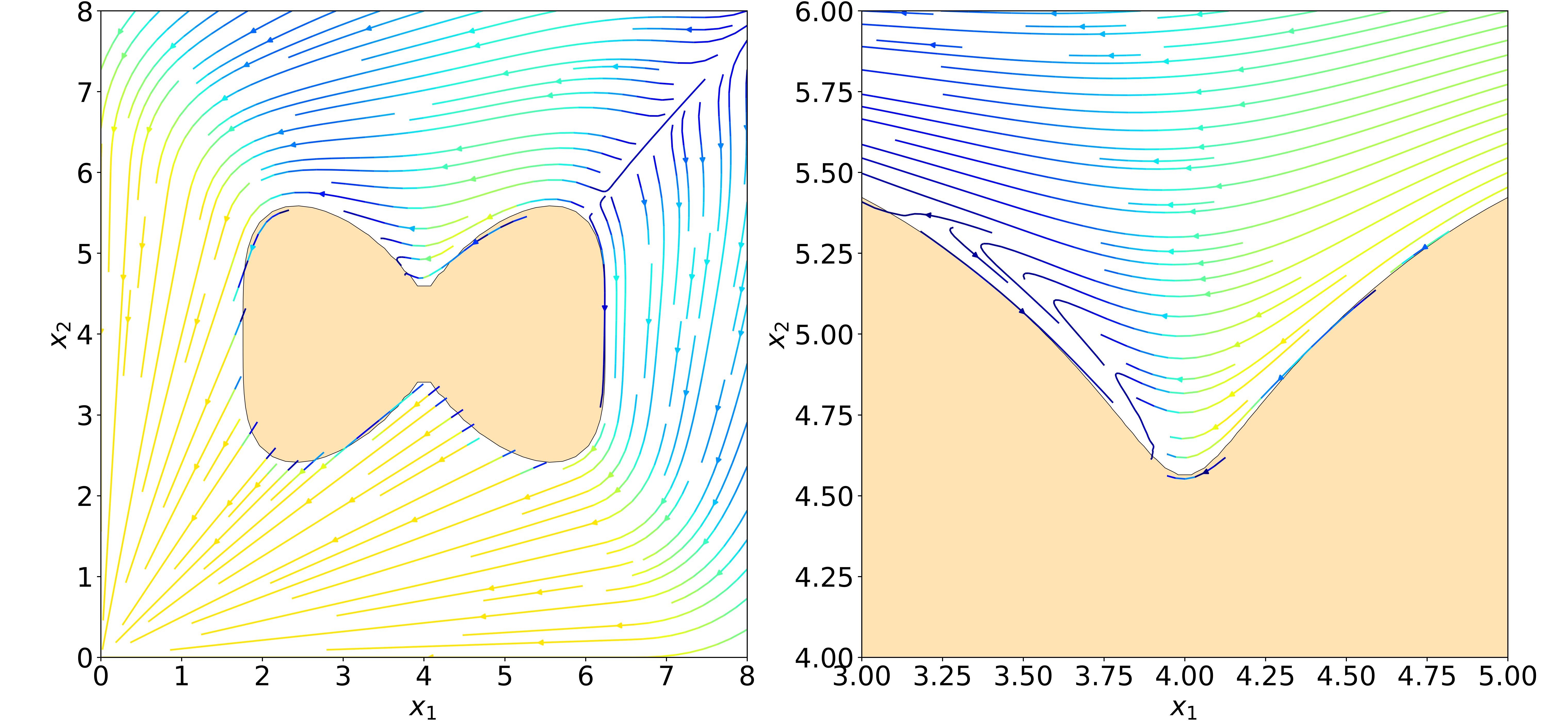}%
        \label{fig: cbf_qp for equivalence}
    }
    \subfloat[Equivalent normal Mod-DS]{%
        \includegraphics[width=0.49\linewidth]{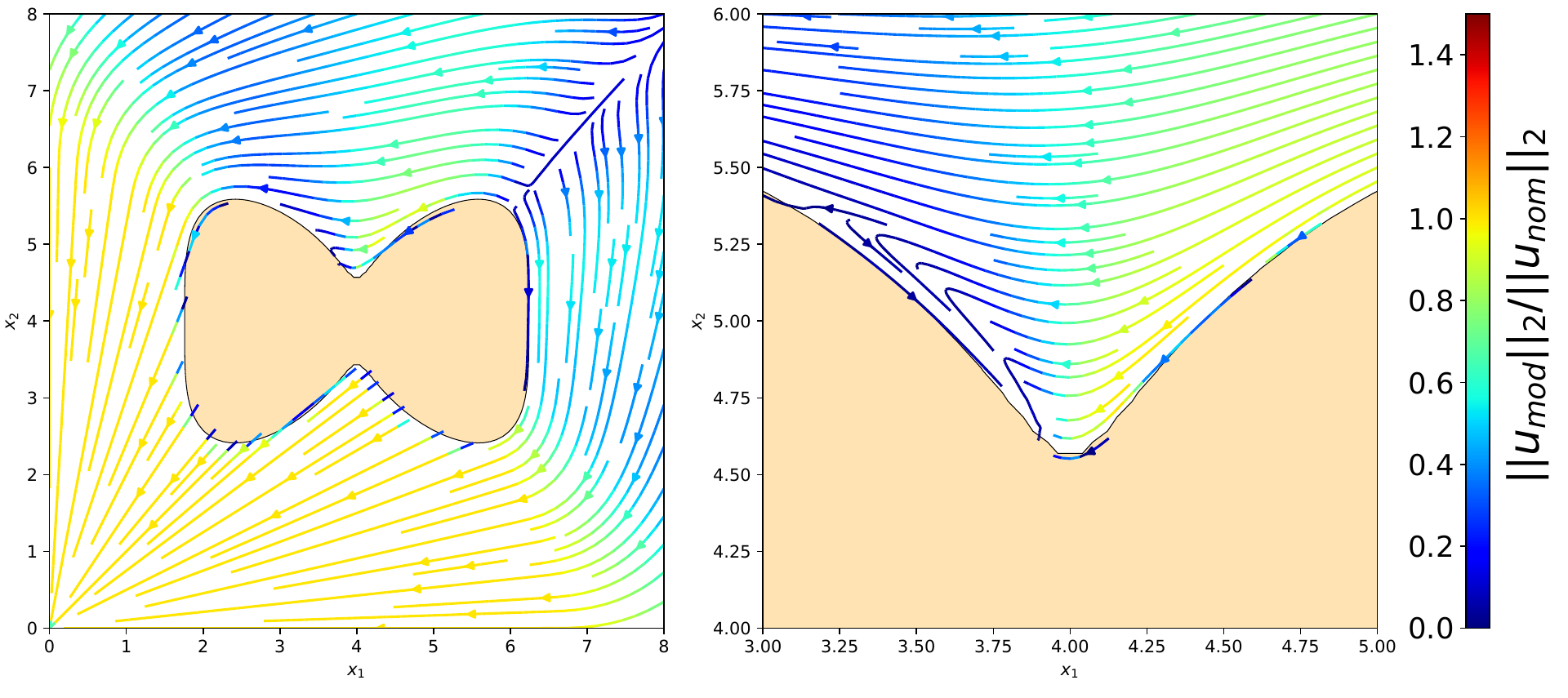}%
        \label{fig: normal mod-ds for equivalence}
    }
    
    \caption{Numerical validation of Theorem~\ref{theorem: equivalance}: the CBF-QP and normal Mod-DS produce identical streamline directions and magnitudes when $\lambda$ and $\lambda_e$ are selected according to the derived equivalence conditions in \autoref{theorem: equivalance}.}
    \label{fig:mod-cbf validation}
\end{figure*}

To further validate Theorem~\ref{theorem: equivalance}, we set $\alpha(h)=h$ and evaluate the corresponding normal Mod-DS and CBF-QP. The resulting vector fields are shown in \autoref{fig:mod-cbf validation}, which confirms that the two controllers produce identical streamline directions and magnitudes.

\begin{figure}[!tph]
    \centering
    \includegraphics[width=0.7\linewidth]{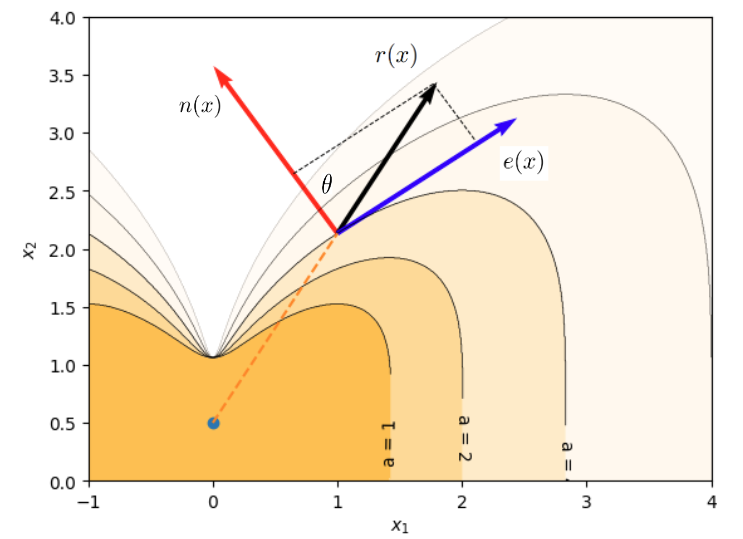}
    \vspace{-10pt}
    \caption{Illustration of the geometric interpretation and relationship between $n(x)$, $e(x)$ and $r(x)$ used for \autoref{theorem:set invariance in reference}. \label{fig:n and r}}
\end{figure}
\subsection{Proof of Theorem~\ref{theorem:set invariance in reference}}
\label{app:proof-set-invariance-reference}

\begin{proof}
For $\forall x\in\mathbb{R}^d$, let $r=r(x,x_o)$, $n=n(x,x_o)$, $e_i=e_i(x,x_o)$, $i\in\{1,\ldots,d-1\}$, and $H=[e_1,\ldots,e_{d-1}]\in\mathbb{R}^{d\times(d-1)}$ denote the reference direction, obstacle normal, and orthonormal basis of the hyperplane orthogonal to $\nabla_xh$, respectively. The reference Mod-DS basis is
\begin{equation}
E_r=[r(x,x_o),H(x,x_o)]\in\mathbb{R}^{d\times d},
\end{equation}
as defined in Eqs.~\eqref{eq:E matrix} and \eqref{eq:reference mod d}. Since $\|r(x,x_o)\|_2=1$, when $\nabla_xh^\top r\neq0$, $E_r$ is invertible as
\begin{align}
\label{eq: Er inv}
E_r^{-1}
&=
\begin{bmatrix}
\dfrac{1}{\nabla_xh^\top r}\nabla_xh^\top\\
-\dfrac{1}{\nabla_xh^\top r}e_1^\top r\nabla_xh^\top+e_1^\top\\
\vdots\\
-\dfrac{1}{\nabla_xh^\top r}e_{d-1}^\top r\nabla_xh^\top+e_{d-1}^\top
\end{bmatrix}\\
&=
\begin{bmatrix}
\dfrac{1}{\nabla_xh^\top r}\nabla_xh^\top\\
H^\top\left(I-\dfrac{1}{\nabla_xh^\top r}r\nabla_xh^\top\right)
\end{bmatrix}.
\end{align}

When $\dot{x}=\mathbf{0}$, the reference-direction component of $u_\text{mod}$ obtained from Eqs.~\eqref{eq:ds-modulation} and \eqref{eq:M matrix} is
\begin{equation}
u_\text{mod}^r
=
\lambda E_r^{-1}[1,:]u_\text{nom}
=
\lambda\frac{\nabla_xh^\top}{\nabla_xh^\top r}u_\text{nom}.
\end{equation}

The decomposition of $u_\text{mod}$ along $r$ can be further projected onto the obstacle-normal direction $n$ and the tangent hyperplane $H$. Since $n^\top H=\mathbf{0}_{1\times(d-1)}$, the projection of $u_\text{mod}$ onto $n$ depends only on its reference-direction component:
\begin{equation}
\label{eq:mod r proj n}
\begin{aligned}
u_\text{mod}^n
&=n^\top(u_\text{mod}^r r+u_\text{mod}^H e)
=u_\text{mod}^r n^\top r\\
&=\left(\lambda\frac{\nabla_xh^\top}{\nabla_xh^\top r}u_\text{nom}\right)
\left(\frac{\nabla_xh}{\|\nabla_xh\|_2}\right)^\top r\\
&=\lambda\frac{\nabla_xh^\top u_\text{nom}}{\|\nabla_xh\|_2}.
\end{aligned}
\end{equation}

It remains to show that this normal component satisfies the CBF condition in Eq.~\eqref{eq:cbf-qp fully actuated}. For a static single-obstacle environment, let $h=h(x,x_o)$ and $\nabla_xh=\nabla_xh(x,x_o)$. Since $\dot{x}_o=\mathbf{0}$, the CBF condition in Eq.~\eqref{eq:cbf-qp fully actuated} reduces to
\begin{equation}
\frac{\nabla_xh^\top}{\|\nabla_xh\|_2}u
\geq
-\frac{\alpha(h)}{\|\nabla_xh\|_2}.
\label{eq:cbf fully actuated constr}
\end{equation}
Substituting the normal component in Eq.~\eqref{eq:mod r proj n} into Eq.~\eqref{eq:cbf fully actuated constr} gives the sufficient condition
\begin{equation}
-\lambda\nabla_xh^\top u_\text{nom}\leq\alpha(h).
\label{eq:alpha sufficient reference mod}
\end{equation}

The bounded-velocity assumption in Eq.~\eqref{eq:velocity_bound} ensures that $\|u_\text{nom}\|$ remains bounded over the relevant trajectories. Since $\lambda$ is bounded by Eq.~\eqref{eq:standard mod lambda}, the left-hand side of Eq.~\eqref{eq:alpha sufficient reference mod} is bounded. Hence, there exists an extended class $K_\infty$ function $\alpha$ that upper-bounds this quantity as a function of $h$. At the boundary $h=0$, Eq.~\eqref{eq:standard mod lambda} gives $\lambda=0$, so the left-hand side of Eq.~\eqref{eq:alpha sufficient reference mod} is zero, consistent with $\alpha(0)=0$. Thus, Eq.~\eqref{eq:alpha sufficient reference mod} holds throughout the safe set, and the reference Mod-DS controller satisfies the CBF condition.
\hfill$\blacksquare$
\end{proof}

\subsection{Proof of Theorem~\ref{theorem:quantitative diff reference}}
\label{app:proof-quantitative-difference}
\textbf{Proof:} Since both normal and reference Mod-DS are reactive safe control approaches with explicit closed-form solutions, given the same $\lambda$ and $\lambda_e$, the differences between reference and normal Mod-DS in regions where both are active can be computed as in \cite{onManifoldMod}. Here, to distinguish reference Mod-DS outputs from the normal ones, we denote results from reference Mod-DS as $u_\text{mod}^R$ and those from the normal one as $u_\text{mod}^N$.
\begin{align}
\nonumber
u_\text{mod}^R&=u_\text{mod}^N-(\lambda_e-\lambda)\sum_{i=1}^{d-1}\frac{e_i^\top r}{n^\top r}e_i n^\top u_\text{nom}\\
&=u_\text{mod}^N-(\lambda_e-\lambda)HH^\top\left(I-\frac{rn^\top}{n^\top r}\right)u_\text{nom}.
\end{align}
Because CBF-QP is shown to be equivalent to normal Mod-DS in \autoref{theorem: equivalance}, we have $u_\text{mod}^N=u_\text{cbf}$. Since the nominal controller satisfies $u_\text{nom}=k(x^*-x)$ with $k>0$, it then follows that
\begin{align}
\label{eq:quantitative diff}
u_\text{mod}^R-u_\text{cbf}
&=
-\lambda^+HH^\top
\left(I-\dfrac{rn^\top}{n^\top r}\right)(x^*-x),
\end{align}
where $\lambda^+=k(\lambda_e-\lambda)$ has the same sign as $\lambda_e-\lambda$.
\hfill$\blacksquare$

\section{Supplementary Proofs of R-MCBF-QP Theoretical Guarantees}
\label{app:proofs rmcbf}
\subsection{Supporting Geometric Lemmas}
\label{app:r-mcbf-geometric-lemmas}
\begin{lemma}[Radially Increasing CBF Representation]
\label{lemma:radially-increasing-cbf}
For any compact, star-shaped obstacle with a $C^1$ boundary, there exists a continuously differentiable barrier function $h(x,x_o)$ such that its zero level set represents the obstacle boundary and $h$ is strictly increasing along every ray emanating from the referent point $x_r(x_o)$ in the safe region. In particular, for any unit radial direction $r$ and $s\geq d_b(r)$,
\begin{equation}
\label{eq:radial-monotonicity-h}
\frac{d}{ds}h\bigl(x_r(x_o)+s r,x_o\bigr)>0,
\end{equation}
where $d_b(r)>0$ denotes the distance from $x_r(x_o)$ to the obstacle boundary along $r$.
\end{lemma}

\begin{proof}
Since the obstacle is compact, star-shaped, and has a $C^1$ boundary, its boundary admits the radial representation
\begin{equation}
x_b(r)
=
x_r(x_o)+d_b(r)r.
\end{equation}
Construct
\begin{equation}
h(x,x_o)
=
\|x-x_r(x_o)\|
-
d_b\bigl(r(x,x_o)\bigr),
\end{equation}
where $r(x,x_o)$ is defined in Eq.~\eqref{eq:reference mod d}. 
The $C^1$ boundary implies $d_b\in C^1(\mathbb{S}^{n-1})$. Since $x_r(x_o)$ lies inside the obstacle, $x\neq x_r(x_o)$ in the safe region, and thus $h\in C^1$. For $x=x_b(r)$,
\begin{equation}
h(x_b(r),x_o)=0.
\end{equation}
Along any fixed ray,
\begin{equation}
h\bigl(x_r(x_o)+s r,x_o\bigr)
=
s-d_b(r),
\end{equation}
and therefore
\begin{equation}
\frac{d}{ds}h\bigl(x_r(x_o)+s r,x_o\bigr)
=
1>0.
\end{equation} \hfill$\blacksquare$
\end{proof}

\begin{lemma}[Radial-Gradient Alignment]
\label{lemma:radial-gradient-alignment}
Given the barrier function $h(x,x_o)$ constructed in Lemma~\ref{lemma:radially-increasing-cbf}, for every $x$ in the safe region $\mathcal{C}$ defined in Eq.~\eqref{eq:safe region},
\begin{equation}
\label{eq:gradient-radial-positive}
\nabla_x h^\top (x-x_r(x_o))>0.
\end{equation}
\end{lemma}

\begin{proof}
Consider the ray passing through $x$, with direction $r(x,x_o)$, and define
\begin{equation}
x(s)=x_r(x_o)+s r(x,x_o).
\end{equation}
Along this ray, $r(x,x_o)$ is constant. Hence, by the chain rule,
\begin{align}
\nonumber \frac{d}{ds}h(x(s),x_o)
&=
\nabla_x h(x(s),x_o)^\top\frac{dx(s)}{ds}\\
&=
\nabla_x h(x(s),x_o)^\top r(x,x_o).
\label{eq:chain rule r}
\end{align}
By the radial monotonicity property in Lemma~\ref{lemma:radially-increasing-cbf},
\begin{equation}
\frac{d}{ds}h(x(s),x_o)>0.
\end{equation}
Combing with Eq.~\eqref{eq:chain rule r} gives
\begin{equation}
\nabla_x h^\top r(x,x_o)>0.
\end{equation}
Since
\begin{equation}
x-x_r(x_o)
=
\|x-x_r(x_o)\|r(x,x_o),
\end{equation}
and $\|x-x_r(x_o)\|>0$, we obtain
\begin{equation}
\nonumber
\nabla_x h^\top(x-x_r(x_o))>0.
\end{equation}
\hfill$\blacksquare$
\end{proof}

\begin{lemma}[CBF Inactivity at Zero Reference Angle]
\label{lemma:cbf-inactivity-zero-angle}
Consider the R-MCBF-QP defined in Eq.~\eqref{eq:mod-r-cbf-affine} for a general control-affine system operating in an environment containing a star-shaped obstacle. Suppose Assumptions~\ref{assumption:nominal-convergence} and~\ref{assumption:nominal-cone-non-increase} hold, and let $\theta(x)$ denote the signed angle between the reference direction $r(x,x_o)$ and the goal direction $x^*-x$. Then, for any $x\in\mathcal{C}\setminus\{x^*\}$ satisfying $\theta(x)=0$, the CBF constraint of the R-MCBF-QP is strictly inactive.
\end{lemma}

\begin{proof}
For $\theta(x)=0$, the reference direction and the goal direction are aligned. By Assumption~\ref{assumption:nominal-cone-non-increase},
\begin{equation}
\dot{\theta}_{\mathrm{nom}}=0.
\end{equation}
Hence,
\begin{equation}
\dot{x}_{\mathrm{nom}}
=
\|\dot{x}_{\mathrm{nom}}\|r(x,x_o).
\end{equation}
By Lemma~\ref{lemma:radial-gradient-alignment} and the reference direction definition in Eq.~\eqref{eq:reference mod d},
\begin{equation}
\nabla_x h^\top r(x,x_o)>0.
\end{equation}
Therefore,
\begin{equation}
\nabla_x h^\top\dot{x}_{\mathrm{nom}}
=
\|\dot{x}_{\mathrm{nom}}\|
\nabla_x h^\top r(x,x_o)
>0.
\end{equation}
Consequently,
\begin{equation}
\nabla_x h^\top\dot{x}_{\mathrm{nom}}+\alpha(h)>0,
\end{equation}
so the CBF constraint is strictly satisfied and cannot be active.\hfill$\blacksquare$
\end{proof}

\subsection{Proof of Theorem~\ref{theorem:r-mcbf-equilibrium}}
\label{app:proof-r-mcbf-equilibrium}

The R-MCBF-QP in Eq.~\eqref{eq:mod-r-cbf-affine} can be considered in two cases depending on whether the CBF constraint is active. When the CBF constraint is inactive, the equality constraint associated with $d_g$ is not imposed, and hence the QP reduces to $u_{\mathrm{mcbf}}=u_{\mathrm{nom}}$. 
For the active case, the equality constraint in Eq.~\eqref{eq:mod-r-cbf-affine} gives
\begin{align}
\label{eq:H_velocity_active}
H^\top\dot{x}_{\mathrm{mcbf}}
&=
H^\top d_g
+
\frac{H^\top r}{\nabla_xh^\top r}
\nabla_xh^\top
(\dot{x}_{\mathrm{mcbf}}-d_g).
\end{align}
Since the CBF constraint is active, it is satisfied with equality, yielding
\begin{equation}
\label{eq:normal_velocity_active}
\nabla_xh^\top\dot{x}_{\mathrm{mcbf}}
=-\alpha.
\end{equation}
Substituting Eq.~\eqref{eq:normal_velocity_active} into Eq.~\eqref{eq:H_velocity_active} yields
\begin{equation}
\label{eq:H_velocity_active_simplified}
H^\top\dot{x}_{\mathrm{mcbf}}
=
H^\top d_g
-
\left(
\nabla_xh^\top d_g+\alpha
\right)
\frac{H^\top r}{\nabla_xh^\top r}.
\end{equation}
Furthermore, using the unit gradient direction $n$ defined in Eq.~\eqref{eq:standard mod d}, the normal component of the velocity is
\begin{align}
\label{eq:n_velocity_active}
n^\top\dot{x}_{\mathrm{mcbf}}
&=
\frac{\nabla_xh^\top\dot{x}_{\mathrm{mcbf}}}{\|\nabla_xh\|}
=
-\frac{\alpha}{\|\nabla_xh\|}.
\end{align}
Since $HH^\top=I-nn^\top$, 
the state update rate can therefore be reconstructed from its tangent and normal components as
\begin{align}
\dot{x}_{\mathrm{mcbf}}
&=
n(n^\top\dot{x}_{\mathrm{mcbf}})
+
HH^\top\dot{x}_{\mathrm{mcbf}}
\nonumber\\
&=
-\frac{\alpha\nabla_xh}{\|\nabla_xh\|^2}
+
HH^\top d_g
-
\left(
\nabla_xh^\top d_g+\alpha
\right)
\frac{HH^\top r}{\nabla_xh^\top r}\nonumber\\
&=
d_g
-
\left(
n^\top d_g+\frac{\alpha}{\|\nabla_xh\|}
\right)
\frac{r}{n^\top r}.\nonumber
\end{align}

Consequently, the closed-loop state update rate of the R-MCBF-QP can be expressed explicitly as
\begin{align}
\label{eq:explicit_dot_x_mcbf}
&~~~~~~\dot{x}_{\mathrm{mcbf}}\\
&\nonumber =
\begin{cases}
\dot{x}_{\mathrm{nom}},
&
\nabla_xh^\top\dot{x}_{\mathrm{nom}}\geq-\alpha,
\\[4pt]
\displaystyle
d_g
-
\left(
n^\top d_g+\frac{\alpha}{\|\nabla_xh\|}
\right)
\frac{r}{n^\top r},
&
\nabla_xh^\top\dot{x}_{\mathrm{nom}}<-\alpha.
\end{cases}
\end{align}
According to the definition of an undesirable equilibrium in \autoref{def:equilibria}, the R-MCBF-QP leads to an undesirable equilibrium if and only if $\dot{x}_{\mathrm{mcbf}}=\mathbf{0}$ while $\dot{x}_{\mathrm{nom}}\neq\mathbf{0}$. Such an equilibrium cannot occur when $\nabla_x h^\top\dot{x}_{\mathrm{nom}}+\alpha\geq0$, since in this case $\dot{x}_{\mathrm{mcbf}}=\dot{x}_{\mathrm{nom}}\neq\mathbf{0}$, according to Assumption~\ref{assumption:velocity-bounds}. Therefore, an undesirable equilibrium can only occur when the CBF constraint is active. From the active-case expression for $\dot{x}_{\mathrm{mcbf}}$, the condition $\dot{x}_{\mathrm{mcbf}}=\mathbf{0}$ is equivalent to
\begin{align}
d_g
=
\left(
n^\top d_g+\frac{\alpha}{\|\nabla_xh\|}
\right)
\frac{r}{n^\top r}.
\label{eq:dg_equilibrium}
\end{align}
Hence, $d_g\parallel r$. 
Since
\begin{align}
d_g
=
\frac{\|f(x)+g(x)u_{\mathrm{nom}}\|}
{\|x^*-x\|}
(x^*-x),
\end{align}
and $\dot{x}_{\mathrm{nom}}=f(x)+g(x)u_{\mathrm{nom}}\neq\mathbf{0}$ at an undesirable equilibrium, we have $d_g\neq\mathbf{0}$. Therefore,
\begin{equation}
x^*-x\parallel r(x,x_o).
\end{equation}
Finally, taking the inner product of Eq.~\eqref{eq:dg_equilibrium} with $n$ yields
\begin{align}
n^\top d_g
&=
\left(
n^\top d_g+\frac{\alpha}{\|\nabla_xh\|}
\right)
\frac{n^\top r}{n^\top r}
\nonumber\\
&=
n^\top d_g+\frac{\alpha}{\|\nabla_xh\|},
\end{align}
which implies
\begin{equation}
\alpha=0.
\end{equation}
This implies $h(x,x_o)=0$, so the robot lies on the obstacle boundary. When $d_g$ and $r$ are positively collinear, $\left(
n^\top d_g+\frac{\alpha}{\|\nabla_xh\|}
\right)>0$ and $\theta=0$. By Lemma~\ref{lemma:cbf-inactivity-zero-angle}, the CBF constraint is strictly inactive, and hence
\begin{equation}
\dot{x}_{\mathrm{mcbf}}
=
\dot{x}_{\mathrm{nom}}
\neq 0.
\end{equation}
In conclusion, an undesirable equilibrium can only be induced by the R-MCBF-QP when the robot lies on the obstacle boundary and the goal direction $x^*-x$ is inversely collinear with the reference direction $r(x,x_o)$.

\subsection{Proof of Theorem~\ref{theorem:r-mcbf-convergence}}
\label{app:proof-r-mcbf-convergence}
We establish convergence using a hybrid Lyapunov function $V_H$ that combines Euclidean distance to the goal with the angular progress characterized by $\theta$.

\begin{lemma}[Hybrid Cone-Based Lyapunov Function]
\label{lemma:cone-lyapunov}
Under Assumption~\ref{assumption:nominal-convergence}, let $\theta(x)$ denote the signed reference angle between the reference direction $r(x,x_o)$ and the goal direction $x^*-x$. For any fixed $w_N>0$, define the hybrid cone-based Lyapunov function
\begin{equation}
\label{eq:cone-lyapunov}
V_H(x)
=
w_N V_N(x)+\theta(x)^2.
\end{equation}
Then $V_H$ is positive definite with respect to $x^*$, i.e.,
\begin{equation}
V_H(x)\geq0,\qquad
V_H(x)=0\Longleftrightarrow x=x^*.
\end{equation}
Moreover, $V_H$ is proper and continuously differentiable away from the collinear configuration.
\end{lemma}

\begin{proof}
Since $V_N$ is positive definite under Assumption~\ref{assumption:nominal-convergence} and $\theta(x)^2\geq0$,
\begin{equation}
\nonumber
V_H(x)
=
w_N V_N(x)+\theta(x)^2
\geq
w_N V_N(x)
\geq0,
\end{equation}
with equality if and only if $x=x^*$. Thus,
\begin{equation}
V_H(x)=0
\quad\Longleftrightarrow\quad
x=x^*.
\end{equation}
Furthermore, since $V_N$ is proper and
\begin{equation}
V_H(x)\geq w_N V_N(x),
\end{equation}
every sublevel set of $V_H$ is contained in a sublevel set of $V_N$ and is therefore bounded. Hence, $V_H$ is proper. Finally, $V_N$ is continuously differentiable and $\theta$ is continuously differentiable away from the collinear configuration. Thus, $V_H$ is continuously differentiable away from the collinear configuration.
\hfill$\blacksquare$
\end{proof}

Following Lemma~\ref{lemma:cone-lyapunov}, it remains to show that, for any initial condition satisfying
\begin{equation}
|\theta(x(0))|
\leq
\pi-\epsilon_\pi,
\qquad
\epsilon_\pi>0,
\end{equation}
there exists $w_N>0$ such that $V_H$ strictly decreases along the corresponding closed-loop trajectory for all $x\neq x^*$. Consequently, every such closed-loop trajectory asymptotically converges to the goal $x^*$. Since, in a single-obstacle environment, the reference constraint is active if and only if the CBF constraint is active, the proof considers separately the cases in which the CBF constraint is nonbinding and binding.

\textbf{Case I: CBF constraint nonbinding.}
When the CBF constraint is nonbinding, the R-MCBF-QP reduces to the nominal controller, and hence
\begin{equation}
\dot{x}_{\mathrm{mcbf}}
=
\dot{x}_{\mathrm{nom}}.
\end{equation}
By Assumption~\ref{assumption:nominal-cone-non-increase}, the nominal controller does not increase $|\theta|$, which implies
\begin{equation}
\frac{d}{dt}\theta^2
\leq
0.
\end{equation}
Under the nominal convergence condition of Assumption~\ref{assumption:nominal-convergence},
\begin{equation}
\dot{V}_N
=
\nabla V_N(x)^\top\dot{x}_{\mathrm{nom}}
<0,
\qquad
x\neq x^*.
\end{equation}
Therefore,
\begin{equation}
\dot{V}_H
=
w_N\dot{V}_N
+
\frac{d}{dt}\theta^2
<0,
\qquad
x\neq x^*.
\end{equation}

\textbf{Case II: CBF constraint binding.}
When the CBF constraint is binding, Eq.~\eqref{eq:explicit_dot_x_mcbf} gives
\begin{equation}
\dot{x}_{\mathrm{mcbf}}
=
d_g-\lambda r,
\qquad
\lambda=
\left(
n^\top d_g+\frac{\alpha}{\|\nabla_xh\|}
\right)
\frac{1}{n^\top r}.
\end{equation}

Consider the two-dimensional subspace
\begin{equation}
\label{eq:2d-subspace}
\mathcal{S}=\operatorname{span}\{d_g,r\}.
\end{equation}
Since $\dot{x}_{\mathrm{mcbf}}$ is a linear combination of $d_g$ and $r$, $\dot{x}_{\mathrm{mcbf}}\in\mathcal{S}$. Hence, the convergence analysis can be restricted without loss of generality to $\mathcal{S}$. Choose coordinates in $\mathcal{S}$ such that
\begin{equation}
x^*=[0\;\;0]^\top,
\qquad
x_r=[d_1\;\;0]^\top,
\qquad
d_1>0,
\end{equation}
and define
\begin{equation}
\label{eq:tilde-x}
\widetilde{x}=x-x_r=[x_1-d_1\;\;x_2]^\top.
\end{equation}
For $x\neq x_r$, let $\theta$ denote the signed angle from the negative $x_1$-axis to $\widetilde{x}$. Hence,
\begin{equation}
\label{eq:reference-angle}
\tan\theta
=
\frac{x_2}{d_1-x_1}.
\end{equation}
In particular, $\theta$ and $x_2$ have the same sign for
$\theta\in(-\pi,\pi)$.
Differentiating Eq.~\eqref{eq:reference-angle} gives
\begin{equation}
\label{eq:theta-dot}
\dot{\theta}
=
\frac{(d_1-x_1)\dot{x}_2+x_2\dot{x}_1}
{(d_1-x_1)^2+x_2^2}.
\end{equation}
Substituting $\dot{x}_{\mathrm{mcbf}}=d_g-\lambda r$ and
$r=\widetilde{x}/\|\widetilde{x}\|$ into Eq.~\eqref{eq:theta-dot}, the contribution of the $r$-term cancels, yielding
\begin{equation}
\label{eq:theta-dot-final}
\dot{\theta}
=
-\frac{d_1\|f(x)+g(x)u_{\mathrm{nom}}\|x_2}
{\|x\|\|\widetilde{x}\|^2}.
\end{equation}
By Assumption~\ref{assumption:nominal-convergence},
$\|f(x)+g(x)u_{\mathrm{nom}}\|>0$ for $x\neq x^*$. Since $\theta$ and $x_2$ have the same sign in the chosen coordinates in $\mathcal{S}$, Eq.~\eqref{eq:theta-dot-final} gives
\begin{equation}
\label{eq:theta-theta-dot}
\theta\dot{\theta}<0,
\qquad
\theta\neq0.
\end{equation}
Consequently,
\begin{equation}
\label{eq:Vcone-dot-binding}
\dot{V}_{\mathrm{cone}}
=
2\theta\dot{\theta}<0.
\end{equation}

Moreover, $|\theta|$ is non-increasing along every closed-loop trajectory:
when the CBF constraint is slack,
Assumption~\ref{assumption:nominal-cone-non-increase} gives
\begin{equation}
\frac{d}{dt}|\theta|
\bigg|_{\dot{x}=\dot{x}_{\mathrm{nom}}}
\leq0,
\end{equation}
while when the CBF constraint is binding, Eq.~\eqref{eq:theta-theta-dot}
gives
\begin{equation}
\frac{d}{dt}|\theta|<0.
\end{equation}
Therefore, the initial condition $|\theta(x(0))|\leq\pi-\epsilon_\pi$
implies
\begin{equation}
|\theta(x(t))|
\leq
\pi-\epsilon_\pi
\end{equation}
for all $t\geq0$.

By Lemma~\ref{lemma:cbf-inactivity-zero-angle}, the CBF constraint is strictly inactive at $\theta=0$. Since the CBF condition is continuous, there exists $\epsilon_\theta>0$ such that
\begin{equation}
\label{eq:theta-binding-separation}
\text{CBF binding}
\quad\Longrightarrow\quad
|\theta|\geq\epsilon_\theta.
\end{equation}
Hence, on the CBF-binding region,
\begin{equation}
\epsilon_\theta
\leq
|\theta|
\leq
\pi-\epsilon_\pi.
\end{equation}
On the considered compact sublevel set, $V_N\in C^1$ by
Assumption~\ref{assumption:nominal-convergence}, and hence $\|\nabla V_N\|$ is bounded. Additionally, the closed-loop velocity $\dot{x}_{\mathrm{mcbf}}$ is assumed to be bounded as in
Assumption~\ref{assumption:velocity-bounds}. Hence, there exists $C_N\geq0$ such that
\begin{equation}
\label{eq:VN-binding-bound}
\dot{V}_N
=
\nabla V_N^\top\dot{x}_{\mathrm{mcbf}}
\leq C_N.
\end{equation}
Furthermore, Eq.~\eqref{eq:theta-dot-final} gives
\begin{equation}
-\dot{V}_{\mathrm{cone}}
=
\frac{
2d_1\|f(x)+g(x)u_{\mathrm{nom}}\|
|\theta||\sin\theta|
}{
\|x\|\|\widetilde{x}\|
}.
\end{equation}
Since
\begin{equation}
\epsilon_\theta
\leq
|\theta|
\leq
\pi-\epsilon_\pi,
\end{equation}
we have
\begin{equation}
\label{eq:theta-sin-lower-bound}
|\theta||\sin\theta|
\geq
\epsilon_\theta
\min\{\sin\epsilon_\theta,\sin\epsilon_\pi\}
>0.
\end{equation}
Moreover, by Assumption~\ref{assumption:velocity-bounds}, the nominal velocity is bounded away from zero on the considered compact CBF-binding region. Since $\|x\|$ and $\|\widetilde{x}\|$ are bounded above on the considered compact sublevel set, Eq.~\eqref{eq:theta-sin-lower-bound} implies that there exists $c_\theta>0$ such that
\begin{equation}
\label{eq:Vcone-binding-bound}
-\dot{V}_{\mathrm{cone}}\geq c_\theta.
\end{equation}
Choose
\begin{equation}
\label{eq:wN-selection}
0<w_N<\frac{c_\theta}{C_N},
\end{equation}
with any $w_N>0$ admissible when $C_N=0$. Then
\begin{equation}
\dot{V}_H
=
w_N\dot{V}_N+\dot{V}_{\mathrm{cone}}
\leq
w_N C_N-c_\theta
<0.
\end{equation}
Therefore,
\begin{equation}
\label{eq:VH-dot-binding}
\dot{V}_H<0
\end{equation}
throughout the CBF-binding region.

\section{Supplementary Proofs of OnM-MCBF-QP Theoretical Guarantees}
\label{app:proofs om-mcbf}

\subsection{Proof of Theorem~\ref{theorem:om-mcbf-equilibrium}}

\begin{proof}
Let $\dot{x}=f(x)+g(x)u_{\mathrm{mcbf}}$. From feasibility of the QP constraint,
\[
\phi(x,x_{o'})^\top \dot{x} \ge \nu_{\phi} > 0.
\]
Suppose there exists an equilibrium $x'$. Then $\dot{x}(x')=\textbf{0}$, which implies
\[
0 = \phi(x^\star,x_{o'})^\top \dot{x}(x^\star) \ge \nu_{\phi},
\]
contradicting $\nu_{\phi}>0$. Hence no equilibrium can exist in any region where the $\phi$-constraint is active.
\end{proof}

\subsection{Construction and Regularity of the Geodesic Cost-to-Go}

The geodesic cost-to-go introduced in Definition~\ref{def:geodesic-cost} is constructed as follows. Let $\Phi_{\mathrm{nom}}(x,t)$ denote the flow of the nominal closed-loop system introduced in Assumption~\ref{assumption:nominal-convergence}. Define the set of states whose nominal trajectories intersect the selected manifold $\mathcal{M}_{\epsilon}$ defined in Eq.~\eqref{eq:M_e} as
\begin{equation}
\nonumber
\bar{\Omega}_G
=
\left\{
x\in\mathcal{X}
\;\middle|\;
\exists t\in[0,\infty)
\text{ s.t. }
\Phi_{\mathrm{nom}}(x,t)\in\mathcal{M}_{\epsilon}(t)
\right\}.
\end{equation}
For $x\in\bar{\Omega}_G\setminus\mathcal{M}_{\epsilon}$, the entry time of the first geodesic detour is defined by
\begin{equation}
\label{eq:tinit_definition}
t_{\mathrm{init}}(x)
=
\inf
\left\{
t\in[0,\infty)
\;\middle|\;
\Phi_{\mathrm{nom}}(x,t)\in\mathcal{M}_{\epsilon}(t)
\right\},
\end{equation}
with corresponding entry point
\begin{equation}
\label{eq:xinit_definition}
x_{\mathrm{init}}(x)
=
\Phi_{\mathrm{nom}}
\left(
x,t_{\mathrm{init}}(x)
\right).
\end{equation}

More generally, let $x_{\mathrm{init}}^{(i)}(x)$ and $x_{\mathrm{end}}^{(i)}(x)$ denote the entry and exit points, respectively, of the $i$th geodesic detour along $\mathcal{M}_{\epsilon}$. The exit point $x_{\mathrm{end}}^{(i)}(x)$ is defined as the first point along the $i$th geodesic detour from which the nominal controller can proceed toward $x^*$ without returning to the same local detour.

Let $N_G(x)$ denote the number of geodesic detours remaining before reaching $x^*$. For $x\notin\mathcal{M}_{\epsilon}$,
\begin{equation}
\label{eq:VG_off_manifold}
V_G(x,\{x_o\})
=
\sum_{i=1}^{N_G(x)}
d_{\mathcal{M}_{\epsilon}^{(i)}}
\left(
x_{\mathrm{init}}^{(i)}(x),
x_{\mathrm{end}}^{(i)}(x)
\right).
\end{equation}
For $x\in\mathcal{M}_{\epsilon}$,
\begin{align}
\label{eq:VG_on_manifold}
V_G(x,\{x_o\})
&=
d_{\mathcal{M}_{\epsilon}}
\left(
x,x_{\mathrm{end}}(x)
\right)\\
\nonumber&+
\sum_{i=2}^{N_G(x)}
d_{\mathcal{M}_{\epsilon}^{(i)}}
\left(
x_{\mathrm{init}}^{(i)}(x),
x_{\mathrm{end}}^{(i)}(x)
\right),
\end{align}
Thus, a geodesic segment currently being traversed contributes only its remaining distance, whereas future segments contribute their complete detour lengths. The two expressions agree at geodesic entry and exit points, making $V_G$ \textbf{continuous} across these transitions. At the target, no geodesic detours remain, i.e.,
$N_G(x^*)=0$, and hence
\begin{equation}
\label{eq:VG_nonnegative}
V_G(x,\{x_o\})\geq0,
\end{equation}
with
\begin{equation}
\label{eq:VG_goal}
V_G(x^*,\{x_o\})=0.
\end{equation} 

\subsection{Proof of Theorem~\ref{theorem:om-mcbf-convergence}}
\label{app:onm-mcbf-proofs}
Convergence is established using the hybrid Lyapunov function $V_H$,
which combines the nominal Lyapunov function $V_N$ with the geodesic
cost-to-go $V_G$ to capture both goal convergence and progress along
the geodesic guidance path.

\begin{lemma}[Hybrid Geodesic Lyapunov Function]
\label{lem:geodesic-lyapunov}
For any fixed $w_N>0$, define
\begin{equation}
\label{eq:VH_definition}
V_H(x,\{x_o\})
=
w_NV_N(x)+V_G(x,\{x_o\}).
\end{equation}
Under Assumptions~\ref{assumption:nominal-convergence}
and~\ref{assumption:VG-regularity}, $V_H$ is positive definite with
respect to $x^*$, proper, and locally Lipschitz on the safe
configuration space.
\end{lemma}

\begin{proof}
Since $V_N$ is positive definite and $V_G$ is nonnegative,
\begin{equation}
\label{eq:VH_positive}
V_H(x,\{x_o\})
\geq
w_NV_N(x)
\geq0.
\end{equation}
Moreover, by the positive definiteness and goal properties of $V_N$ and $V_G$ in
\eqref{eq:VN_properties}, \eqref{eq:VG_nonnegative} and \eqref{eq:VG_goal}, respectively,
\begin{equation}
\label{eq:VH_zero}
V_H(x,\{x_o\})=0
\quad\Longleftrightarrow\quad
x=x^*.
\end{equation}

Since $V_N$ is proper and
$V_H\geq w_NV_N$, every sublevel set of $V_H$ is contained in a
sublevel set of $V_N$. Therefore, $V_H$ is proper. Finally, $V_N$ is
continuously differentiable and $V_G$ is locally Lipschitz by
Assumption~\ref{assumption:nominal-convergence} and \ref{assumption:VG-regularity}. Thus, $V_H$ is locally
Lipschitz. \hfill$\blacksquare$
\end{proof}

Following Lemma~\ref{lem:geodesic-lyapunov}, it remains to show that there exist an admissible class-$\mathcal{K}$ function $\alpha(\cdot)$, a constant $\nu_{\phi}>0$, and a weight $w_N\in(0,1)$ such that $\dot V_H\leq0$ along every closed-loop trajectory in the relevant safe configuration space, with strict decrease for $x\neq x^*$. 

Given $\phi$ and $n_h$ computed using Eq.~\eqref{eq:geo approxi} and \eqref{eq:nh_definition}, the nominal Lyapunov gradient and the robot-state component of a geodesic cost-to-go generalized gradient admit the decompositions
\begin{align}
\label{eq:gradVN_decomposition}
\nabla V_N&=-a_N\phi+b_Nn_h+r_N,\\
\zeta_{G,x}&=-\lambda_G\phi+b_Gn_h+r_G.
\end{align}
where
\begin{align}
\nonumber
a_N&=-\phi^\top\nabla V_N,
&
b_N&=n_h^\top\nabla V_N,
&
r_N&=P_T\nabla V_N,\\
\nonumber
\lambda_G&=-\phi^\top\zeta_{G,x},
&
b_G&=n_h^\top\zeta_{G,x},
&
r_G&=P_T\zeta_{G,x}.
\end{align}
Since $P_T$ projects onto the subspace orthogonal to $\phi$ and $n_h$,
\begin{equation}
\phi^\top r_N=n_h^\top r_N
=\phi^\top r_G=n_h^\top r_G=0.
\end{equation}
By Assumption~\ref{assumption:residual-bounds},
\begin{equation}
\label{eq:residual_bounds}
\|r_N\|\leq\epsilon_N,
\qquad
\|r_G\|\leq\epsilon_G\approx0.
\end{equation}
By continuity of $\nabla V_N$ and compactness of the relevant region, there exists $\bar V_N>0$ such that
\begin{equation}
\label{eq:VN_gradient_bound}
\|\nabla V_N\|\leq\bar V_N.
\end{equation}
Hence,
\begin{equation}
\label{eq:aN_bN_bounds}
|a_N|\leq\bar V_N,
\qquad
|b_N|\leq\bar V_N.
\end{equation}

The proof proceeds by considering the three possible active sets of the proposed on-manifold MCBF-QP.

\noindent\textbf{Case 1: Neither constraint is binding.}

When neither the CBF nor the geodesic constraint is binding, the
closed-loop dynamics coincide with the nominal dynamics. Hence, by
Assumption~\ref{assumption:nominal-convergence},
\begin{equation}
\label{eq:case1_VN_descent}
\dot V_N
=
\nabla V_N^\top\dot{x}_{\mathrm{nom}}
\leq0,
\end{equation}
with strict inequality for all $x\neq x^*$ in the region of
attraction.

If $x\in\bar{\Omega}_G$, the state evolves along the nominal trajectory used
to define $x_{\mathrm{init}}(x)$ in
\eqref{eq:tinit_definition}--\eqref{eq:xinit_definition}. Therefore,
over this interval, $x_{\mathrm{init}}(x(t))$ remains fixed, and the
geodesic cost-to-go in \eqref{eq:VG_off_manifold} is constant. Hence,
\begin{equation}
\label{eq:case1_VG_descent}
\dot V_G=0.
\end{equation}
If $x\notin\bar{\Omega}_G$, then $V_G=0$ by definition, and again
$\dot V_G=0$. Therefore,
\begin{equation}
\label{eq:case1_VH_descent}
\dot V_H
=
w_N\dot V_N
<0,
\qquad
x\neq x^*.
\end{equation}

\noindent\textbf{Case II: The CBF constraint is binding and the
geodesic constraint is slack.}

Since the geodesic constraint is slack, define
\begin{equation}
\label{eq:gamma_prime_definition}
\nu_{\phi}'
:=
\phi^\top\dot{x}
>
\nu_{\phi}.
\end{equation}
Since the CBF constraint is binding, the normal component of the
closed-loop velocity is fixed by
\begin{equation}
\label{eq:normal_binding}
n_h^\top\dot{x}
=
-\frac{
\alpha(\bar h)+\dot{\bar h}_o
}{
\|\nabla_x\bar h\|
}
=:\nu_h.
\end{equation}

At a differentiable point of $V_G$, using the gradient
decompositions in Assumption~\ref{assumption:residual-bounds}, the
derivative of the hybrid Lyapunov function can be decomposed as
\begin{align}
\dot V_H
={}&
-w_Na_N\phi^\top\dot{x}
+w_Nb_N\nu_h
+w_Nr_N^\top\dot{x}
\nonumber\\
&\nonumber -\lambda_G\phi^\top\dot{x}
+b_G\nu_h
+r_G^\top\dot{x}
+
\sum_{o\in\mathcal O}
\nabla_{x_o}V_G^\top\dot{x}_o.
\end{align}
The terms weighted by $w_N$ satisfy
\begin{align}
&-a_N\phi^\top\dot{x}
+b_N\nu_h
+r_N^\top\dot{x}
\nonumber\\
&\qquad\leq
|a_N|\,|\phi^\top\dot{x}|
+
|b_N|\,|\nu_h|
+
\|r_N\|\,\|\dot{x}\|.
\end{align}
By Eq.~\eqref{eq:aN_bN_bounds}, the residual bounds in
Assumption~\ref{assumption:residual-bounds}, and
Eq.~\eqref{eq:velocity_bound}, together with
$\|\phi\|=\|n_h\|=1$, it follows that
\begin{equation}
\label{eq:w_terms_caseII}
\nonumber w_N\left(
-a_N\phi^\top\dot{x}
+b_N\nu_h
+r_N^\top\dot{x}
\right)
\leq
w_N(2\bar V_N+\epsilon_N)\bar v.
\end{equation}

Next, since
\begin{equation}
\nabla_xV_G
=
-\lambda_G\phi+b_Gn_h+r_G,
\end{equation}
and $\phi^\top n_h=\phi^\top r_G=0$, we have
\begin{equation}
\label{eq:lambdaG_gamma_prime}
-\lambda_G\phi^\top\dot{x}
=
-\lambda_G\nu_{\phi}'
=
\nu_{\phi}'\nabla_xV_G^\top\phi.
\end{equation}
Therefore,
\begin{align}
&-\lambda_G\phi^\top\dot{x}
+
\sum_{o\in\mathcal O}
\nabla_{x_o}V_G^\top\dot{x}_o
\nonumber=
\nu_{\phi}'\nabla_xV_G^\top\phi
+
\sum_{o\in\mathcal O}
\nabla_{x_o}V_G^\top\dot{x}_o.
\end{align}

The geodesic guidance consistency in Assumption~\ref{assumption:geodesic-guidance-consistency} applied with $\nu_{\phi}'$ gives
\begin{equation}
\label{eq:geodesic_descent_gamma_prime}
\nu_{\phi}'\nabla_xV_G^\top\phi
+
\sum_{o\in\mathcal O}
\nabla_{x_o}V_G^\top\dot{x}_o
\leq
-c_G(\nu_{\phi}').
\end{equation}
Since $\nu_{\phi}'>\nu_{\phi}$, the monotonicity condition in
Eq.~\eqref{eq:cG_monotonicity} implies
\begin{equation}
\label{eq:cG_gamma_prime}
c_G(\nu_{\phi}')\geq c_G(\nu_{\phi}).
\end{equation}
Consequently,
\begin{equation}
\label{eq:geodesic_descent_caseII}
-\lambda_G\phi^\top\dot{x}
+
\sum_{o\in\mathcal O}
\nabla_{x_o}V_G^\top\dot{x}_o
\leq
-c_G(\nu_{\phi}).
\end{equation}

The remaining geodesic residual satisfies
\begin{equation}
\label{eq:rG_caseII}
r_G^\top\dot{x}
\leq
\|r_G\|\,\|\dot{x}\|
\leq
\epsilon_G\bar v.
\end{equation}
Combining Eqs.~\eqref{eq:normal_binding}, \eqref{eq:w_terms_caseII},
\eqref{eq:geodesic_descent_caseII}, and
\eqref{eq:rG_caseII} gives
\begin{align}
\label{eq:caseII_alpha_gamma_bound}
\nonumber\dot V_H
&\leq
w_N(2\bar V_N+\epsilon_N)\bar v
-
\frac{b_G}{\|\nabla_x\bar h\|}
\left(
\alpha(\bar h)+\dot{\bar h}_o
\right)\\
&\quad -c_G(\nu_{\phi})
+
\epsilon_G\bar v.
\end{align}

At a nondifferentiable point of $V_G$, let
$\zeta_G\in\partial_CV_G(x,\{x_o\})$ and write
\begin{equation}
\zeta_G
=
\left(
\zeta_{G,x},
\{\zeta_{G,o}\}_{o\in\mathcal O}
\right),
\end{equation}
with
\begin{equation}
\zeta_{G,x}
=
-\lambda_G\phi+b_Gn_h+r_G.
\end{equation}
The same decomposition gives
\begin{align}
&-\lambda_G\nu_{\phi}'
+
\sum_{o\in\mathcal O}
\zeta_{G,o}^\top\dot{x}_o
\nonumber=
\nu_{\phi}'\zeta_{G,x}^\top\phi
+
\sum_{o\in\mathcal O}
\zeta_{G,o}^\top\dot{x}_o.
\end{align}
By the generalized geodesic descent condition in
Eq.~\eqref{eq:VG_phi_descent_clarke} and the monotonicity condition
in Eq.~\eqref{eq:cG_monotonicity}, this term is bounded above by
$-c_G(\nu_{\phi})$. Thus, Eq.~\eqref{eq:caseII_alpha_gamma_bound}
also holds for the corresponding Clarke generalized derivative.

Since $\bar V_N$, $\epsilon_N$, and $\bar v$ are finite, $w_N>0$
can then be chosen sufficiently small such that
\begin{equation}
\label{eq:caseII_w_choice}
w_N(2\bar V_N+\epsilon_N)\bar v
<
\delta_N,
\end{equation}
where $\delta_N>0$ can be chosen arbitrarily small. Thus, the
$w_N$-weighted nominal contribution can be made arbitrarily small
and does not affect the sufficient condition on $\alpha(\cdot)$ and $\nu_{\phi}$.

Therefore, Case II is guaranteed to admit strict decrease provided
\begin{equation}
\label{eq:caseII_sufficient_condition}
-
\frac{b_G}{\|\nabla_x\bar h\|}
\left(
\alpha(\bar h)+\dot{\bar h}_o
\right)
-c_G(\nu_{\phi})
+\epsilon_G\bar v
<0.
\end{equation}

\noindent\textbf{Case III: Both the CBF and geodesic constraints are binding.}

In this case,
\begin{equation}
\label{eq:case3_active_constraints}
\phi^\top\dot{x}=\nu_{\phi},
\qquad
n_h^\top\dot{x}=\nu_h.
\end{equation}
Under Assumption~\ref{assumption:active-control-authority},
$\Delta_g>0$. The corresponding minimum-norm state-space directions
$P_\phi$ and $P_h$ are given by
\begin{align}
P_\phi
&=
\frac{
g g^\top
\left[
\|g^\top n_h\|^2\phi
-
(g^\top\phi)^\top(g^\top n_h)n_h
\right]
}{
\Delta_g
},
\label{eq:case3_Pphi}
\\
P_h
&=
\frac{
g g^\top
\left[
\|g^\top\phi\|^2n_h
-
(g^\top\phi)^\top(g^\top n_h)\phi
\right]
}{
\Delta_g
}.
\label{eq:case3_Ph}
\end{align}
They satisfy
\begin{align}
\phi^\top P_\phi&=1,
&
n_h^\top P_\phi&=0,
\\
\phi^\top P_h&=0,
&
n_h^\top P_h&=1.
\end{align}

Hence, the closed-loop velocity admits the decomposition
\begin{equation}
\label{eq:case3_velocity_decomposition}
\dot{x}
=
\nu_{\phi} P_\phi
+
\nu_hP_h
+
\dot{x}_{\perp},
\end{equation}
where $\dot{x}_{\perp}$ denotes the component of $\dot{x}$ in the
null space of the active constraint directions. Consequently,
\begin{equation}
\label{eq:case3_xperp_orthogonality}
\phi^\top\dot{x}_{\perp}=0,
\qquad
n_h^\top\dot{x}_{\perp}=0.
\end{equation}

On the relevant compact active set, continuity and
$\Delta_g>0$ imply the existence of finite constants
$\bar P_\phi,\bar P_h>0$ such that
\begin{equation}
\label{eq:case3_P_bounds}
\|P_\phi\|\leq\bar P_\phi,
\qquad
\|P_h\|\leq\bar P_h.
\end{equation}

Using Eq.~\eqref{eq:case3_velocity_decomposition}, the derivative of
the hybrid navigation function can be decomposed as
\begin{align}
\dot V_H
={}&
\nu_{\phi}\nabla_xV_H^\top P_\phi
+
\nu_h\nabla_xV_H^\top P_h
+
\nabla_xV_H^\top\dot{x}_{\perp}
\nonumber\\
&+
\sum_{o\in\mathcal O}
\nabla_{x_o}V_G^\top\dot{x}_o.
\label{eq:case3_dVH_decomposition}
\end{align}

We first consider the contribution associated with the active
geodesic constraint. Define
\begin{equation}
\label{eq:case3_Cphi}
C_\phi
=
\nu_{\phi}\nabla_xV_H^\top P_\phi
+
\sum_{o\in\mathcal O}
\nabla_{x_o}V_G^\top\dot{x}_o.
\end{equation}
Using the gradient decomposition in
Assumption~\ref{assumption:residual-bounds} and the properties of
$P_\phi$ gives
\begin{align}
C_\phi
={}&
-\nu_{\phi} w_Na_N
-\nu_{\phi}\lambda_G
+\nu_{\phi}(w_Nr_N+r_G)^\top P_\phi
\nonumber\\
&+
\sum_{o\in\mathcal O}
\nabla_{x_o}V_G^\top\dot{x}_o.
\label{eq:case3_Cphi_exact}
\end{align}
Since
\begin{equation}
-\lambda_G
=
\nabla_xV_G^\top\phi,
\end{equation}
the geodesic terms in Eq.~\eqref{eq:case3_Cphi_exact} satisfy
\begin{align}
&-\nu_{\phi}\lambda_G
+
\sum_{o\in\mathcal O}
\nabla_{x_o}V_G^\top\dot{x}_o
\nonumber\\
&\qquad=
\nu_{\phi}\nabla_xV_G^\top\phi
+
\sum_{o\in\mathcal O}
\nabla_{x_o}V_G^\top\dot{x}_o
\leq
-c_G(\nu_{\phi}),
\end{align}
where the inequality follows from
Assumption~\ref{assumption:geodesic-guidance-consistency}.
Consequently, given boundedness of $\nabla V_H$ in Eq.~\eqref{eq:aN_bN_bounds},
\begin{align}
C_\phi
\leq{}&
-c_G(\nu_{\phi})
+
\nu_{\phi} w_N
\left(
\bar V_N+\epsilon_N\bar P_\phi
\right)
+
\nu_{\phi}\epsilon_G\bar P_\phi.
\label{eq:case3_Cphi_bound}
\end{align}

Next, consider the contribution associated with the active CBF
constraint:
\begin{equation}
\label{eq:case3_Ch}
C_h
=
\nu_h\nabla_xV_H^\top P_h.
\end{equation}
Using the gradient decomposition in Eq.~\eqref{eq:gradVN_decomposition} and the properties of $P_h$,
\begin{align}
C_h
={}&
\nu_h
\left[
w_Nb_N+b_G
+
(w_Nr_N+r_G)^\top P_h
\right].
\end{align}
Hence,
\begin{align}
C_h
\leq{}&
|\nu_h|
\left[
w_N
\left(
\bar V_N+\epsilon_N\bar P_h
\right)
+
\epsilon_G\bar P_h
\right]+ \nu_h b_G.
\label{eq:case3_Ch_bound}
\end{align}

The remaining component $\dot{x}_{\perp}$ is orthogonal to both
$\phi$ and $n_h$. Therefore,
\begin{equation}
\nabla_xV_H^\top\dot{x}_{\perp}
=
(w_Nr_N+r_G)^\top\dot{x}_{\perp},
\end{equation}
and
\begin{equation}
\label{eq:case3_xperp_bound}
\nabla_xV_H^\top\dot{x}_{\perp}
\leq
\left(
w_N\epsilon_N+\epsilon_G
\right)\bar{v}.
\end{equation}

Combining Eqs.~\eqref{eq:case3_Cphi_bound},
\eqref{eq:case3_Ch_bound}, and
\eqref{eq:case3_xperp_bound} yields
\begin{align}
\dot V_H
\leq{}&
-c_G(\nu_{\phi})
+
\nu_{\phi}\epsilon_G\bar P_\phi
+\nu_h b_G+ \epsilon_G|\nu_h|\bar P_h
+
\epsilon_G\bar{v}
\nonumber\\
&+
w_N
\Big[
\nu_{\phi}
\left(
\bar V_N+\epsilon_N\bar P_\phi
\right)
+
|\nu_h|
\left(
\bar V_N+\epsilon_N\bar P_h
\right)\nonumber\\
&+
\epsilon_N\bar{v}
\Big].
\label{eq:case3_full_bound}
\end{align}

All quantities multiplying $w_N$ are finite on the relevant active
set. Thus, for any $\delta_N>0$, $w_N>0$ can be selected sufficiently
small such that
\begin{equation}
\label{eq:case3_w_choice}
w_N
\Big[
\nu_{\phi}
\left(
\bar V_N+\epsilon_N\bar P_\phi
\right)
+
|\nu_h|
\left(
\bar V_N+\epsilon_N\bar P_h
\right)
+
\epsilon_N\bar{v}
\Big]
<
\delta_N,\nonumber
\end{equation}

Accordingly, after making the $w_N$-dependent contribution arbitrarily small and substituting $\nu_h$ from Eq.~\eqref{eq:normal_binding}, the $w_N$-independent sufficient condition for Case III is
\begin{align}
\label{eq:case3_alpha_gamma_condition}
c_G(\nu_{\phi})
&>
\nu_{\phi}\epsilon_G\bar P_\phi
-
\frac{
b_G
}{
\|\nabla_x\bar h\|
}(\alpha(\bar h)+\dot{\bar h}_o)\\
&\quad + \epsilon_G\Big[\frac{
\bar P_h
}{
\|\nabla_x\bar h\|
}|\alpha(\bar h)+\dot{\bar h}_o|
+\bar{v}\Big] \nonumber.
\end{align}

\noindent\textbf{Conclusion:}
Case I requires no additional condition on $\alpha(\cdot)$ or
$\nu_{\phi}$. For Cases II and III, sufficient conditions are given by
Eqs.~\eqref{eq:caseII_sufficient_condition} and
\eqref{eq:case3_alpha_gamma_condition}, respectively. Under
Assumption~\ref{assumption:residual-bounds},
$\epsilon_G\approx0$. Since the relevant active set is compact,
$\alpha(\bar h)+\dot{\bar h}_o$ is continuous, and
$\|\nabla_x\bar h\|>0$ on the active set, there exists a finite
constant $\bar q_h>0$ such that
\begin{equation}
\label{eq:q_h_bound}
\frac{
\left|
\alpha(\bar h)+\dot{\bar h}_o
\right|
}{
\|\nabla_x\bar h\|
}
\leq
\bar q_h.
\end{equation}

Thus, neglecting the arbitrarily small residual terms, the dominant
condition shared by Cases II and III is
\begin{equation}
\label{eq:common_alpha_gamma_condition}
c_G(\nu_{\phi})
>
-
\frac{
b_G
}{
\|\nabla_x\bar h\|
}
\left(
\alpha(\bar h)+\dot{\bar h}_o
\right).
\end{equation}
The signed term on the right-hand side may be negative when the
CBF-induced normal motion assists geodesic descent and positive when it opposes geodesic descent.

By Assumption~\ref{assumption:VG-regularity}, $V_G$ is locally
Lipschitz on the relevant compact active set. Hence, its Clarke
generalized gradient is bounded, so there exists $L_G>0$ such that
\begin{equation}
\label{eq:VG_generalized_gradient_bound}
\|\zeta_G\|\leq L_G,
\qquad
\forall\zeta_G\in\partial_C V_G.
\end{equation}
Consequently,
\begin{equation}
\label{eq:bG_bound}
|b_G|
=
|n_h^\top\zeta_{G,x}|
\leq
\|\zeta_{G,x}\|
\leq
\|\zeta_G\|
\leq
L_G.
\end{equation}
Therefore, the worst-case CBF-induced contribution satisfies
\begin{equation}
\label{eq:common_alpha_gamma_worst_case}
-
\frac{
b_G
}{
\|\nabla_x\bar h\|
}
\left(
\alpha(\bar h)+\dot{\bar h}_o
\right)
\leq
L_G\bar q_h.
\end{equation}

Hence, $\nu_{\phi}$ can be selected such that
\begin{equation}
\label{eq:gamma_existence_condition}
c_G(\nu_{\phi})
>
L_G\bar q_h.
\end{equation}
Together with the sufficient conditions for Cases I, II and III, this
establishes
\begin{equation}
\dot V_H\leq0
\end{equation}
along every closed-loop trajectory, with strict decrease for
$x\neq x^*$. By Lemma~\ref{lem:geodesic-lyapunov},
\begin{equation}
x(t)\rightarrow x^*
\qquad\text{as}\qquad
t\rightarrow\infty.
\end{equation}
This completes the proof.

\subsection{Discussion of Convergence Conditions}
\label{app:onm-mcbf-discussion}
The sufficient convergence condition of the OnM-MCBF-QP provides further insight into the factors governing the convergence margin. In particular, it characterizes the interplay among the tangent progress requirement, CBF-induced normal motion, geodesic approximation error, and available control authority.

First, $\nu_{\phi}$ and $\alpha(\cdot)$ affect the closed-loop behavior through fundamentally different mechanisms. The parameter $\nu_{\phi}$ sets the required tangent progress and increases the geodesic descent margin $c_G(\nu_{\phi})$. In contrast, $\alpha(\cdot)$ determines $\nu_h$, the normal velocity imposed by the CBF. Its effect on convergence is therefore determined by the orientation of the geodesic gradient relative to the safety-induced normal motion, as captured by the signed term $b_G\nu_h$. Consequently, a stronger CBF intervention can either assist or oppose geodesic descent.

Second, $\epsilon_G$ characterizes the accuracy of the computed
geodesic direction relative to the desired geodesic descent
direction. A smaller $\epsilon_G$ reduces the residual contributions in both Cases II and III, thereby providing greater tolerance to other perturbations in the system. In practice, sensing noise, obstacle-prediction error, numerical approximation, and finite sampling can potentially increase $\epsilon_G$ and reduce the available convergence margin.

Third, the interaction between $\nu_{\phi}$, $\epsilon_G$, and the actuation matrix $g(x)$ is particularly important in Case III. The term$\nu_{\phi}\epsilon_G\bar P_\phi$ is qualitatively different from the other residual terms because it is multiplied by the same parameter $\nu_{\phi}$ that increases the required geodesic progress. Moreover, $\bar P_\phi$ reflects how efficiently the available control directions can generate motion along $\phi$ while remaining orthogonal to the CBF normal direction $n_h$. If the control directions provided by $g(x)$ have weak projection onto the required tangent direction, or if the two active constraint directions become nearly indistinguishable after mapping through $g(x)$, then $\bar P_\phi$ can become large. This does not by itself preclude convergence, since the resulting residual penalty can remain small when $\epsilon_G$ is sufficiently small. Thus, the benefit of increasing $\nu_{\phi}$ depends on whether the resulting increase in the geodesic descent margin $c_G(\nu_{\phi})$ is sufficient to offset the $\nu_{\phi}\epsilon_G\bar P_\phi$ penalty, which in turn depends on both the accuracy of the geodesic approximation and the effectiveness of the available actuation in realizing the required tangent motion.

Finally, obstacle motion and the nominal controller affect the convergence margin through different channels. Obstacle motion enters the safety intervention through $\dot{\bar h}_o$ and also contributes directly to the geodesic descent condition through
$\sum_{o\in\mathcal O}\nabla_{x_o}V_G^\top\dot{x}_o$.
Rapid obstacle motion can therefore increase the normal safety intervention while simultaneously changing the available geodesic descent margin. The nominal controller affects the closed-loop velocity bound $\bar v$, which in turn determines the magnitude of the residual perturbations. Hence, the practical convergence margin is also influenced by the motion of the environment and the behavior of the nominal controller.

\section{Constrained Mod-DS in Fully Actuated Systems}
\label{sec:constraing-mod}
All Mod-DS approaches have limitations in handling input constraints. For closed-form methods in general, implementing velocity constraints poses challenges as modifications to the modulated velocity might compromise the agent's guaranteed impenetrability. Here in this section, by posing Mod-DS's input constraining process as an independent convex optimization problem and proving mathematically that such optimization problem either has simple closed-form solutions or can be solved sufficiently fast, we argue that Mod-DS's constraints enforcement ability in fully actuated systems is equivalent to that of CBF-QP. In robot applications, inputs constraints typically can be categorized into 2 types: speed constraints $\|\dot{x}\|_2 = \|u\|_2 \leq u_\text{ub}$, and velocity or box constraints $U_\text{lb}\leq u \leq U_\text{ub}$, where $u_\text{ub}\in \mathbb{R}$ and $U_\text{lb}$, $U_\text{ub} \in \mathbb{R}^d$. Following, one optimization problem is constructed for each constraint type. We again abused notation here and set $n = n(x,x_o)$, which is defined in Eq.~\eqref{eq:standard mod d}.

\subsection{Speed-Constraining Mod-DS}
\label{appendix:speed-modulation}
When the unconstrained output from Mod-DS $u_\text{unc} = \dot{x}_\text{unc}$ exceeds the speed limit, solving the following Quadratically Constrained Linear Programming (QCLP) problem finds the constrained output $u_\text{c}$ that is most similar to $u_\text{unc}$ and ensures safety defined in Eq.~\eqref{eq:const safe} while satisfying speed input constraint $u_\text{ub}$ in Eq.~\eqref{eq:const speed}. In both speed-constraining and velocity-constraining Mod-DS, robot safety is ensured by restricting the robot's velocity moving towards the obstacle, i.e. the projection of $u_\text{c}$ onto vector $n(x,x_o)$, to be no larger than that of $u_\text{unc}$. 
\begin{gather}
\label{eq:obj}
u_\text{c}= \argmax_{u} \; \langle u_\text{unc}, u \rangle\\
\label{eq:const speed} \|u\|_2 \leq u_\text{ub}\\
\label{eq:const safe} \langle n, u \rangle \geq \min(\langle n,\bar{\dot{x}}_{o}\rangle,\langle n, u_\text{unc}\rangle)
\end{gather}

In speed-constraining Mod-DS, we make the design choice to measure similarity between the constrained and unconstrained Mod-DS outputs using dot products between the two vectors so that the QP has simple closed-form solutions. Using dot products as cost functions is reasonable because $\langle u_\text{unc}, u \rangle = \|u_\text{unc}\|_2\|u\|_2\cos \theta_d$, where $\theta_d \in [0,\pi]$ is the angle difference between the two vectors. $\|u_\text{unc}\|_2$ is fixed because $u_\text{unc}$ is a constant vector input into the QP. When $\|u\|_2$ is also fixed, the dot product between $u_\text{unc}$ and $u$ is inversely correlated to $\theta_d$. In other words, the larger the dot product is, the more similar $u$ is to $u_\text{unc}$ orientation-wise. We know the magnitude of the optimal solution of the QP $u_\text{c}$ must be fixed to $\|u_\text{c}\|_2=u_\text{ub}$, because the above optimization problem would only be called to modify $u_\text{unc}$ when $\|u_\text{unc}\|_2 > u_\text{ub}$. Therefore, the optimization problem can be reformulated as the following. Note that $\langle a, b \rangle = a^\top b$ when $a$, $b \in \mathbb{R}^d$ and $\|u\|^2_2=u^\top u$.
\begin{gather}
\label{eq:qp speed}
u_\text{c}= \argmax_{u} \; u_\text{unc}^\top u \\
\nonumber u^\top u = (u_\text{ub})^2\\
\nonumber n^\top u \geq \min(n^\top \bar{\dot{x}}_{o},n^\top u_\text{unc})
\end{gather}

Denote $v_n = \min(n^\top\bar{\dot{x}}_{o},n^\top u_\text{unc})$. Solving for the explicit solutions of the above convex optimization problem using Karush–Kuhn–Tucker (KKT) conditions, a simple explicit solution is found as in Eq.~\eqref{eq:mod speed}, where $v_e = \sqrt{\|u_\text{unc}\|_2^2-\|n^\top u_\text{unc}\|_2^2} = \|H^\top u_\text{unc}\|_2$ is the projection of $u_\text{unc}$ onto the tangent hyperplane $H=H(x,x_o)$  defined in Eq.~\eqref{eq:H matrix}.
\begin{equation}
\label{eq:mod speed}
\begin{cases}
\frac{u_\text{ub}}{\|u_\text{unc}\|_2}u_\text{unc} \quad \qquad \qquad \qquad \text{if} \quad \frac{u_\text{ub}}{\|u_\text{unc}\|_2}n^\top u_\text{unc}\geq v_n\\
v_nn + \frac{\sqrt{(u_\text{ub})^2-(v_n)^2}}{v_\text{e}}(u_\text{unc}-u_\text{unc}^\top n n)\quad  \text{otherwise}  \\
\end{cases}
\end{equation}

\begin{figure}
    \centering
     \includegraphics[trim={0.75cm 0.2cm 1cm 0cm},clip,width=\linewidth]{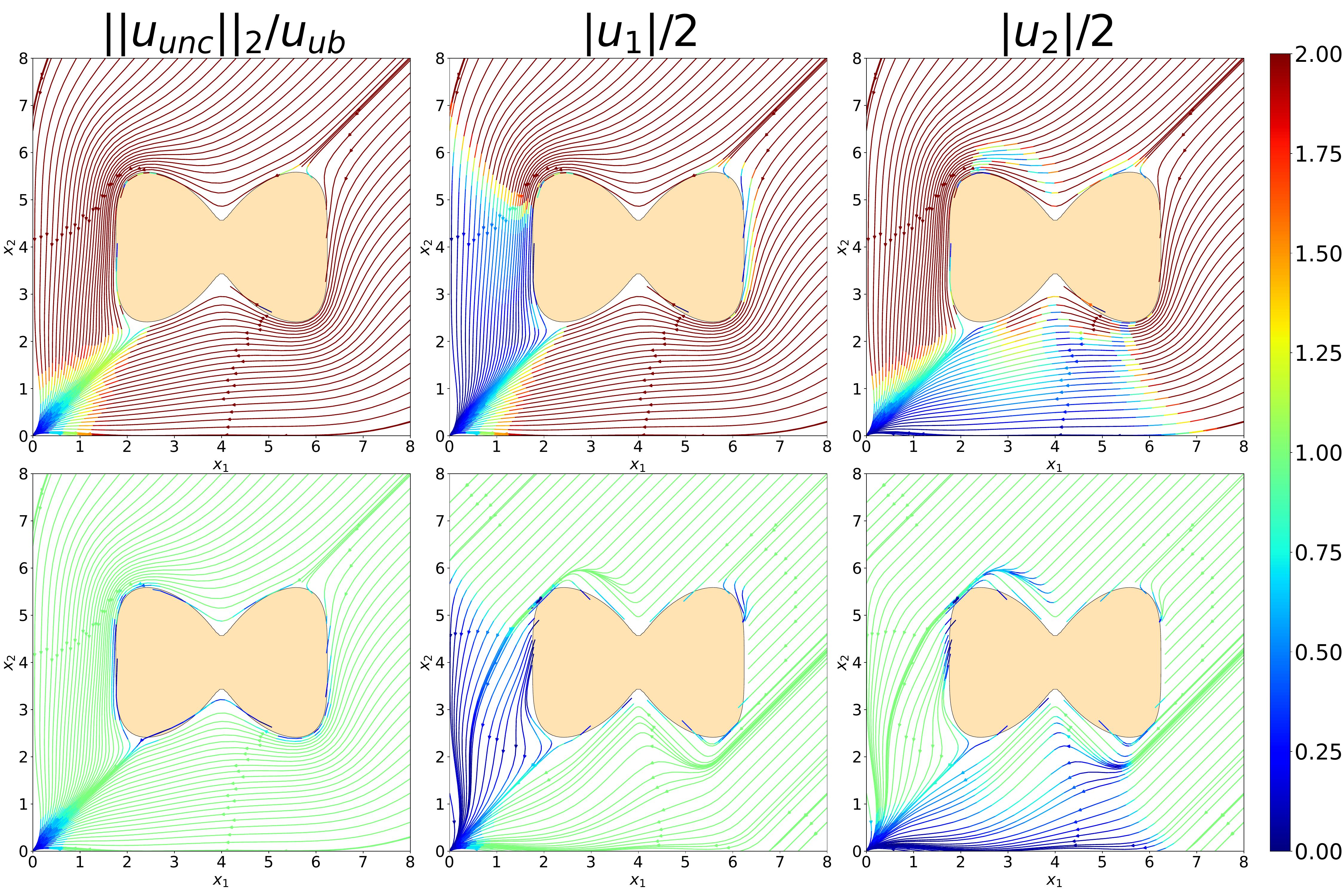}
     \caption{Unconstrained reference Mod-DS (first row) vs. speed-constrained (second row, left) and velocity-constrained reference Mod-DS with $U_\text{ub}= -U_\text{lb} = [2, 2]^\top$(second row, middle and right). The color codes for images in the left, middle, and right columns respectively represent $\frac{\|u_\text{unc}\|_2}{u_\text{ub}}$, $\frac{|u_1|}{2}$, and $\frac{|u_2|}{2}$, given that $u=[u_1, u_2]^\top$.}
        \label{fig:modulation constrained static}
\vspace{-15pt}
\end{figure}

The difference between the proposed method with the strategy in \cite{constrainedModu} is that ours considers the unconstrained Mod-DS outputs $n^\top u_\text{unc}$ when determining whether the robot is safe, which improves the smoothness of the trajectories generated by the constrained outputs and maximumly preserves the Mod-DS characteristics (see \autoref{fig:modulation constrained static}). 

\subsection{Velocity-Constraining Mod-DS} \label{appendix:vel-modulation}
Similarly, solving optimization problem in Eq.~\eqref{eq:lp velocity} offers solutions to enforcing velocity constraints $U_\text{lb}\leq u \leq U_\text{ub}$ for Mod-DS in fully actuated systems while guaranteeing safety. 
\begin{gather}
\label{eq:lp velocity}
u_\text{c} = \argmin_u \; \|u - u_\text{unc}\|_2^2\\
\nonumber U_\text{lb}\leq u \leq U_\text{ub}\\
\nonumber n^\top u \geq \min(n^\top \bar{\dot{x}}_{o},n^\top u_\text{unc})
\end{gather}

Unlike the speed-constraining Mod-DS, the QP problem for enforcing velocity constraints no longer has a straightforward explicit solution, because the complexity of the explicit solution derived from KKT conditions grows exponentially with the number of inequality constraints in the problem. Note that both speed-constraining and velocity-constraining Mod-DS methods proposed here remain valid for high dimensional state-spaces when $d>3$.

\subsection{Theoretical Guarantees} Both speed and velocity constraining Mod-DS preserve the \textit{impenetrability} guarantees of the original Mod-DS, under the assumption that the speed that the obstacle travels towards the agent is no more than the speed limit. We showcase stream plots in \autoref{fig:modulation constrained static} to demonstrate performances of speed-constraining and velocity-constraining Mod-DS compared to unconstrained Mod-DS. 

\begin{theorem}
Consider an obstacle whose boundary is defined as $h(x,x_o) = 0$. Any trajectory $x(t)$, $t\in [0,\infty)$ starting outside the obstacle boundary, i.e., $h(x(0),x_o(0)\geq 0$, evolving according to Eq.~\eqref{eq:ds-modulation} and being constrained by the Speed-Constraining and Velocity-Constraining approaches Mod-DS presented in Appendix \ref{appendix:speed-modulation} and \ref{appendix:vel-modulation}, will never penetrate the obstacle boundary, i.e., $h(x(t),x_o(t))\geq 0$, $\forall t\in [0,\infty)$. 
\end{theorem}
\noindent \textbf{Proof:} 
In \cite{khansari2012dynamical},~\cite{LukesDS}, and \cite{onManifoldMod}, the unconstrained outputs $u_\text{unc}$ from respectively normal, reference and on-manifold Mod-DS approaches are proven to achieve obstacle impenetrability in terms of the von Neuman boundary condition, listed in Eq.~\eqref{eq:Neuman}.
\begin{equation}
\label{eq:Neuman}
    n(x,x_o)^\top u_\text{unc} = 0 \quad \forall x \in \partial \mathcal{C}_o
\end{equation}

Given safe controller $u_\text{unc}$ in Eq.~\eqref{eq:Neuman}, we want to show that the constrained solutions $u_\text{c}$ of the QPs in \eqref{eq:qp speed} and \eqref{eq:lp velocity} also satisfies the set invariance condition of impenetrability. In other words, 
\begin{equation}
\label{eq:safe uc}
    n(x,x_o)^\top u_\text{c} \geq 0 \quad \forall x \in \partial \mathcal{C}_o
\end{equation}

Since any solutions $u_\text{c}$ from QPs in \eqref{eq:qp speed} and \eqref{eq:lp velocity} must satisfy the safety constraints in \eqref{eq:const safe}, $n(x,x_o)^\top u_\text{c} \geq n(x,x_o)^\top u_\text{unc}$ must be true for all state $x$. This implies that $n(x,x_o)^\top u_\text{c} \geq 0$ if $n(x,x_o)^\top u_\text{unc} = 0$. Therefore, the statement in \eqref{eq:safe uc} always holds. \hfill $\blacksquare$

\end{document}